\documentclass{article}
\usepackage{graphicx} % Required for inserting images
\usepackage[a4paper,top=3cm,bottom=2cm,left=2.3cm,right=2.3cm,marginparwidth=1.75cm]{geometry}
\usepackage{graphicx}
\usepackage{verbatim}
\usepackage{hyperref}       % hyperlinks
\usepackage{url}            % simple URL typesetting
\usepackage{booktabs}       % professional-quality tables
\usepackage{amsfonts}       % blackboard math symbols
\usepackage{nicefrac}       % compact symbols for 1/2, etc.
\usepackage{microtype}      % microtypography
\usepackage{amsmath}
\usepackage{multirow}
\usepackage{amsthm}
\usepackage{amssymb}

\usepackage{subfigure}
\usepackage{times}
\usepackage{color}
\usepackage{xcolor}

\usepackage[normalem]{ulem}
\useunder{\uline}{\ul}{}
\usepackage{algorithm}
\usepackage{algpseudocode}
\title{State-space models are accurate and efficient neural operators for dynamical systems}

\author{Zheyuan Hu\thanks{Department of Computer Science, National University of Singapore, Singapore, 119077 (\href{mailto:e0792494@u.nus.edu}{zheyuan.hu@u.nus.edu}, \href{mailto:shenqianli@u.nus.edu}{shenqianli@u.nus.edu}, \href{mailto:kenji@nus.edu.sg}{kenji@nus.edu.sg})} \and Nazanin Ahmadi Daryakenari\thanks{Center for Biomedical Engineering, School of Engineering, Brown University, Providence, RI 02912, USA (\href{mailto:nazanin\_ahmadi\_daryakenari@brown.edu}{nazanin\_ahmadi\_daryakenari@brown.edu})} \and Qianli Shen\footnotemark[1] \and Kenji Kawaguchi\footnotemark[1] \and   George Em Karniadakis\thanks{Division of Applied Mathematics, Brown University, Providence, RI 02912, USA (\href{mailto:george\_karniadakis@brown.edu}{george\_karniadakis@brown.edu})} \ \thanks{Advanced Computing, Mathematics and Data Division, Pacific Northwest National Laboratory, Richland, WA, United States}}
\date{}

\begin{document}

\maketitle

\begin{abstract}
Physics-informed machine learning (PIML) has emerged as a promising alternative to classical methods for predicting dynamical systems, offering faster and more generalizable solutions. However, existing models, including recurrent neural networks (RNNs), transformers, and neural operators, face challenges such as long-time integration, long-range dependencies, chaotic dynamics, and extrapolation, to name a few.
To this end, this paper introduces state-space models implemented in Mamba for accurate and efficient dynamical system operator learning. Mamba addresses the limitations of existing architectures by dynamically capturing long-range dependencies and enhancing computational efficiency through reparameterization techniques. 
To extensively test Mamba and compare against another 11 baselines, we introduce several strict extrapolation testbeds that go beyond the standard interpolation benchmarks. 
We demonstrate Mamba's superior performance in both interpolation and challenging extrapolation tasks. Mamba consistently ranks among the top models while maintaining the lowest computational cost and exceptional extrapolation capabilities.
Moreover, we demonstrate the good performance of Mamba for a real-world application in quantitative systems pharmacology for assessing the efficacy of drugs in tumor growth under limited data scenarios. 
Taken together, our findings highlight Mamba's potential as a powerful tool for advancing scientific machine learning in dynamical systems modeling.
(The code will be available at \url{https://github.com/zheyuanhu01/State_Space_Model_Neural_Operator} upon acceptance.) 
\end{abstract}

\section{Introduction}

A dynamical system describes a state that evolves over time according to a set of fixed rules that are dependent on its current state and inputs. These systems can be found across various fields, such as physics, biology, economics, engineering, etc. They allow us to model and understand complex real-world phenomena. For instance, climate models are dynamical systems that help predict weather patterns and climate change. The pharmacokinetic (PK) and pharmacodynamic (PD) models are essential tools in biomedicine and drug development to understand how drugs interact with biological systems. The Lorenz dynamical system \cite{lorenz1963deterministic} is a set of three nonlinear differential equations introduced initially by Edward Lorenz in 1963 to model atmospheric convection. 
%It is known for its chaotic solutions, leading to the famous ``butterfly effect," where small changes in initial conditions can result in vastly different outcomes. 
The Duffing system is used to model certain damped and driven oscillators, exhibiting complex dynamics, including chaotic behavior, and is characterized by a nonlinear restoring force, which is widely adopted in mechanical and electrical engineering.
Solving such dynamical systems is thus of critical importance as it enables us to model, predict, and understand complex phenomena in various scientific fields, providing insights into how systems evolve and respond to different conditions over time. However, since many critical dynamical systems cannot be solved analytically, numerical solutions are pursued using a plethora of computational methods.

Traditional numerical solvers, such as those based on multi-step and multistage methods, have been extensively used to solve dynamical systems. However, they come with certain limitations. 
For instance, traditional numerical solvers can be computationally expensive, especially for high-dimensional systems or systems requiring fine time-step resolution. 
In contrast, physics-informed machine learning (PIML) \cite{raissi2019physics,lu2019deeponet,li2020fourier,karniadakis2021physics} methods have emerged as a powerful alternative to traditional numerical solvers for solving dynamical systems \cite{robinson2022physics,saha2021physics}, and more generally, differential equations. 
Once trained, PIML models can provide real-time and mesh-free solutions significantly faster than traditional numerical methods \cite{ovadia2023ditto} and tackle high-dimensional systems where traditional methods fail due to the curse of dimensionality \cite{hu2023tackling}.
They can generalize well from training data, potentially providing accurate predictions even for unseen scenarios, demonstrating great progress in stiff chemical mechanics \cite{goswami2023learning}, shape optimization \cite{shukla2023deep}, undamped oscillators \cite{cao2023lno}, weather forecast \cite{ovadia2023ditto}, and materials science problems \cite{goswami2022physics}, to just name a few.

Despite the recent progress in PIML, designing a backbone model architecture to efficiently capture long-time integration, long-range dependencies, chaotic properties, and stiff solutions in dynamical systems remains challenging. Existing architectures, including time-series models like recurrent neural networks (RNNs) \cite{hochreiter1997long,cho2014learning}, Transformers \cite{vaswani2017attention,litransformer,hao2023gnot,cao2021choose}, and other general neural operators \cite{lu2019deeponet,li2020fourier,cao2023lno}, all have certain limitations.

RNNs are cost-efficient as their computational cost scale linearly with the sequence length or the dynamical system terminal time. However, they cannot capture long-term dependencies. This is because RNNs have a limited effective context window, meaning they can only consider a limited amount of preceding context due to their sequential nature.
Furthermore, RNNs process input sequentially in a step-by-step manner, which means they cannot take full advantage of the parallel processing capabilities of modern hardware like GPUs~\cite{sutskever2014sequence, gu2021efficiently, litransformer}. This makes RNNs slower to train, even if their model structures are relatively simple.

Transformers \cite{vaswani2017attention} capture long-term dependencies and thus have been the state-of-the-art for sequential modeling, widely adopted in computer vision \cite{liu2021swin} and natural language processing \cite{vaswani2017attention} communities. 
This is achieved via the self-attention mechanism, which computes the input sequence's attention score across all tokens/time steps that can directly relate distant positions in a sequence.
However, the cost of Transformers scales quadratically with the sequence length, rendering them inefficient in handling difficult tasks such as long-time integration of dynamical systems.

Although numerous efficient low-cost Transformers have been proposed to alleviate the quadratic cost, e.g., Galerkin attention \cite{cao2021choose}, whose cost scales linearly, they still face a serious tradeoff between model efficiency and capacity.  Relying on approximations of the self-attention mechanism, these efficient Transformers may not be as good as the original Transformer, especially for tasks heavily dependent on precise attention calculations.

Regarding neural operators, such as DeepONet \cite{lu2019deeponet}, LNO \cite{cao2023lno}, and FNO \cite{li2020fourier}, there has been great recent progress in approximating nonlinear ODE/PDE operators, but they fail to consider the temporal information in dynamical systems, handle variable-length input, or generalize to longer sequences and longer time integration \cite{michalowska2023neural}.

To this end, we introduce the state-space model and Mamba \cite{gu2023mamba,dao2024mamba2} for dynamical system operator learning.
Mamba~\cite{gu2023mamba} introduces a selection mechanism that dynamically adjusts to capture long-range dependencies and adopt reparameterization techniques~\cite{gu2021efficiently} for parallelization and better computational efficiency, demonstrating competitive empirical performance against Transformers across a wide range of benchmarks while maintaining linear-time efficiency in long-sequence modeling.
Compared with RNNs, Mamba maintains linear complexity with input sequence length and is more parallelizable.
More recently, Mamba2~\cite{dao2024mamba2} introduced a State-Space Duality (SSD) framework, establishing theoretical connections between State-Space Models (SSMs) and various attention mechanisms. Additionally, various refinements were introduced to the Mamba module, achieving a 2-8$\times$ speed-up without compromising performance.

Herein, we conduct extensive experiments on various dynamical systems and go beyond standard interpolation benchmarking, where the train and test sets follow the same distribution, as summarized in Table \ref{tab:experiment_summary_in_intro}. In addition to the relatively easier interpolation tests, we introduce nontrivial and strict testbeds containing the following unusual and challenging tasks:
(1) Finite regularity and discontinuous solution in Section \ref{sec:finite_regularity}.
(2) Chaotic Lorenz systems with or without an external force in Sections \ref{sec:lnoode} and \ref{sec:Lorenz_init}.
(3) Out-of-distribution generalization to unseen test solutions following a different distribution from the train solution distribution in Sections \ref{sec:lnoode} and \ref{sec:pk_pd_model_extrapolation}.
(4) Quantification of extrapolation errors by gradually moving the test distribution away from the train one in Section \ref{sec:quantify_extrapolation_error_zhu2023reliable}.
(5) Extrapolation to longer-time problems with training only in a shorter time in Section \ref{sec:extrapolation_long_time_1D_DS_deeponet}.
(6) Long-time integration in Section \ref{sec:long_time_integration_lno_pend}.
(7) Application of Mamba to a real-world dynamical system in Section \ref{sec:pk_pd_model}.
In terms of the tested dynamical system types, we consider an anti-derivative operator, nonlinear ODE, gravity pendulum, the Izhikevich neuron model, a tempered fractional LIF neuron model, Lorenz system, Duffing oscillator, and the PK/PD model, covering a diverse set of chaotic, undamped, stiff, and real-world systems.
Regarding models for comparison, we conduct comprehensive tests against GRU, LSTM, DeepONet, FNO, LNO, (conventional) Transformers, Oformer with Vanilla/Galerkin/Fourier attention, and GNOT.
Mamba performs stably across almost all dynamical systems and various problem types, and ranks in the top three except for only one test case, demonstrating consistent performance across various settings. Regarding computational and memory costs, Mamba's outstanding performance is achieved at a significantly lower cost than Transformers and at a similar cost to the light RNN models.
Most importantly, Mamba also demonstrates outstanding extrapolation capability, as well as efficiency and effectiveness in handling long-time integration problems, in addition to its in-distribution interpolation performances.
These results demonstrate Mamba's great potential in efficiently modeling dynamical systems and capturing long-range dependencies, long-time integration, and other complex phenomena in scientific machine learning compared with other state-of-the-art models. Overall, this paper opens the door for using novel state-space models and Mamba for dynamical system operator learning.

\begin{table}[htbp]
\centering
\begin{tabular}{|c|c|}
\hline
Section & Problems \\ \hline
$\S$\ref{sec:1D_DS_DeepONet} & 1D Dynamical Systems from DeepONet Benchmarks \\ \hline
$\S$\ref{sec:finite_regularity} & Finite Regularity and Discontinuous Solutions
 \\ \hline
$\S$\ref{sec:lnoode} & Out-of-Distribution Operator Learning Tasks \\ \hline
$\S$\ref{sec:long_time_integration_lno_pend} & Long-Time Integration \\ \hline
$\S$\ref{sec:extrapolation_long_time_1D_DS_deeponet} & Extrapolation to Long-Time Integration Problems \\ \hline
$\S$\ref{sec:Lorenz_init} &  Chaotic Lorenz System with Various Initial Conditions \\ \hline
$\S$\ref{sec:quantify_extrapolation_error_zhu2023reliable} &  Quantifying the Extrapolation Error \\ \hline
$\S$\ref{sec:pk_pd_model} & Real-World Application: Quantitative Systems Pharmacology \\ \hline
\end{tabular}
\caption{Eight problems tested in this paper.}
\label{tab:experiment_summary_in_intro}
\end{table}

\section{Related Work}

In the following, we briefly review neural operators, transformers, and state-space models.

\subsection{Neural Operators}
Neural operators are a special class of machine learning models designed to learn (nonlinear) mappings between infinite-dimensional spaces, making them particularly suitable for efficiently solving partial differential equations (PDEs). 
The dynamical systems problem under consideration in this paper also falls into this field. 
Unlike traditional numerical methods that require mesh generation and can be computationally intensive, neural operators learn a representation of the PDE solution operator directly, allowing for fast evaluations of new inputs once trained. This approach provides significant advantages in terms of generalization and scalability, enabling more rapid and flexible solutions across a variety of complex scenarios in physics and engineering.

The first neural operator, DeepONet \cite{lu2019deeponet}, is based on the operators' universal approximation theorem \cite{chen1995universal}. It is a general mesh-free neural operator containing a branch net encoding the input function and a trunk net encoding the query domain.
The Fourier neural operator (FNO) \cite{li2020fourier} adopts the trainable kernel integral operator to transform the input function into the output function on a uniform grid to approximate the unknown PDE operator, which can be efficiently computed in the Fourier space via fast Fourier transforms (FFTs).
Lu et al. \cite{lu2021comprehensive} compared  DeepONet and FNO systematically on multiple representative benchmarks and demonstrated that they perform similarly in relatively simple test cases. In contrast, in irregular geometries and more complicated cases, DeepONet can outperform FNO, especially in high dimensions.
In addition to the data-driven approach, DeepONet can also incorporate physics into its training, obtaining physics-informed DeepONet \cite{wang2021learning,goswami2023physics}, which demonstrates better generalization thanks to the physics-informed regularization under data-limited test cases.
Neural operator's theoretical approximation rates have also been studied extensively in \cite{deng2022approximation,franco2023approximation}. 
The Laplace neural operator (LNO) \cite{cao2023lno} also considers the trainable kernel integral operator, but in Laplace space, and parameterizes the trainable kernel in the pole-residue form.
The hyena neural operator (HNO) \cite{patil2023hno} is based on the hyena model \cite{pmlr-v202-poli23a}, which is one particular kind of SSM utilizing long convolutions and element-wise multiplication gating components. HNO further combines it with a Transformer-based encoder-decoder model architecture and demonstrates comparable performance with FNO \cite{li2020fourier} on two PDEs.
Our approach differs in model type and experimental setup.
We employ a different and state-of-the-art SSM, namely Mamba \cite{gu2023mamba}, to dynamical system operator learning, conduct eight comprehensive tests beyond standard interpolation tasks to extrapolate, and outperform other strong baselines. 
The Markov neural operator \cite{li2021learning} learns chaotic systems based on their Markov property by conducting time-step by time-step prediction and employing a dissipativity regularization.

Recurrent neural networks (RNNs), including long-short term memory (LSTM) \cite{hochreiter1997long} and gated recurrent unit (GRU) \cite{cho2014learning}, are popular sequentially modeling architectures suitable for time-dependent PDE/ODE operator learning. 
Michałowska et al. \cite{michalowska2023neural} utilized RNNs for long-time integration in dynamical systems, while Vlachas et al. \cite{vlachas2020backpropagation} applied them to learning chaotic spatiotemporal systems.

Furthermore, Transformers \cite{vaswani2017attention} have gained popularity as neural operator models due to their ability to model complex dependencies and interactions across inputs through self-attention mechanisms, which is crucial for accurately learning the underlying dynamics of PDEs \cite{shih2024transformers}.
However, the Transformer's quadratic cost with the input sequence length can be a substantial computational bottleneck in modeling dynamical systems, especially when dealing with long-time integration or large-scale PDE problems.
The Galerkin Transformer \cite{cao2021choose} applied Transformers to neural operator learning for the first time and improved their efficiency by proposing the linear cost of Galerkin and Fourier self-attentions based on the operator approximation theory in the Hilbert space.
Oformer \cite{litransformer} further adapted the Transformer to neural operator learning via combining self-attention, cross-attention between input function and query point, and point-wise multi-layer perception transformations. The attention mechanism in Oformer can also be vanilla quadratic attention or the more efficient Galerkin/Fourier attention.
GNOT \cite{hao2023gnot} proposes a heterogeneous normalized attention mechanism to incorporate multiple inputs and irregular meshes. Furthermore, GNOT's geometric gating mechanism adopting a soft domain decomposition \cite{hu2023augmented} can efficiently model multi-scale and multi-physics problems.
Finally, diffusion-inspired Temporal Transformer Operator \cite{ovadia2023ditto} models extrapolation in time-dependent PDEs via Transformer blocks and temporal conditioned residual blocks.

\subsection{Transformers}
Transformers~\cite{vaswani2017attention} revolutionized sequence modeling by discarding the recurrent structure in favor of a self-attention mechanism. The self-attention mechanism enables Transformers to directly model dependencies between any two positions in a sequence, allowing for better parallelization and scalability. The original Transformer architecture, particularly the encoder-decoder structure, has been exceptionally successful in machine translation. Subsequent adaptations, such as BERT~\cite{devlin2018bert} and GPT~\cite{brown2020language}, have achieved state-of-the-art results in a wide range of NLP tasks, including text understanding and text generation. Despite their success, the quadratic complexity of the self-attention mechanism remains a challenge for processing long sequences.

\subsection{State-Space Models and Mamba}
General state-space models (SSMs) utilize state variables to represent and model dynamical systems. In the context of deep learning, SSMs refer to a family of methodologies that involve recurrent updates on hidden state variables to model the dynamics of sequential data. The most emblematic examples of early SSMs are Recurrent Neural Networks (RNNs), which dominated sequence modeling tasks prior to the advent of Transformers~\cite{vaswani2017attention}. However, as emphasized in~\cite{sutskever2014sequence, gu2021efficiently, litransformer}, the intrinsic sequential nature of RNNs fundamentally restricts their capacity for parallelization during the training process.

Structured state space models (S4) have introduced significant advancements in computational efficiency through reparameterization techniques~\cite{gu2021efficiently}, providing an efficient alternative to the traditional attention mechanisms used in transformers. Prominent variants of S4 models include Linear Attention~\cite{katharopoulos2020transformers}, H3~\cite{fu2022hungry}, Gated State Space~\cite{mehta2022long}, Hyena~\cite{nguyen2024hyenadna}, RetNet~\cite{sun2023retentive}, and RWKV~\cite{peng2023rwkv}.

\begin{figure}[htbp]
\centering
\includegraphics[width=0.7\linewidth]{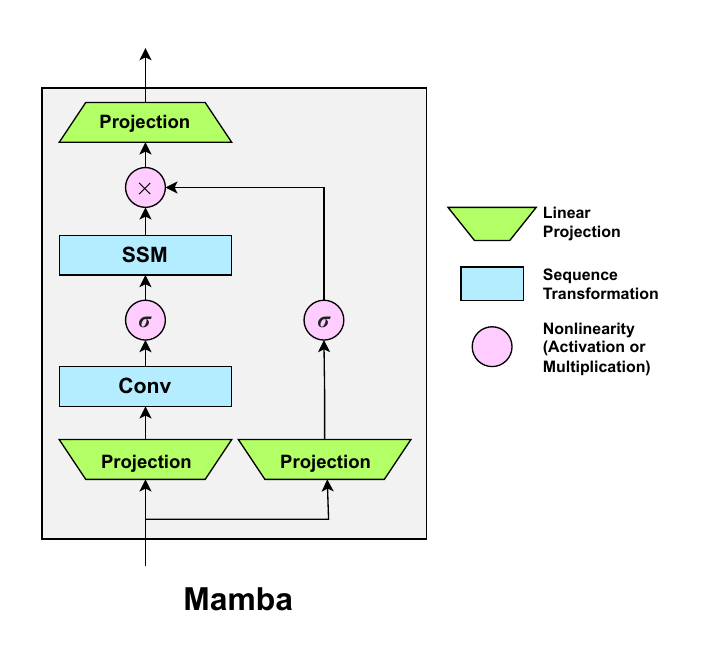}
\caption{\textbf{The Mamba block architecture} is constructed based on SSM's mathematical formulation in equation (\ref{eq:ssm_continuous}) corresponding to the blue SSM block (Sequence Transformation) plus other additional components to strengthen the model.
The figure is based on Figure 3 (right) following Gu and Dao \cite{gu2023mamba}.
Specifically, a Mamba block contains two branches.
The left branch is SSM-related. It first uses a green linear projection to map the input sequence's each time step to have more feature channels.
Then, a blue one-dimensional convolution block (Conv), a nonlinear activation $\sigma$, and finally, the SSM block are applied sequentially.
The SSM block intakes the input sequence and model parameters $\boldsymbol{A}, \boldsymbol{B}, \boldsymbol{C}$ for computation. 
Furthermore, another right branch is the skip connection \cite{he2016deep} branch, which is a linear projection followed by a nonlinear activation.
The results from the two branches are multiplied and linearly transformed into the final output.
}
\label{fig:mamba}
\end{figure}

The recently developed Mamba framework~\cite{gu2023mamba} builds on the S4 models by introducing a selection mechanism that dynamically adjusts to capture long-range context, see Figure \ref{fig:mamba}. Mamba has demonstrated competitive empirical performance against Transformers across a wide range of benchmarks while maintaining linear-time efficiency in long-sequence modeling.
Mamba-2~\cite{dao2024mamba2} introduced a 
State-Space Duality (SSD) framework, establishing theoretical connections between SSMs and various attention mechanisms. Additionally, it proposed refinements to the Mamba module, achieving a 2-8$\times$ speed-up without compromising performance.
%
% \begin{figure}[h!]
% \centering
% \includegraphics[width=\linewidth]{mamba.pdf}
% \caption{\textbf{Mamba and Mamba-2 model architectures:} The Mamba-2 architecture introduces significant simplifications to the original Mamba block by eliminating sequential linear projections. In the Mamba-2 design, the SSM parameters A, B, and C are computed at the outset of the block rather than being dynamically generated based on the SSM input X. To enhance stability, Mamba-2 incorporates an additional normalization layer, inspired by the NormFormer approach \cite{shleifer2021normformer}. Moreover, the B and C projections in Mamba-2 utilize a single head shared across all X heads, optimizing the architecture. Mamba requires two all-reduce operations per layer for tensor model parallelism, while Mamba-2 reduces this to one, thereby improving computational efficiency, particularly with a parallel size of two.}
% \label{fig:mamba}
% \end{figure}

\section{Methods for Comparative Study}
This section defines our dynamical system operator learning problem and introduces the main Mamba and SSM models, which serve as the neural operators.
Furthermore, we introduce competitive models for time-series modeling, including recurrent neural networks (RNNs) and Transformers. We compare and contrast them with the proposed Mamba model.
We also briefly introduce other popular neural operators, including DeepONet, Laplace neural operator, and Fourier neural operator, which will serve as solid baselines in our computational experiments.

\subsection{Notation and Problem Definition}
The unknown neural operator is denoted by  $\mathcal{O}$, while the input function is denoted $x(t)$, and the output is $y(t) = \left(\mathcal{O}x\right)(t)$.
The input functions and their corresponding output functions are represented by their pointwise values on a discrete temporal grid $\left\{ t_1, t_2, \cdots, t_{N_{\text{grid}}} \right\}$.
This motivates us to formulate the problem as a sequence-to-sequence learning problem from the input sequence $\left\{ x(t_1), x(t_2), \cdots, x(t_{N_{\text{grid}}}) \right\}$ to the output sequence $\left\{ y(t_1), y(t_2), \cdots, y(t_{N_{\text{grid}}}) \right\}$.
In the following, we will introduce the proposed SSM and Mamba, as well as other sequential modeling models, including recurrent neural nets (RNNs) and transformers. 
In addition to these sequential models, since dynamical systems can also be viewed as a special class of ODEs, we also review neural operators for general ODE/PDE operator learning.

\subsection{State Space Model (SSM) and Mamba}

The state-space model (SSM) was originally introduced as a continuous-time sequence-to-sequence (seq2seq) mapping:

\begin{equation}\label{eq:ssm_continuous}
    \begin{aligned}
    h'(t) &= \boldsymbol{A} h(t) + \boldsymbol{B} x(t),\\
    y(t) &= \boldsymbol{C} h(t),
\end{aligned}
\end{equation}
to map the input sequence $x(t)$ to $y(t)$ via a hidden state $h(t)$.
In real-world applications, data is commonly discrete, necessitating the discretization of equation~\eqref{eq:ssm_continuous}.
We follow~\cite{gu2023mamba} to employ the zero-order hold (ZOH) discretization rule defined in
\begin{equation}\label{eq:ssm_recurrence}
    \begin{aligned}
    h_t &= \boldsymbol{\bar{A}} h_{t-1} + \boldsymbol{\bar{B}} x_t,\\
    y_t &= \boldsymbol{C} h_t,
\end{aligned}
\end{equation}
where $\boldsymbol{\bar{A}} := \exp (\Delta \cdot \boldsymbol{A})$, $\boldsymbol{\bar{B}} := (\Delta \cdot \boldsymbol{A})^{-1} (\exp(\Delta \cdot \boldsymbol{A}) - I) \cdot \Delta\boldsymbol{B}$, and $\Delta$ is the discretization step size.
The recurrence formulas~\eqref{eq:ssm_recurrence} inherently enable the modeling of unbounded context and linear-time inference, but they are not conducive to efficient parallel training.
Fortunately, equation~\eqref{eq:ssm_recurrence} can be further rewritten using a convolution formula as
\begin{equation}\label{eq:ssm_convolution}
    \begin{aligned}
        \boldsymbol{\bar{K}} &= (\boldsymbol{C}\boldsymbol{\bar{B}},
        \boldsymbol{C}\boldsymbol{\bar{A}\bar{B}},\dots,\boldsymbol{C}\boldsymbol{\bar{A}}^k\boldsymbol{\bar{B}},\dots),\\
        \boldsymbol{y} = \begin{pmatrix}y_0\\y_1\\ \vdots \\ y_k \\ \vdots \end{pmatrix} = \boldsymbol{u} * \boldsymbol{\bar{K}} &:= \begin{pmatrix} \boldsymbol{C}\boldsymbol{\bar{B}} & \bold{0} & \cdots & \bold{0}\\ \boldsymbol{C}\boldsymbol{\bar{A}\bar{B}} & \boldsymbol{C}\boldsymbol{\bar{B}}  & \cdots & \bold{0}\\ \vdots & \vdots & \ddots & \vdots \\ \boldsymbol{C}\boldsymbol{\bar{A}}^k\boldsymbol{\bar{B}} & \boldsymbol{C}\boldsymbol{\bar{A}}^{k-1}\boldsymbol{\bar{B}} & \cdots & \bold{0} \\ \vdots & \vdots & \vdots & \vdots \end{pmatrix} \begin{pmatrix}x_0\\x_1\\ \vdots \\ x_k \\ \vdots\end{pmatrix},
    \end{aligned}
\end{equation}
where $*$ denotes the convolution operator, and $\boldsymbol{\bar{K}}$ is referred to as the convolution kernel, which can be efficiently computed by the Fast Fourier Transform (FFT).
The convolution operation is well supported by modern hardware, thus enabling efficient parallel training with careful implementation.
Mamba~\cite{gu2023mamba} introduces a selection mechanism to further overcome the fundamental limitations of SSMs' Linear Time Invariance (LTI) property, achieving competitive empirical performance against Transformers while maintaining linear-time efficiency in long-sequence modeling.

We present Mamba's model architecture in Figure \ref{fig:mamba} following Figure 3 (right) in Gu and Dao \cite{gu2023mamba}.
A Mamba block contains two branches.
The left one is the SSM branch, containing linear transformation, convolution, nonlinearity, and the SSM transformation corresponding to equation (\ref{eq:ssm_continuous}).
The right is the skip connection block, consisting of a linear transformation and a nonlinear activation.
Then, the left and right outputs are merged via multiplication.
The entire Mamba model is constructed via stacking multiple Mamba blocks in Figure \ref{fig:mamba} sequentially.

\subsection{Transformers}
A Transformer block consists of two modules: a self-attention module $\mathsf{Attn}(\cdot)$ and a feed-forward network module $\mathsf{FFN}(\cdot)$, each with a shortcut~\cite{he2016deep}.
More formally, let $X \in \mathbb{R}^{n \times d}$ denote the input sequence and $Y \in \mathbb{R}^{n \times d}$ denote the output sequence, where $n$ denotes the sequence length, and $d$ denotes the hidden size, then the forward process of a Transformer block can be defined as
\begin{equation}
    X' = \mathsf{Attn}(X) + X, \quad Y = \mathsf{FFN}(X') + X'.\\
\end{equation}
The core innovation of Transformers~\cite{vaswani2017attention} lies in the self-attention mechanism, which enables the model to capture long-range dependencies and contextual relationships more effectively:
\begin{equation}
\begin{aligned}
    \quad \mathsf{Attn}(X) = \text{Attention}(W_Q X, W_K X, W_V X), \\
\end{aligned}
\end{equation}
where $W_Q, W_K, W_V \in \mathbb{R}^{d \times d}$ denote trainable linear mappings. 
The most widely recognized choice for the attention function is the scale-dot softmax attention:
\begin{equation}\label{eq:attention}
\begin{aligned}
    \text{Attention}(Q, K, V) = \text{softmax}\left(\frac{QK^T}{\sqrt{d}}\right)V,
\end{aligned}
\end{equation}
where $Q, K, V \in \mathbb{R}^{n \times d}$ denote the query, key, and value matrix, respectively; $n$ denotes the sequence length, and $d$ denotes the hidden size.
The feed-forward networks consist of two linear transformations with a $\mathsf{ReLU}$ activation in between
\begin{equation}
    \mathsf{FFN}(X') = \mathsf{ReLU}(X' W_1 + B_1) W_2 + B_2.
\end{equation}

The efficiency bottleneck of Transformers lies in the computation of the scale-dot softmax attention defined in~\eqref{eq:attention}, which requires computing the attention scores for every pair of tokens in the sequence, leading to computational complexity of $\mathcal{O}(n^2d)$ and memory complexity of $\mathcal{O}(n^2)$.
The quadratic complexity in terms of the sequence length hinders the application of Transformers in long-context tasks.

To address the efficiency limitation, several attention variants with a linear complexity have been recently proposed and studied.
In the context of operator learning, a Galerkin-type attention was first introduced by \cite{cao2021choose} as a linear variant of attention and later employed by~\cite{litransformer,hao2023gnot},
\begin{equation}\label{eq:g_attention}
\begin{aligned}
    \text{Attention}_{G}(Q, K, V) &= \frac{Q(\tilde{K}^T\tilde{V})}{d}, &\text{(Galerkin-type Attention)}
\end{aligned}
\end{equation}
where $\tilde{\cdot}$ denotes layer normalization~\cite{ba2016layer}.
The Galerkin-type attention involves two matrix product operations, which have complexity $\mathcal{O}(nd^2)$, remitting the dependence on the sequence length from quadratic to linear.

\subsection{Recurrent Neural Networks}

Recurrent Neural Networks (RNNs) have been pivotal in sequence modeling due to their ability to maintain temporal dependencies through recurrent connections. 
The classic RNNs can be formulated as follows with input sequence $x_t$, hidden state $h_t$, and output sequence $y_t$:
\begin{equation}\label{eq:rnn}
    \begin{aligned}
        h_t &= f(h_{t-1}, x_t),\\
    y_t &= g(h_t),
    \end{aligned}
\end{equation}
where $f$ is the hidden-to-hidden mapping and $g$ is the hidden-to-output mapping, both typically involving non-linear operations.
Classic RNNs are well-known to suffer from vanishing and exploding gradients~\cite{bengio1994learning, pascanu2013difficulty}.
To address these limitations, advanced variants such as Long Short-Term Memory (LSTM) networks~\cite{hochreiter1997long} and Gated Recurrent Units (GRUs)~\cite{cho2014learning} were developed by introducing gating mechanisms.

We note that equation~\eqref{eq:rnn} is closely related to the recurrence formulas of SSMs in equation~\eqref{eq:ssm_recurrence}, except for the non-linearity of $f$ and $g$.
Despite the expressivity gained, the non-linearity hinders parallelization and thereby significantly limits their efficiency and applicability in large-scale and real-time tasks.
In addition, the gating mechanisms employed by LSTM and GRUs are closely connected to the discretization of continuous-time systems as discussed in~\cite{funahashi1993approximation, tallec2018can} and are instances of selection mechanism introduced in~\cite{gu2023mamba}.

\subsection{Neural Operators}
In this subsection, we review neural operators (NOs), including DeepONet \cite{lu2019deeponet}, Laplace neural operator (LNO) \cite{cao2023lno}, and Fourier neural operator (FNO) \cite{li2020fourier}. These NOs have been popular choices in learning PDE operators and will serve herein as strong baselines in our computational experiments.
Compared with neural operators such as DeepONet, LNO, and FNO, sequential modeling models like Mamba take good advantage of the time-dependency. More specifically, the dynamical system solution at time $t$ only depends on the input function from zero to $t$ but does not rely on a later time. As a sequential modeling architecture, Mamba naturally encodes this prior knowledge into its prediction. In contrast, neural operators fail to do that and may be less generalizable and prone to overfitting.

Technically, DeepONet \cite{lu2019deeponet} was the first neural operator relying on the universal approximation theorem of operators \cite{chen1995universal}. It is a general mesh-free neural operator containing a branch network that encodes the input function's pointwise values into a latent vector, and another trunk network encodes the query point into another latent vector. Then, the inner product between them forms the final DeepONet's output.
FNO \cite{li2020fourier} and LNO \cite{cao2023lno} adopt the trainable kernel integral operator to transform the input function into the output function on a uniform grid, where FNO computes the integral and parameterizes the kernel in the Fourier space. At the same time, LNO does this in the Laplace space and Laplace transform.
A detailed mathematical formulation of these operators is presented in Appendix \ref{appendix:baseline_models_NO}.

\subsection{Application of Mamba to dynamical systems}
As a general sequence-to-sequence mapping, the application of Mamba is versatile and can be utilized in the following scientific computing scenarios. We will present multiple diverse test cases in our computational experiments.
\begin{enumerate}
\item Sequence-to-sequence mapping, applied to dynamical systems with external forces or other related inputs.
\item Solving ODEs with initial conditions only with or without the external forces. If we are given the ODE initial conditions solely, then we can construct the input sequence as $u(t) = u_0$, i.e., a constant sequence.
\item Other additional ODE information, including the ODE coefficients and initial conditions, can be concatenated to the input sequence and modeled to solve a bunch of diverse ODEs. Specifically, suppose the input sequence is $u(t)$, and the additional PDE coefficient/parameter is $\alpha$, which can be a high-dimensional vector. Then, the concatenated input to the Mamba model will be $\Tilde{u}(t) = [u(t), \alpha]$, which is a vector-valued input function of $t$.
\end{enumerate}

\section{Computational Experiments}
We conduct all experiments on an A100 GPU with 80 Gigabytes (GB) memory.

\subsection{1D Dynamical Systems from DeepONet Benchmarks}\label{sec:1D_DS_DeepONet}
Following the first neural operator, DeepONet \cite{lu2019deeponet}, we consider the 1D dynamical systems:
\begin{align}
s'(t) = g\left(s(t), u(t), t\right), \quad t \in [0, 1],
\end{align}
with an initial condition $s(0) = 0$. Our goal is to predict $s(t)$ over the whole domain $t \in [0, 1]$ for the input $u(t)$, i.e.,  we learn the operator $u(t) \mapsto s(t)$, given different $g$ to construct various test cases:
\begin{itemize}
\item Linear \textbf{antiderivative operator} case: $g\left(s(t), u(t), t\right) = u(t)$.
\item \textbf{Noninear ODE} case: $g\left(s(t), u(t), t\right) = u(t)^2$.
\end{itemize}
In addition, we test the \textbf{gravity pendulum} with an external force
\begin{align}
s_1'(t) = s_2(t), \quad s_2'(t) = -\sin\left(s_1(t)\right) + u(t), \quad t \in [0, 1],
\end{align}
with an initial condition $s_1(0) = s_2(0) = 0$. We aim to learn the operator $u(t) \mapsto s_1(t)$.

We describe the implementation in detail. We generate 10K train, {\color{green!50!black}validation}, and test data with a sensor/discretization size of 100, i.e., the sensors are on $\{0.01, 0.02, \cdots, 0.99, 1\}$, where the input function $u(t)$ and the target function $s(t)/s_t(t)$ are discretized. 
Following the original setting in DeepONet, the input functions $u(t)$ are randomly drawn from a mean zero Gaussian random field (GRF) with a length scale of 0.2, i.e., the input functions follow the distribution $\mathcal{GP}\left(0, k_l(x, y)\right)$ where $k_l(x, y) = \exp\left(-\frac{(x - y)^2}{2 l^2}\right)$ is the RBF kernel with length scale $l$ and $\mathcal{GP}$ denotes a Gaussian process. The output function $s(t)$ or $s_1(t)$ can correspondingly be solved via traditional numerical methods.
The train loss and the test metric are all the mean square errors (MSE).
We train all models for 10,001 epochs. 
In each epoch, we traverse the dataset with a minibatch size of 128.
We use the Adam optimizer \cite{kingma2014adam} with a 1e-3 initial learning rate, which decays linearly to zero at the end of the training.
We keep the parameter numbers of different models similar for a fair comparison. 
We run the code for five independent random seeds and report the average.

The model structure hyperparameter settings are summarized below.
{\color{green!50!black} We set an approximate parameter budget of 10,000.
The time series model search space is model depth in $\{1,2,3\}$, and hidden dimensions satisfying the budget. We set all Transformer-related models' head numbers as four and the number of experts in GNOT as two.
We set the Transformer and Mamba feedforward dimensions as their respective hidden dimension.
We opt for one-layer models with various hidden dimensions to satisfy the budget for all time series models, including LSTM, GRU, Mamba, and Transformers.
Specifically, LSTM, GRU, Mamba, and Mamba2 are one-layer models with 32 hidden dimensions.
Transformer \cite{vaswani2017attention} has 40 hidden and feedforward dimensions, four heads, and one layer.
For all three Oformers with vanilla, Galerkin, or Fourier attention \cite{litransformer}, we opt for one-layer Oformers with 24 hidden and feedforward dimensions and four heads.
GNOT has one layer, 16 hidden and feedforward dimensions, four heads, and two experts.
We set the same depth for these time series models for a fair comparison, and we use the validation dataset to verify that shallow and wide time series models are better than deep ($2\sim3$ layers) and narrow ones in terms of both speed and error.
Next, we report the hyperparameter choice on NOs.
We keep DeepONet's trunk and branch nets as MLP with the same structure, whose depth is $2\sim5$. We find the deepest model is the best, i.e., DeepONet's \cite{lu2019deeponet} trunk and branch nets are all five-layer networks with 24 hidden dimensions.
FNO \cite{li2020fourier} is a four-layer model with 16 hidden dimensions and eight modes. We validate other FNO models with $2\sim4$ layers, $8/16$ modes, and a corresponding hidden dimension to satisfy the parameter count budget and choose the best one.
We follow Cao et al. \cite{cao2023lno} and choose a one-layer LNO following their experiments on 1D dynamical systems. LNO has 16 hidden dimensions and eight modes.
We validate other LNO models with one layer, $4/8/16$ modes, and a corresponding hidden dimension to satisfy the parameter count budget and choose the best one.
}
\begin{table}[htbp]
\small
\centering
\begin{tabular}{|c|c|c|c|c|c|c|}
\hline
Model & Params & Memory & Time & Anti-Derivate & Nonlinear ODE & Pendulum \\ \hline
LSTM & 8545 & 1643MiB & 31min & 1.625E-7$\pm$1.185E-7 & 9.509E-6$\pm$8.892E-6 & 1.322E-8$\pm$9.719E-9 \\ \hline
GRU & 6433 & 1629MiB & 30min & 3.627E-8$\pm$2.771E-8 & 3.215E-6$\pm$2.276E-6 & 7.117E-9$\pm$1.953E-9 \\ \hline
DeepONet & 7272 & 1381MiB & 34min & 1.296E-7$\pm$4.094E-8 & 2.147E-5$\pm$1.435E-6 & 3.500E-8$\pm$3.479E-8 \\ \hline
FNO & 9329
 & 1833MiB & 73min & 1.536E-4$\pm$6.927E-6 & 6.519E-4$\pm$1.032E-4 & 4.008E-5$\pm$8.730E-6 \\ \hline
LNO & 6721 & 1797MiB & 52min & 2.214E-7$\pm$1.928E-7 & 4.775E-5$\pm$1.944E-5 & 1.073E-8$\pm$4.329E-9 \\ \hline
Transformer & 10201 & 1443MiB & 49min & 5.197E-5$\pm$2.183E-5 & 4.314E-4$\pm$1.098E-5 & 6.786E-6$\pm$7.092E-7 \\ \hline
Oformer-V & 11209 & 1439MiB & 41min & 4.210E-8$\pm$1.139E-8 & 1.352E-6$\pm$6.782E-7 & 7.549E-9$\pm$4.378E-9 \\ \hline
Oformer-G & 11161 & 1379MiB & 49min &\textcolor{blue}{3.472E-8$\pm$3.089E-9} & 6.036E-7$\pm$1.256E-7 & \textcolor{blue}{4.582E-9$\pm$1.115E-9} \\ \hline
Oformer-F & 11161 & 1379MiB & 49min & 3.598E-8$\pm$6.595E-9 & \textcolor{blue}{4.582E-7$\pm$9.735E-8} & \textcolor{red}{4.302E-9$\pm$1.538E-9} \\ \hline
GNOT & 11187 & 1403MiB & 91min & 4.155E-8$\pm$7.239E-9 & 6.253E-7$\pm$2.714E-7 & 5.547E-9$\pm$1.246E-9 \\ \hline
Mamba & 6593 & 1425MiB & 35min & \textbf{3.333E-9$\pm$1.288E-9} & \textbf{4.351E-8$\pm$1.782E-8} & \textbf{1.301E-9$\pm$3.273E-10} \\ \hline
Mamba2 & 6009 & 1282MiB & 109min & \textcolor{red}{2.106E-8$\pm$8.198E-9} & \textcolor{red}{2.015E-7$\pm$9.302E-8} &  4.464E-8$\pm$3.981E-8 \\ \hline
\end{tabular}
\caption{Test MSE error results for 1D dynamical systems, including the anti-derivative operator, nonlinear ODE, and gravity pendulum, from DeepONet's benchmark cases \cite{lu2019deeponet} corresponding to Section \ref{sec:1D_DS_DeepONet} in this paper. The errors are MSE following Lu et al. \cite{lu2019deeponet}. 
Optimization trajectories are visualized in Figure \ref{fig:1D_DS_loss}.
\textbf{The best model is bold}, \textcolor{red}{the second best is red}, and \textcolor{blue}{the third best is blue}.}
\label{table:1D_DS}
\end{table}

\begin{figure}[htbp]
\centering
\includegraphics[width=0.32\linewidth]{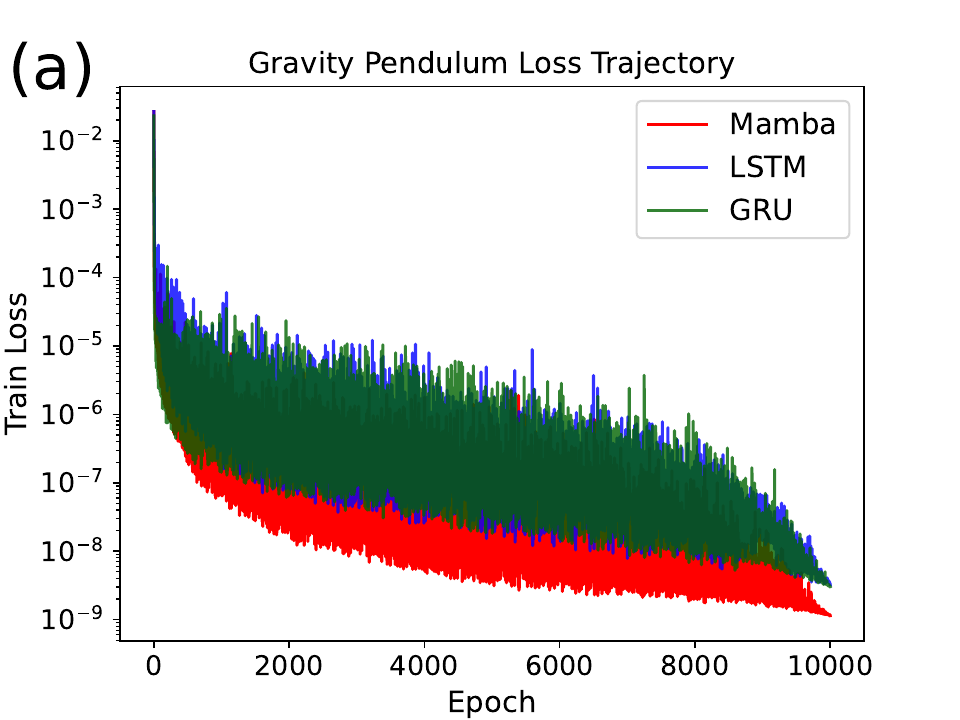}
\includegraphics[width=0.32\linewidth]{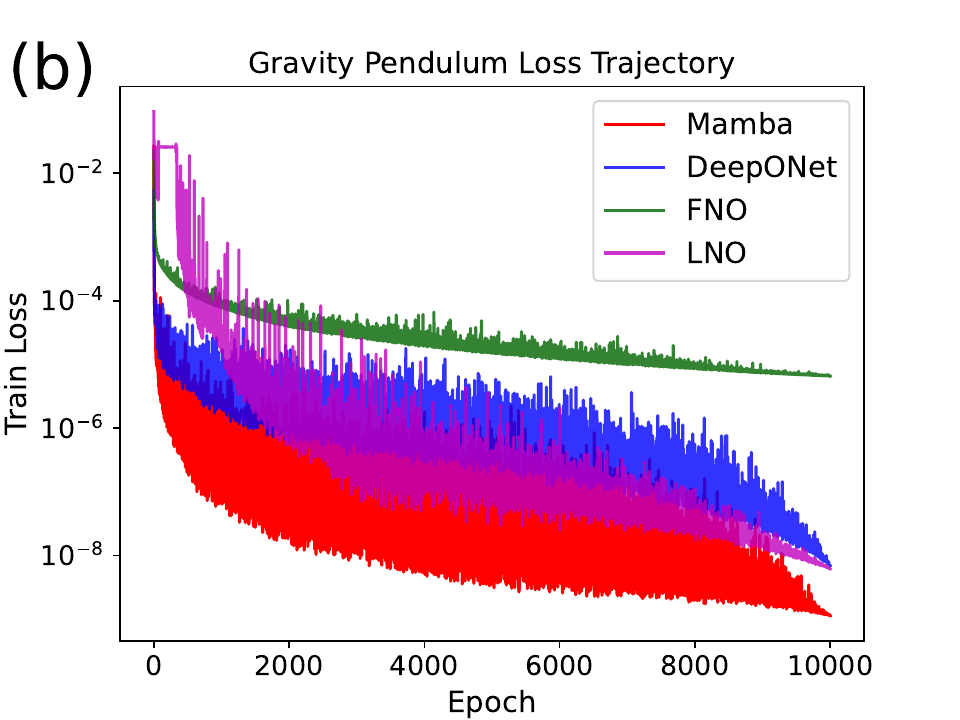}
\includegraphics[width=0.32\linewidth]{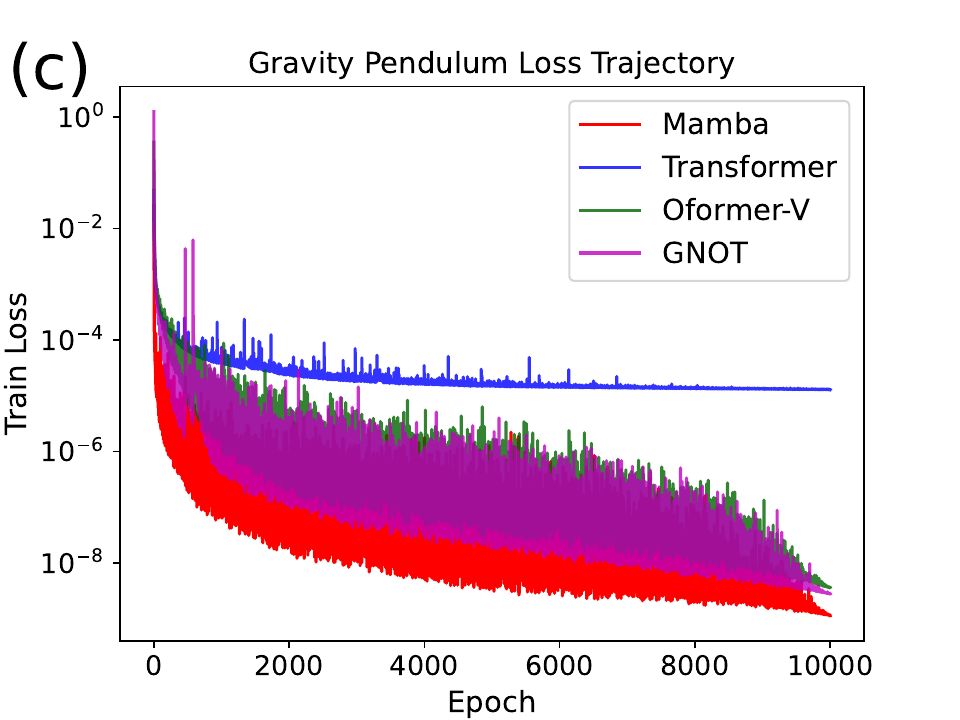}
\caption{{\color{green!50!black}Visualization of various models' loss trajectories for the 1D Pendulum system following Lu et al. \cite{lu2019deeponet} corresponding to Section \ref{sec:1D_DS_DeepONet} in this paper. The detailed quantitative results for this setting are presented in Table \ref{table:1D_DS}. 
Subfigure (a) is Mamba versus RNNs; subfigure (b) is Mamba versus NOs; and subfigure (c) is Mamba versus Transformers.}}
\label{fig:1D_DS_loss}
\end{figure}

Table \ref{table:1D_DS} shows the computational results, where we report the model name, the number of total parameters in the model, the memory cost in Mebibyte (MiB) obtained from the command ``nvidia-smi" on the Linux GPU server where only the training part is considered into the memory cost evaluation, the total training time in minutes (min), and the model MSE prediction error on the three operator learning tasks.  The MSE error metric was initially adopted by Lu et al. \cite{lu2019deeponet}, and we strictly follow them for a clear comparison.

The main experimental observations are summarized as follows.
\begin{itemize}
\item Regarding the proposed Mamba's performances, Mamba is the best model for all three operator learning tasks. It outperforms the second-best model by a large margin.
Mamba2 is generally good and ranks the second best in the anti-derivative and nonlinear ODE operators. Even with a limited 10,000 parameter count budget, Mamba can already achieve a low error.
\item Mamba2 is slower and more costly than Mamba due to its more sophisticated structure; Mamba2 is mostly written by triton, a language and compiler for large-scale GPU parallel programming. Since the Mamba2 model used here is tiny, has a small hidden width and contains only one layer, it will be comparatively slow since triton's acceleration effect is not obvious.
In contrast, in the large language model (LLM) setting, if we construct multiple Mamba2 layers to design a large model containing billions of model parameters, the acceleration effect of triton will be more obvious.
\item Regarding other models, Oformer is the strongest baseline, which is a variant of the neural operator Transformer, demonstrating the effectiveness of the self-attention mechanism, but still worse and more costly than Mamba, signifying Mamba's potential to replace Transformers in operator learning tasks.
\item Regarding memory cost, Mamba is as memory efficient and fast as RNNs (specifically LSTM and GRU). Transformers are much more costly, so Mamba outperforms Transformers in accuracy while maintaining a low cost.
\item Compared with convolution-based models such as FNO and LNO, Mamba has a similar cost and outperforms them. LNO significantly outperforms FNO.
\item Mamba outperforms DeepONet since DeepONet cannot capture the time dependence as Mamba does.
\item We construct models with a similar number of parameters. Hence, their memory costs are similar. However, the total training time differs drastically due to different models' computational complexity. Specifically, the expensive self-attention mechanism makes Transformer-related models (Transformer, Oformer, and GNOT) the slowest among time series models, while RNNs and Mamba are relatively faster.
\item {\color{green!50!black}The training loss trajectory in the Pendulum problem is visualized in Figure \ref{fig:1D_DS_loss}. Mamba always maintains its advantage in faster convergence throughout the optimization. In later studies in Sections 
\ref{sec:extrapolation_long_time_1D_DS_deeponet} and \ref{sec:quantify_extrapolation_error_zhu2023reliable}, we adopt the same ODE data and Mamba model and report some relative error results, which is at the scale of $10^{-4} \sim 10^{-3}$ for in-distribution IID results. This demonstrates Mamba's outstanding representation ability under a limited parameter count budget, which does not require larger model sizes for low errors.}
\end{itemize}

\subsection{Finite Regularity and Discontinuous Solutions}\label{sec:finite_regularity}
Following the neural operator Transformer by Shih et al. \cite{shih2024transformers}, we aim to test various models and Mamba on problems with discontinuous solutions and finite regularity. Specifically, the following two ODEs are considered.
\begin{itemize}
\item \textbf{The Izhikevich Model} describing the spiking phenomenon in biological neurons is given by:
\begin{align}
u'(t) &= 0.04u^2 + 5u + 140 - v + I(t),\\
v'(t) &= a(bu - v).
\end{align}
The initial condition is determined by $u_{thres}$, the voltage threshold, and the condition $ u \geq u_{thres} $.
We aim to learn the mapping $I(t) \mapsto u(t)$, where $u(t)$ is the membrane recovery, $v(t)$ is the membrane potential, and $I(t)$ is the forcing spike function. We opt for $a = 0.02, b = 0.25, c = -55, d=0.05$ and $u_{thres} = -64$. 
We follow Shih et al. \cite{shih2024transformers} to generate 2,400 training samples and 150 testing samples. The sequence length is 401.

\item \textbf{The Tempered Fractional LIF Model} is another spiking neuron model given by:
\begin{align}
\tau \mathcal{D}^{\alpha,\sigma}_t v(t) = - (v - v_{rest}) + R\cdot I(t).
\end{align}
Here, $\tau$ is the membrane time constant, $R$ is the membrane resistance, $I(t)$ is the forcing spike function, $v_{rest}$ is the resting membrane potential, and $\mathcal{D}^{\alpha,\sigma}_t$ is the tempered fractional time-derivative with fractional order $\alpha$ and tempering coefficient $\sigma$.
We aim to learn the mapping $\left(I(t), \alpha\right) \mapsto v(t)$.
We construct the input by concatenating the fractional order $\alpha$ to each time step of the input sequence $I(t)$. Hence, the input sequence dimension is two. This example demonstrates how Mamba can incorporate ODE coefficients for flexible and general modeling.
We follow Shih et al. \cite{shih2024transformers} to generate 1,900 training samples and 480 testing samples. The sequence length is 200.
\end{itemize}
The training loss is MSE, and the test metric is the relative $L_2$ error. The test metric is similar to the one used by Shih et al. \cite{shih2024transformers} for a direct comparison.
We train all models for 10,001 epochs. 
In each epoch, we traverse the dataset with a minibatch size of 128.
We use the Adam optimizer \cite{kingma2014adam} with a 1e-3 initial learning rate, which decays linearly to zero at the end of training.
We keep the number of parameters of different models similar for a fair comparison. 
We run the code for five independent random seeds and report the average.

The model structure hyperparameter settings are summarized below.
{\color{green!50!black} During validation, we train the model on 80\% of the training set and validate on the rest 20\%. The model search space is the same as the previous Section \ref{sec:1D_DS_DeepONet}, and we find the following outperforming model structures.}
LSTM and GRU are one-layer models with 32 hidden dimensions.
DeepONet's trunk and branch nets are all five-layer networks with 24 hidden dimensions.
FNO is a four-layer model with 16 hidden dimensions and eight modes.
We choose a one-layer LNO with eight hidden dimensions and four modes.
We opt for a one-layer Transformer with 40 hidden and feedforward dimensions and four heads.
For all three Oformers with vanilla, Galerkin, or Fourier attention, we opt for one-layer Oformers with 24 hidden and feedforward dimensions and four heads.
GNOT has three layers, 16 hidden and feedforward dimensions, four heads, and two experts.
Regarding Mamba, we opt for a block with skip connection and layer normalization, with 16 hidden and intermediate dimensions.
{\color{green!50!black} Again, for all time series models, including RNNs, Mamba, and Transformers, we keep the same one-layer structure and similar parameter count budgets for a fair comparison.}

\begin{table}[]
\footnotesize
\centering
\begin{tabular}{|ccccc|ccccc|}
\hline
\multicolumn{5}{|c|}{The Izhikevich Model} & \multicolumn{5}{c|}{Tempered Fractional LIF Model} \\ \hline
\multicolumn{1}{|c|}{Model} & \multicolumn{1}{c|}{Params} & \multicolumn{1}{c|}{Memory} & \multicolumn{1}{c|}{Time} & Relative $L_2$ Error & \multicolumn{1}{c|}{Model} & \multicolumn{1}{c|}{Params} & \multicolumn{1}{c|}{Memory} & \multicolumn{1}{c|}{Time} & Relative $L_2$ Error \\ \hline
\multicolumn{1}{|c|}{LSTM} & \multicolumn{1}{c|}{8545} & \multicolumn{1}{c|}{1863MiB} & \multicolumn{1}{c|}{10min} & \textcolor{red}{1.225E-1$\pm$9.256E-3} & \multicolumn{1}{c|}{LSTM} & \multicolumn{1}{c|}{8577} & \multicolumn{1}{c|}{1707MiB} & \multicolumn{1}{c|}{7min} & \textcolor{blue}{2.181E-2$\pm$9.073E-5} \\ \hline
\multicolumn{1}{|c|}{GRU} & \multicolumn{1}{c|}{6433} & \multicolumn{1}{c|}{1821MiB} & \multicolumn{1}{c|}{10min} & \textbf{1.138E-1$\pm$8.422E-3} & \multicolumn{1}{c|}{GRU} & \multicolumn{1}{c|}{6465} & \multicolumn{1}{c|}{1703MiB} & \multicolumn{1}{c|}{7min} & \textbf{2.056E-2$\pm$1.556E-4} \\ \hline
\multicolumn{1}{|c|}{DeepONet} & \multicolumn{1}{c|}{14496} & \multicolumn{1}{c|}{1383MiB} & \multicolumn{1}{c|}{9min} & 2.150E-1$\pm$1.155E-4 & \multicolumn{1}{c|}{DeepONet} & \multicolumn{1}{c|}{14472} & \multicolumn{1}{c|}{1383MiB} & \multicolumn{1}{c|}{7min} & 2.782E-1$\pm$3.581E-3 \\ \hline
\multicolumn{1}{|c|}{FNO} & \multicolumn{1}{c|}{9329} & \multicolumn{1}{c|}{1859MiB} & \multicolumn{1}{c|}{17min} & 1.517E-1$\pm$2.508E-4 & \multicolumn{1}{c|}{FNO} & \multicolumn{1}{c|}{9345} & \multicolumn{1}{c|}{1839MiB} & \multicolumn{1}{c|}{15min} & 2.203E-2$\pm$3.385E-4 \\ \hline
\multicolumn{1}{|c|}{LNO} & \multicolumn{1}{c|}{1889} & \multicolumn{1}{c|}{1917MiB} & \multicolumn{1}{c|}{12min} & 1.786E-1$\pm$1.842E-2 & \multicolumn{1}{c|}{LNO} & \multicolumn{1}{c|}{1897} & \multicolumn{1}{c|}{1849MiB} & \multicolumn{1}{c|}{9min} & 2.635E-1$\pm$3.198E-2 \\ \hline
\multicolumn{1}{|c|}{Mamba} & \multicolumn{1}{c|}{9649} & \multicolumn{1}{c|}{1599MiB} & \multicolumn{1}{c|}{12min} & \textcolor{blue}{1.359E-1$\pm$1.421E-2} & \multicolumn{1}{c|}{Mamba} & \multicolumn{1}{c|}{9665} & \multicolumn{1}{c|}{1493MiB} & \multicolumn{1}{c|}{8min} & \textcolor{red}{2.132E-2$\pm$3.087E-4} \\ \hline
\multicolumn{1}{|c|}{Transformer} & \multicolumn{1}{c|}{10201} & \multicolumn{1}{c|}{2951MiB} & \multicolumn{1}{c|}{44min} & 2.274E-1$\pm$1.012E-2 & \multicolumn{1}{c|}{Transformer} & \multicolumn{1}{c|}{10241} & \multicolumn{1}{c|}{1735MiB} & \multicolumn{1}{c|}{13min} & 4.549E-2$\pm$2.509E-3 \\ \hline
\multicolumn{1}{|c|}{Oformer-V} & \multicolumn{1}{c|}{11209} & \multicolumn{1}{c|}{2743MiB} & \multicolumn{1}{c|}{39min} & 2.325E-1$\pm$2.189E-2 & \multicolumn{1}{c|}{Oformer-V} & \multicolumn{1}{c|}{11233} & \multicolumn{1}{c|}{1711MiB} & \multicolumn{1}{c|}{12min} & 2.395E-2$\pm$5.660E-4 \\ \hline
\multicolumn{1}{|c|}{Oformer-G} & \multicolumn{1}{c|}{11161} & \multicolumn{1}{c|}{1579MiB} & \multicolumn{1}{c|}{31min} & 2.243E-1$\pm$2.541E-2 & \multicolumn{1}{c|}{Oformer-G} & \multicolumn{1}{c|}{11185} & \multicolumn{1}{c|}{1443MiB} & \multicolumn{1}{c|}{14min} & 6.419E-2$\pm$9.055E-3 \\ \hline
\multicolumn{1}{|c|}{Oformer-F} & \multicolumn{1}{c|}{11161} & \multicolumn{1}{c|}{1579MiB} & \multicolumn{1}{c|}{31min} & 2.202E-1$\pm$2.491E-2 & \multicolumn{1}{c|}{Oformer-F} & \multicolumn{1}{c|}{11185} & \multicolumn{1}{c|}{1443MiB} & \multicolumn{1}{c|}{14min} & 6.623E-2$\pm$7.160E-3 \\ \hline
\multicolumn{1}{|c|}{GNOT} & \multicolumn{1}{c|}{11187} & \multicolumn{1}{c|}{1631MiB} & \multicolumn{1}{c|}{61min} & 1.782E-1$\pm$2.864E-2 & \multicolumn{1}{c|}{GNOT} & \multicolumn{1}{c|}{11203} & \multicolumn{1}{c|}{1445MiB} & \multicolumn{1}{c|}{27min} & 6.262E-2$\pm$1.207E-2 \\ \hline
\end{tabular}
\caption{Results for finite regularity and discontinuous solutions adopted from Shih et al. \cite{shih2024transformers} corresponding to Section \ref{sec:finite_regularity} in this paper. Visualization of the solutions is presented in Figure \ref{fig:finite_regularity}. Here, we present the test relative $L_2$ errors. \textbf{The best model is bold}, \textcolor{red}{the second best is red}, and \textcolor{blue}{the third best is blue}. Some loss trajectories are visualized in Figure \ref{fig:finite_reg_loss}.}
\label{tab:discontinuous_sol}
\end{table}

\subsubsection{Main Results}

The results are presented in Table \ref{tab:discontinuous_sol}, and the main observations are summarized as follows.
\begin{itemize}
\item Regarding the Izhikevich model, RNNs (GRU and LSTM) perform the best, while Mamba is the third best.
Regarding the LIF model, Mamba performs the second best.
Hence, Mamba can capture discontinuous dynamics.
\item Overall, due to the longer sequence length with 401 steps in the Izhikevich model and 200 steps in the LIF model, Transformers are more costly, even with linear cost attentions such as Oformer with Galerkin/Fourier attention and GNOT. In contrast, RNNs and Mamba maintain a lower cost and converge faster.
\item With better hyperparameter choices, we outperform the original results by Shih et al. \cite{shih2024transformers}. Specifically, Shih et al. achieve relative errors of 0.49 and 0.03 on the Izhikevich model and the LIF model, respectively. We report relative $L_2$ errors here for a clear comparison following Shih et al. \cite{shih2024transformers}.
\item Since the LIF inputs are both $I(t)$ and the fractional order $\alpha$, its input dimension is two. Hence, the input dimension of all models for learning the LIF operator is two, which is larger than the one-dimensional input in the Izhikevich model. Consequently, models in the LIF operator have slightly more parameters.
In addition, since the sequence in the Izhikevich model is 401, which is longer than that in the LIF model (200), models are trained for longer.
\item Although we chose a small network hidden size for DeepONet, the number of parameters of DeepONet is still large due to the large input dimension since conventional DeepONet cannot use the temporal information for parameter saving; hence, its performance is limited.
\item Regarding convolutional models (FNO, LNO), FNO is more stable, while LNO diverges in the LIF model.
\end{itemize}

\begin{figure}[htbp]
\centering
\includegraphics[width=0.32\linewidth]{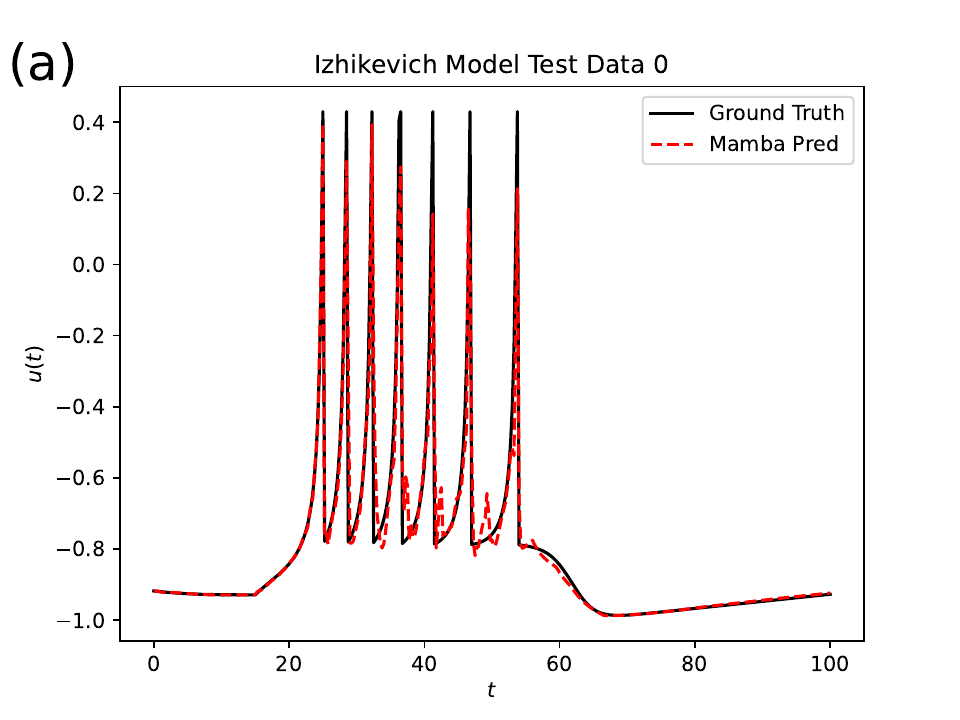}
\includegraphics[width=0.32\linewidth]{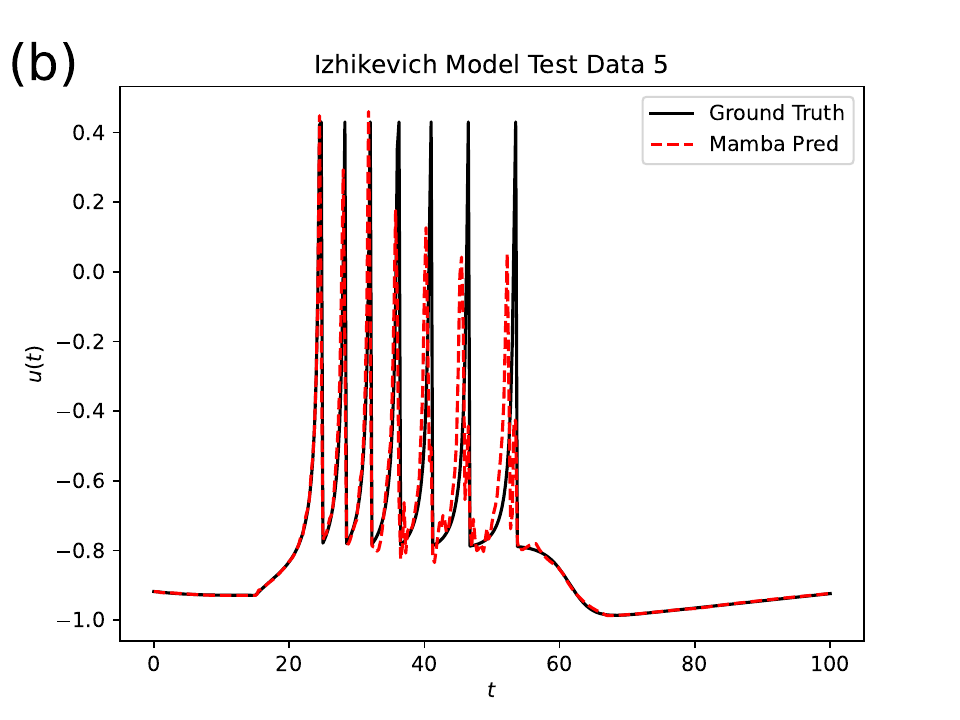}
\includegraphics[width=0.32\linewidth]{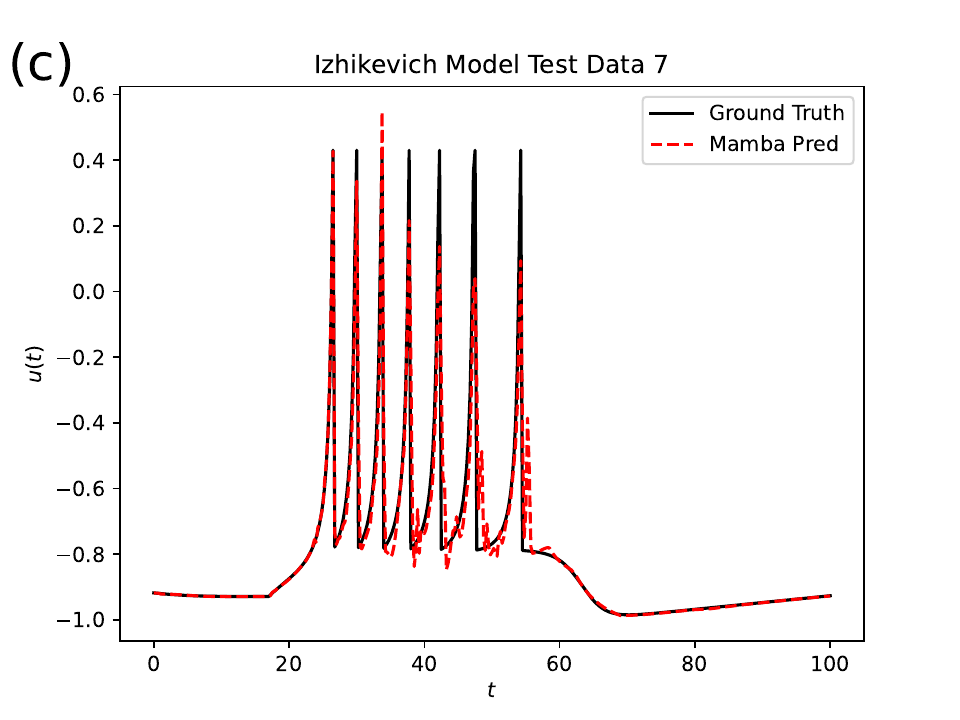}
\includegraphics[width=0.32\linewidth]{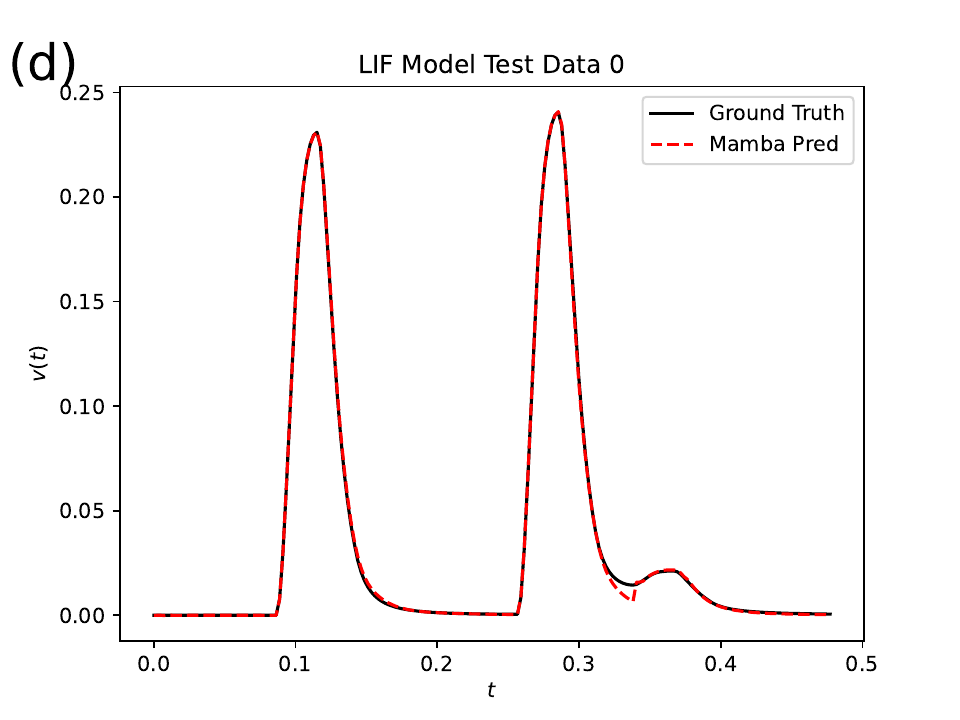}
\includegraphics[width=0.32\linewidth]{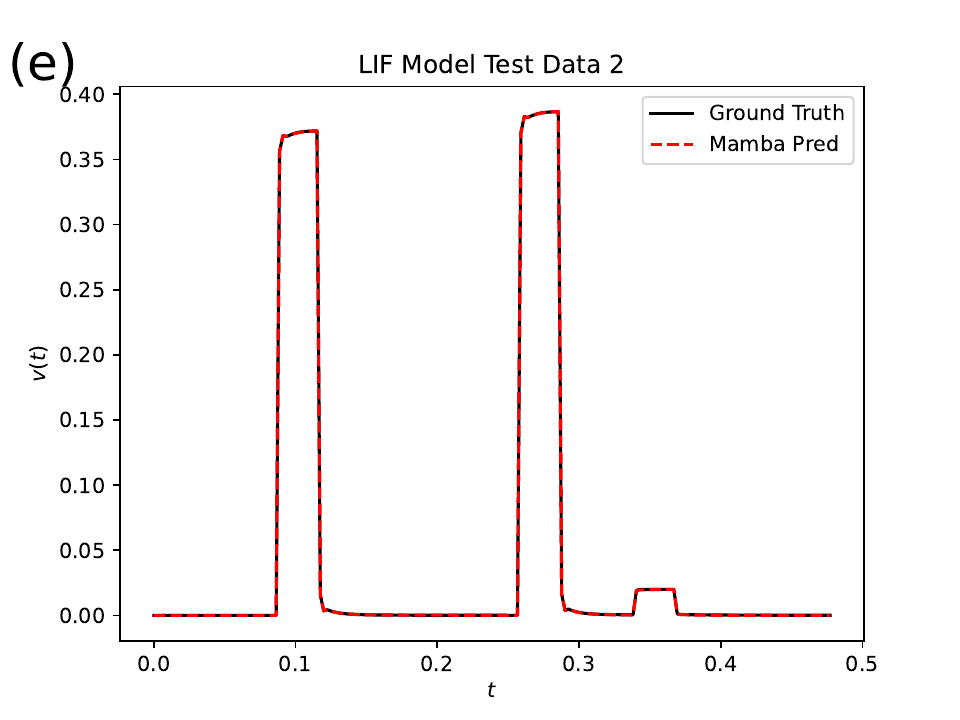}
\includegraphics[width=0.32\linewidth]{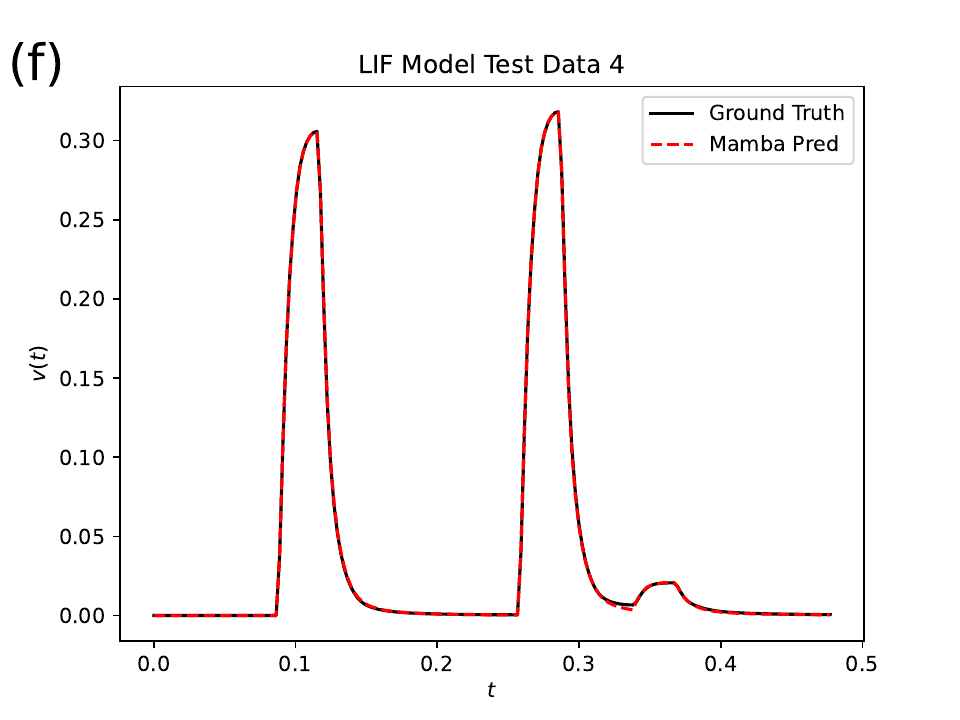}
\caption{Visualization of Mamba's prediction in finite regularity solutions adopted from Shih et al. \cite{shih2024transformers} corresponding to Section \ref{sec:finite_regularity} in this paper. The detailed quantitative results for this setting have been presented in Table \ref{tab:discontinuous_sol}. 
First row: subfigure (a) is the Izhikevich model with test data number 0, subfigure (b) is the Izhikevich model with test data number 5, and subfigure (c) is the Izhikevich model with test data number 7. 
Second row: subfigure (d) is the LIF model with test data number 0, subfigure (e) is the LIF model with test data number 2, and subfigure (f) is the LIF model with test data number 4.}
\label{fig:finite_regularity}
\end{figure}

We visualize some representative Mamba’s predictions in Figure \ref{fig:finite_regularity}.
The first row shows the Mamba prediction visualization on the Izhikevich model, where Mamba's relative $L_2$ error is 1.257E-1, while the second row shows that on the LIF model,  where Mamba's relative $L_2$ error is 2.126E-2. It is obvious that the first row's errors are larger than the second row's errors, which is consistent with our quantitative results. The visualization also suggests that the larger error in the Izhikevich may stem from its more stiff peaks compared with the LIF model.
Specifically, the discontinuous solution in the Izhikevich model tends to have six peaks, while the LIF model only has two to three peaks.
The visualization results further confirm Mamba's capability to capture rough solutions in ODE operators with finite regularity.

\begin{figure}[htbp]
\centering
\includegraphics[width=0.48\linewidth]{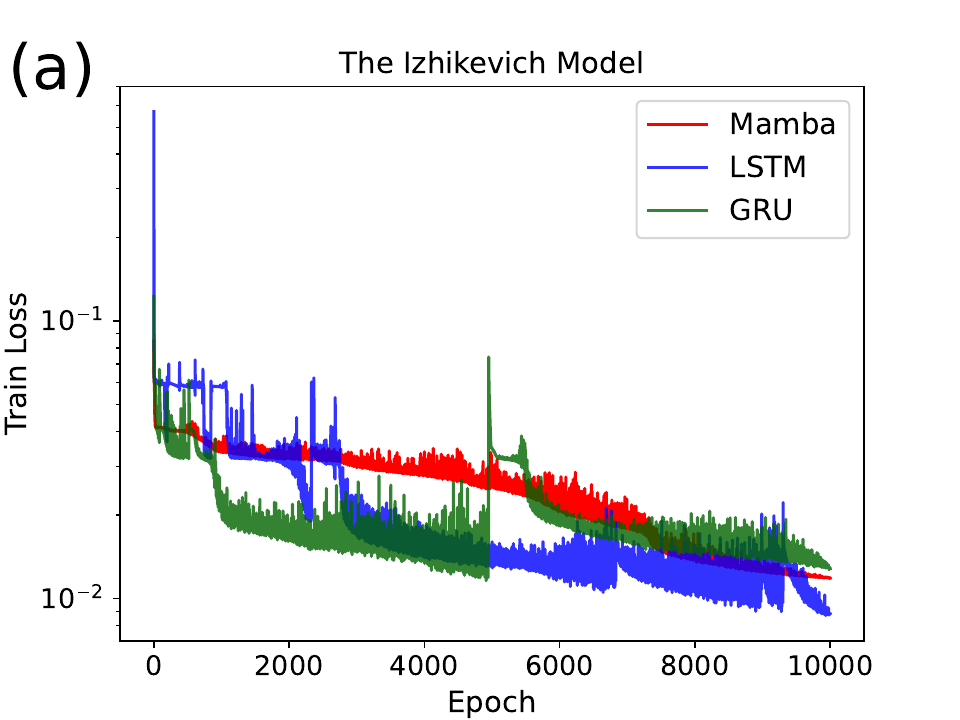}
\includegraphics[width=0.48\linewidth]{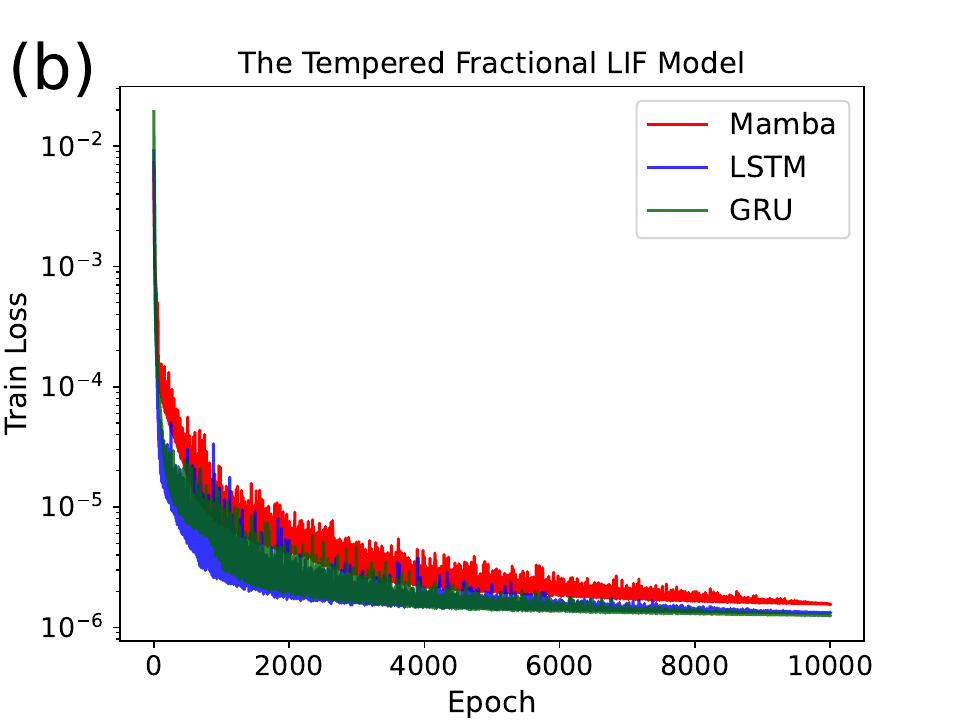}
\caption{{\color{green!50!black}Visualization of top three models' loss trajectories for operators with finite regularity following Shih et al. \cite{shih2024transformers} corresponding to Section \ref{sec:finite_regularity} in this paper. The detailed quantitative results for this setting are presented in Table \ref{tab:discontinuous_sol}. 
Subfigure (a) is the Izhikevich model; subfigure (b) is the LIF model.}}
\label{fig:finite_reg_loss}
\end{figure}

{\color{green!50!black} To further understand and compare the top three models, namely GRU, LSTM, and Mamba, we plot their respective optimization trajectories on the two problems in Figure \ref{fig:finite_reg_loss}.
Since the LIF model has two to three peaks and is easier to learn in terms of model accuracy, the three models' train loss decreases continuously throughout the optimization in (b).
However, due to the stronger discontinuity in the Izhikevich model in (a), LSTM and GRU exhibit unstable optimization with spikes in loss during training, while Mamba remains stable. Despite Mamba's slightly larger error, it demonstrates better training stability under solution discontinuity in operator learning.}

\subsubsection{Additional Study: Larger Model for Better Performances}\label{sec:large_model_izhikevich}
{\color{green!50!black}
Due to the discontinuous solutions, the relative $L_2$ test error results for the Izhikevich model are larger than $10^{-1}$, which is not satisfactory. Hence, we will conduct an additional study to achieve better performance with larger model sizes and further understand different models' accuracy under larger parameter counts. We selected the top three models for this study: LSTM, GRU, and Mamba, with various depths and hidden dimensions. We keep other settings the same except for controlling the model size.
The additional results are shown in Table \ref{tab:large_model_izhikevich}. The lowest error is achieved by Mamba with three layers and 16 hidden dimensions, demonstrating Mamba's advantage under various model sizes.

\begin{table}[htbp]
\centering
\begin{tabular}{|ccccccc|}
\hline
\multicolumn{7}{|c|}{Additional Study on the Izhikevich Model} \\ \hline
\multicolumn{1}{|c|}{Model} & \multicolumn{1}{c|}{Depth} & \multicolumn{1}{c|}{Width} & \multicolumn{1}{c|}{Params} & \multicolumn{1}{c|}{Memory} & \multicolumn{1}{c|}{Time} & Relative $L_2$ Error \\ \hline
\multicolumn{1}{|c|}{\multirow{4}{*}{LSTM}} & \multicolumn{1}{c|}{1} & \multicolumn{1}{c|}{32} & \multicolumn{1}{c|}{8545} & \multicolumn{1}{c|}{1863MiB} & \multicolumn{1}{c|}{10min} & 1.225E-1$\pm$9.256E-3 \\ \cline{2-7} 
\multicolumn{1}{|c|}{} & \multicolumn{1}{c|}{2} & \multicolumn{1}{c|}{32} & \multicolumn{1}{c|}{16993} & \multicolumn{1}{c|}{1885MiB} & \multicolumn{1}{c|}{13min} & 1.240E-1$\pm$2.110E-2 \\ \cline{2-7} 
\multicolumn{1}{|c|}{} & \multicolumn{1}{c|}{3} & \multicolumn{1}{c|}{32} & \multicolumn{1}{c|}{25441} & \multicolumn{1}{c|}{1923MiB} & \multicolumn{1}{c|}{15min} & 1.218E-1$\pm$6.069E-3 \\ \cline{2-7} 
\multicolumn{1}{|c|}{} & \multicolumn{1}{c|}{1} & \multicolumn{1}{c|}{64} & \multicolumn{1}{c|}{33473} & \multicolumn{1}{c|}{2175MiB} & \multicolumn{1}{c|}{11min} & 1.216E-1$\pm$1.190E-2 \\ \hline
\multicolumn{1}{|c|}{\multirow{4}{*}{GRU}} & \multicolumn{1}{c|}{1} & \multicolumn{1}{c|}{32} & \multicolumn{1}{c|}{6433} & \multicolumn{1}{c|}{1821MiB} & \multicolumn{1}{c|}{10min} & 1.138E-1$\pm$8.422E-3 \\ \cline{2-7} 
\multicolumn{1}{|c|}{} & \multicolumn{1}{c|}{2} & \multicolumn{1}{c|}{32} & \multicolumn{1}{c|}{12769} & \multicolumn{1}{c|}{1861MiB} & \multicolumn{1}{c|}{14min} & 1.111E-1$\pm$5.581E-3 \\ \cline{2-7} 
\multicolumn{1}{|c|}{} & \multicolumn{1}{c|}{3} & \multicolumn{1}{c|}{32} & \multicolumn{1}{c|}{19105} & \multicolumn{1}{c|}{1905MiB} & \multicolumn{1}{c|}{16min} & 1.193E-1$\pm$1.792E-2 \\ \cline{2-7} 
\multicolumn{1}{|c|}{} & \multicolumn{1}{c|}{1} & \multicolumn{1}{c|}{64} & \multicolumn{1}{c|}{25153} & \multicolumn{1}{c|}{2087MiB} & \multicolumn{1}{c|}{12min} & 1.249E-1$\pm$1.497E-2 \\ \hline
\multicolumn{1}{|c|}{\multirow{4}{*}{Mamba}} & \multicolumn{1}{c|}{1} & \multicolumn{1}{c|}{16} & \multicolumn{1}{c|}{9649} & \multicolumn{1}{c|}{1599MiB} & \multicolumn{1}{c|}{12min} & 1.359E-1$\pm$1.421E-2 \\ \cline{2-7} 
\multicolumn{1}{|c|}{} & \multicolumn{1}{c|}{2} & \multicolumn{1}{c|}{16} & \multicolumn{1}{c|}{19217} & \multicolumn{1}{c|}{1759MiB} & \multicolumn{1}{c|}{18min} & 1.175E-1$\pm$6.209E-3 \\ \cline{2-7} 
\multicolumn{1}{|c|}{} & \multicolumn{1}{c|}{3} & \multicolumn{1}{c|}{16} & \multicolumn{1}{c|}{28785} & \multicolumn{1}{c|}{1919MiB} & \multicolumn{1}{c|}{28min} & \textbf{1.045E-1$\pm$2.698E-3} \\ \cline{2-7} 
\multicolumn{1}{|c|}{} & \multicolumn{1}{c|}{1} & \multicolumn{1}{c|}{32} & \multicolumn{1}{c|}{22497} & \multicolumn{1}{c|}{1655MiB} & \multicolumn{1}{c|}{14min} & 1.305E-1$\pm$5.578E-3 \\ \hline
\end{tabular}
\caption{Results for larger models on the Izhikevich model corresponding to Section \ref{sec:large_model_izhikevich}.}
\label{tab:large_model_izhikevich}
\end{table}
}

\subsection{Out-of-Distribution Operator Learning Tasks}\label{sec:lnoode}

Following the work on Laplace Neural Operator (LNO) by Cao et al. \cite{cao2023lno}, we consider several nonlinear ODEs with external force.
\begin{itemize}
\item The chaotic Lorenz system with external force $f(t)$:
\begin{align}
\Dot{x} &= \sigma (y - x).\\
\Dot{y} &= x(\rho - z) - y.\\
\Dot{z} &= xy - \beta z - f(t).
\end{align}
We aim to learn the operator from the external force $f(t)$ to the state variable $x(t)$. The initial condition is $x(0) = 1, y(0) = 0, z(0) = 0$.
Here, $\sigma = 10$ and $\beta = 8/3$, and Cao et al. designed two settings with a Rayleigh number $\rho = 5$ and Rayleigh number $\rho = 10$.
Furthermore, Cao et al. demonstrated that $\rho = 5$ is relatively easier while $\rho = 10$ can only be solved via LNO, but other models like RNN fail to capture high-frequency components in the ODE solution. We name these two settings \textbf{Lorenz-5} and \textbf{Lorenz-10}, respectively.
\item Duffing oscillator:
\begin{align}
m\Ddot{x} + c\Dot{x} + k_1x + k_3x^3 = f(t), \quad x(0) = 0, \quad \Dot{x}(0) = 0.
\end{align}
We aim to learn the operator $f(t) \mapsto x(t)$; $c$ is the damping. With $c=0$, the solution will exhibit high-frequency components, which are hard to capture.
We follow the two settings in Cao et al. \cite{cao2023lno} with the common $m = 1, k_1 = 1, k_3 = 1$ and manipulate $c = 0$ or $c = 0.5$, which are dubbed \textbf{Duffing-0} and \textbf{Duffing-.5}. 
\item Driven gravity pendulum:
\begin{align}
\Ddot{x} + c\Dot{x} + \frac{g}{l} \sin(x) = f (t), \quad x(0) = 0, \quad \Dot{x}(0) = 0.
\end{align}
We aim to learn the operator $f(t) \mapsto x(t)$; $c$ is the damping coefficient. With $c=0$, the solution will exhibit high-frequency components which are hard to fit. We follow the two settings in Cao et al. \cite{cao2023lno} with the common $g / l = 1$ and manipulate $c = 0$ or $c = 0.5$, which are dubbed \textbf{Pend-0} and \textbf{Pend-.5}. 
\end{itemize}

For all three problems, we test the out-of-distribution (OOD) generalization problem following Cao et al. \cite{cao2023lno}, i.e., the training and testing functions/data will follow different distributions. 
Specifically, we sample 200 training samples from $f_{train}(t) =
A\sin(5t)$, where the $A \in \{0.05, 0.1, 0.15, \cdots, 9.95, 10\}$. We sample 50 validation data and 130 testing samples from $f_{test}(t) = A\exp(-0.05t)\sin(5t)$, where $A \in \{0.14, 0.19, \cdots, 9.04, 9.09\}$, i.e., the smallest 50 elements in $A$ forms the validation set and the largest 130 elements in $A$ forms the test set. There are 2048 temporal time steps with the discretization step size 0.01. The best model architectures are chosen based on 50 validation data with the same distribution as the test data.

Here are more implementation details. The training loss is MSE, and the test metric is the relative $L_2$ error.
We train all models for 10,001 epochs, while for LNO (Small), we follow the original setting in LNO \cite{cao2023lno}. 
In each epoch, we traverse the dataset with a minibatch size of 16.
We use the Adam optimizer \cite{kingma2014adam} with a 1e-3 initial learning rate, which decays linearly to zero at the end of training.
We keep the parameter numbers of different models similar for a fair comparison. 

{\color{green!50!black}We select model structures based on the reported validation dataset. We set a parameter count budget of approximately 100,000. The model search space is as follows.
For time series models, including GRU, LSTM, Mamba, Transformer, Oformer, and GNOT, the search space is model depth in $\{1,2,3,4\}$, hidden dimensions in $\{8,12,16,24,32,48,64,96,128\}$. We set all Transformer-related models' head numbers as four and the number of experts in GNOT as two.
We set DeepONet's trunk and branch nets as MLP and search the MLP depth in $\{2,3,4,5\}$, and a corresponding hidden size satisfies the parameter budget.
We search FNO's depth in $\{2,3,4\}$, hidden dimensions in $\{8, 16, 32\}$ and number of modes in $\{4, 8, 16\}$.
We search LNO's depth in $\{1,2,3\}$, hidden dimensions in $\{4, 8, 16\}$ and number of modes in $\{4, 8, 16\}$.
After validation, we choose the following model structures that perform the best among all the six validation sets on average.}
LSTM and GRU are one-layer models with 64 hidden dimensions.
DeepONet's trunk and branch nets are all five-layer networks with 24 hidden dimensions.
FNO is a three-layer model with 32 hidden dimensions and 8 modes.
Regarding LNO, we opt for two versions. The first is the original model from Cao et al. \cite{cao2023lno}, whose model size is smaller (LNO-small), and the other is a larger one-layer LNO with 16 hidden dimensions and 8 modes (LNO-large).
We opt for a four-layer Transformer with 24 hidden dimensions, 96 feedforward dimensions, and 4 heads.
For all three Oformers with vanilla, Galerkin, or Fourier attention, we opt for three-layer Oformers with 24 hidden dimensions, 24 feedforward dimensions, and 4 heads.
GNOT has three layers, 16 hidden dimensions and 16 feedforward dimensions, 4 heads, and 2 experts.
Regarding Mamba, we opt for two blocks with skip connection and layer normalization, with 16 model dimensions and 16 intermediate dimensions.
%
%Regarding Mamba, we opt for two hyperparameter settings via validation on 50 data following Cao et al. \cite{cao2023lno}. First, we opt for two blocks with skip connection and layer normalization, with 16 model dimensions and 16 intermediate dimensions for Lorenz $\rho=5$, Duffing $c=0$, and Pendulum $c=0.5$. For the other three test cases, we opt for six blocks with skip connection and layer normalization, with 8 model dimensions and 8 intermediate dimensions.

\begin{table}[htbp]
\centering
\small
\begin{tabular}{|c|c|c|c|c|c|c|}
\hline
Model & Params & Memory & Time & Lorenz-5 & Lorenz-10 & Duffing-0 \\ \hline
LSTM & 33537 & 1965MiB & 16min & \textcolor{red}{6.381E-3$\pm$7.821E-4} & 5.163E-1$\pm$1.029E-1 & 1.000E+0$\pm$6.831E-2 \\ \hline
GRU & 25217 & 1909MiB & 16min & \textcolor{blue}{1.276E-2$\pm$4.674E-3} & 4.789E-1$\pm$1.313E-1 & 8.089E-1$\pm$6.008E-2 \\ \hline
DeepONet & 103176 & 1385MiB & 5min & 1.216E-1$\pm$2.392E-2 & 7.933E-1$\pm$2.554E-1 & 9.136E-1$\pm$8.952E-2 \\ \hline
FNO & 27873 & 1693MiB & 8min & 2.662E-2$\pm$1.930E-3 & 3.769E-1$\pm$1.002E-1 & 5.895E-1$\pm$6.538E-2 \\ \hline
LNO (Small) & 1309 & 1885MiB & 1min & 1.748E-2$\pm$3.571E-3 & \textbf{1.420E-1$\pm$2.796E-2} & \textbf{2.598E-1$\pm$3.298E-2} \\ \hline
LNO (Large) & 6737 & 1987MiB & 7min & 1.982E-1$\pm$5.609E-2 & 8.006E-1$\pm$2.635E-1 & 9.988E-1$\pm$9.895E-2 \\ \hline
Transformer & 29041 & 20849MiB & 232min & 1.434E-2$\pm$4.190E-3 & 2.877E-1$\pm$1.133E-1 & 1.241E+0$\pm$1.780E-1 \\ \hline
Oformer-V & 33505 & 9607MiB & 162min & 4.264E-2$\pm$2.508E-3 & \textcolor{red}{2.634E-1$\pm$3.451E-2} & \textcolor{blue}{5.110E-1$\pm$2.189E-1} \\ \hline
Oformer-G & 33361 & 1949MiB & 37min & 3.131E-2$\pm$2.412E-3 & 8.388E-1$\pm$8.587E-2 & 1.049E+0$\pm$6.729E-2 \\ \hline
Oformer-F & 33361 & 1949MiB & 37min & 2.704E-2$\pm$1.722E-3 & 7.216E-1$\pm$9.066E-2 & 9.896E-1$\pm$1.153E-1 \\ \hline
GNOT & 33479 & 2333MiB & 78min & 3.259E-2$\pm$1.735E-3 & 8.738E-1$\pm$9.707E-3 & 1.122E+0$\pm$4.003E-2 \\ \hline
Mamba (Ours) & 19233 & 1635MiB & 10min & \textbf{5.956E-3$\pm$1.493E-3} & \textcolor{blue}{2.754E-1$\pm$2.045E-1} & \textcolor{red}{4.445E-1$\pm$2.297E-1} \\ \hline\hline
Model & Params & Memory & Time & Duffing-.5 & Pend-0 & Pend-.5 \\ \hline
LSTM & 33537 & 1965MiB & 16min & 1.283E-1$\pm$7.865E-3 & 7.589E-1$\pm$1.036E-1 & 1.352E-1$\pm$3.432E-3 \\ \hline
GRU & 25217 & 1909MiB & 16min & 1.282E-1$\pm$5.121E-3 & \textcolor{blue}{7.065E-1$\pm$2.152E-1} & \textcolor{blue}{1.159E-1$\pm$5.147E-3} \\ \hline
DeepONet & 103176 & 1385MiB & 5min & 5.887E-1$\pm$1.203E-1 & 1.145E+0$\pm$3.175E-2 & 5.773E-1$\pm$9.989E-2 \\ \hline
FNO & 27873 & 1693MiB & 8min & 1.938E-1$\pm$1.727E-2 & \textcolor{red}{6.835E-1$\pm$1.921E-1} & 1.544E-1$\pm$2.423E-2 \\ \hline
LNO (Small) & 1309 & 1885MiB & 1min & \textcolor{blue}{1.125E-1$\pm$2.098E-2} & \textbf{2.278E-1$\pm$4.556E-2} & 1.624E-1$\pm$2.074E-2 \\ \hline
LNO (Large) & 6737 & 1987MiB & 7min & 9.805E-1$\pm$2.546E-1 & 1.200E+0$\pm$2.873E-1 & 9.309E-1$\pm$2.357E-1 \\ \hline
Transformer & 29041 & 20849MiB & 232min & \textcolor{red}{1.092E-1$\pm$1.777E-2} & 7.754E-1$\pm$2.416E-1 & \textbf{8.497E-2$\pm$2.770E-2} \\ \hline
Oformer-V & 33505 & 9607MiB & 162min & 1.996E-1$\pm$1.725E-2 & 8.701E-1$\pm$3.055E-1 & 2.888E-1$\pm$2.850E-2 \\ \hline
Oformer-G & 33361 & 1949MiB & 37min & 2.948E-1$\pm$2.902E-2 & 9.137E-1$\pm$2.995E-2 & 2.317E-1$\pm$3.961E-2 \\ \hline
Oformer-F & 33361 & 1949MiB & 37min & 2.566E-1$\pm$4.633E-2 & 9.993E-1$\pm$4.672E-2 & 2.926E-1$\pm$1.445E-2 \\ \hline
GNOT & 33479 & 2333MiB & 78min & 2.856E-1$\pm$3.886E-2 & 9.392E-1$\pm$1.594E-2 & 3.285E-1$\pm$1.992E-2 \\ \hline
Mamba (Ours) & 19233 & 1635MiB & 10min & \textbf{8.957E-2$\pm$1.366E-2} & 8.978E-1$\pm$3.487E-2 & \textcolor{red}{8.837E-2$\pm$8.949E-3} \\ \hline
\end{tabular}
\caption{Results for the six test cases in the LNO paper \cite{cao2023lno} corresponding to Section \ref{sec:lnoode} in this paper. 
Some of Mamba's representative predictions are visualized in Figure \ref{fig:lno_prediction_visualization}.
Mamba's relative $L_2$ error with respect to time is plotted in Figure \ref{fig:lno_rel_error_t}.
This table presents the relative $L_2$ test errors.
\textbf{The best model is bold}, \textcolor{red}{the second best is red}, and \textcolor{blue}{the third best is blue}.}
\label{tab:Lorenz}
\end{table}

\begin{figure}[htbp]
\centering
\includegraphics[width=0.32\linewidth]{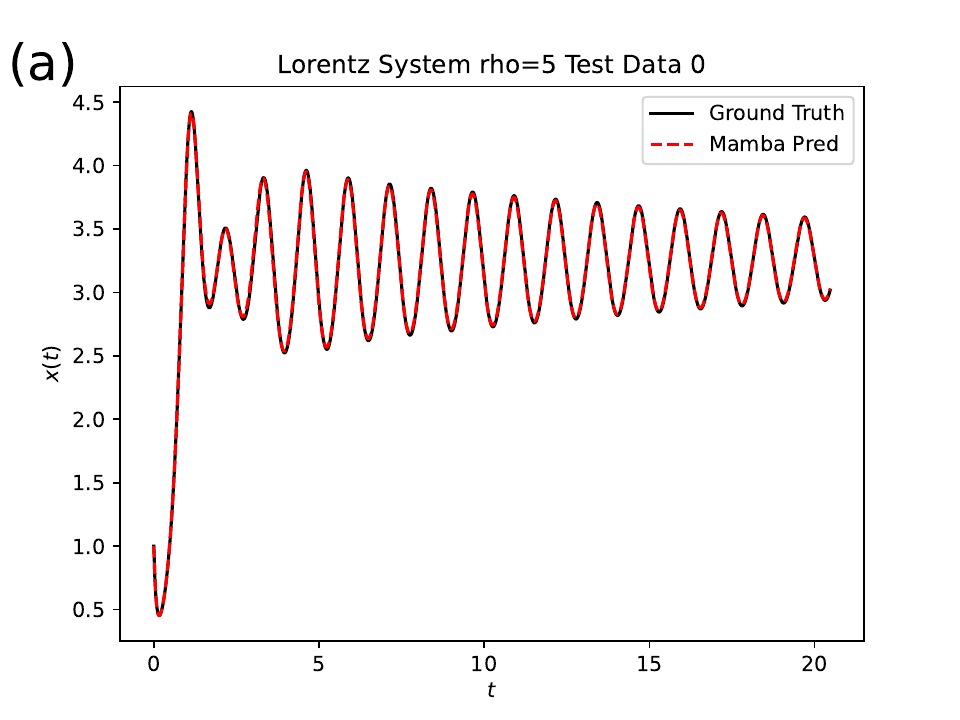}
\includegraphics[width=0.32\linewidth]{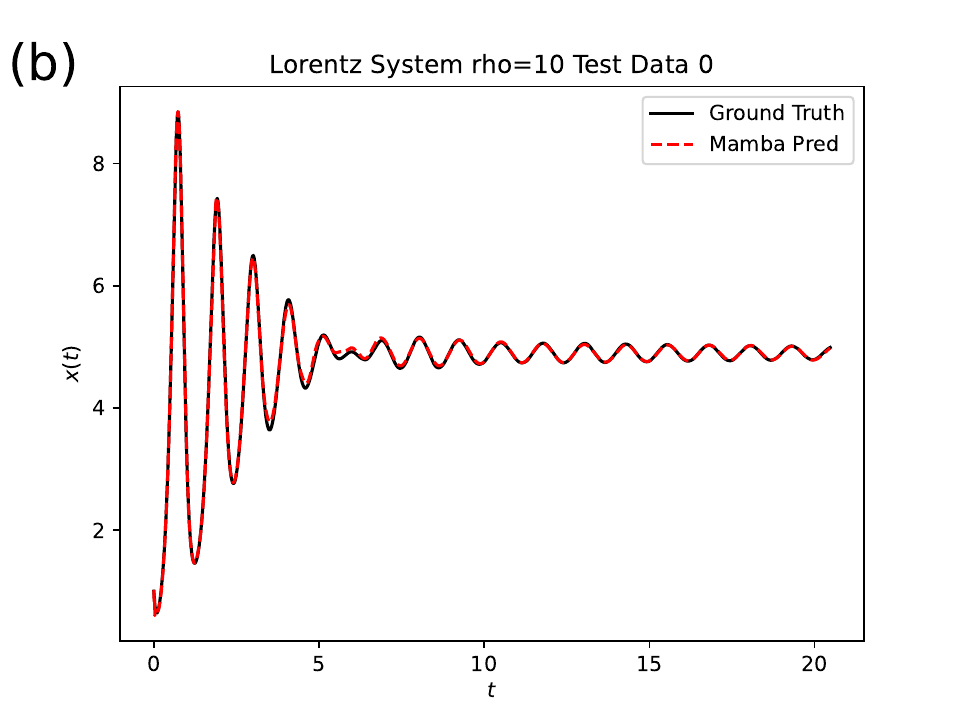}
\includegraphics[width=0.32\linewidth]{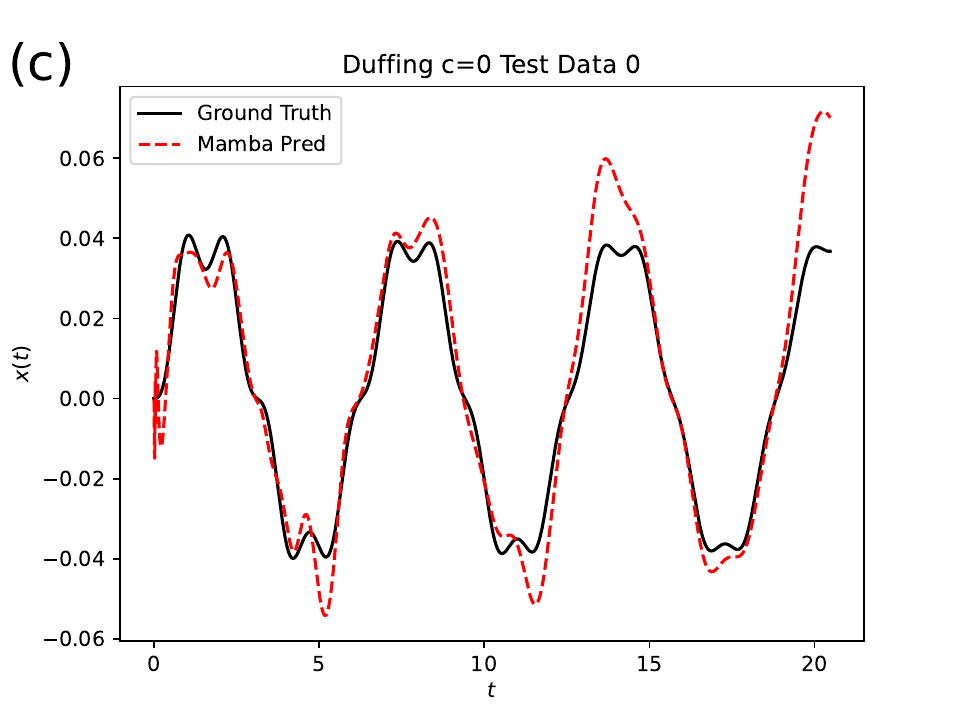}
\includegraphics[width=0.32\linewidth]{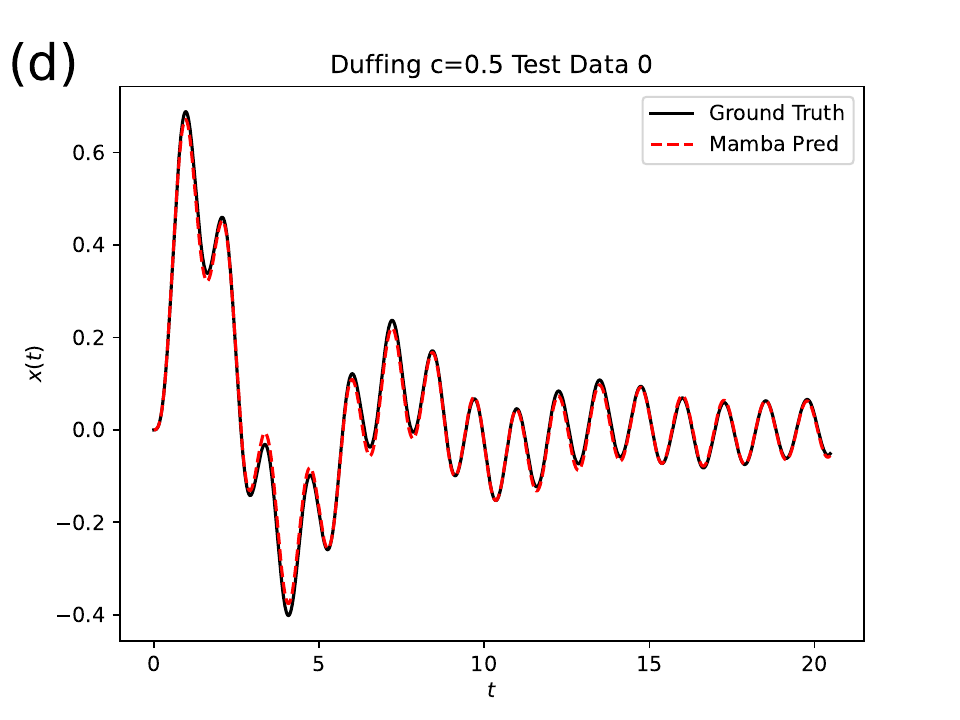}
\includegraphics[width=0.32\linewidth]{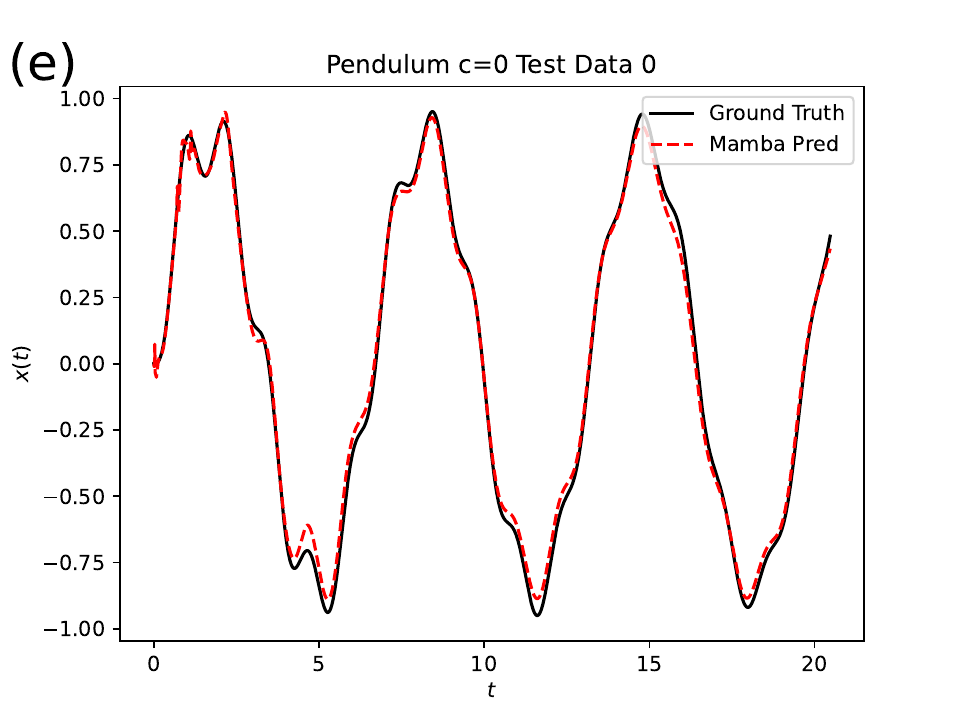}
\includegraphics[width=0.32\linewidth]{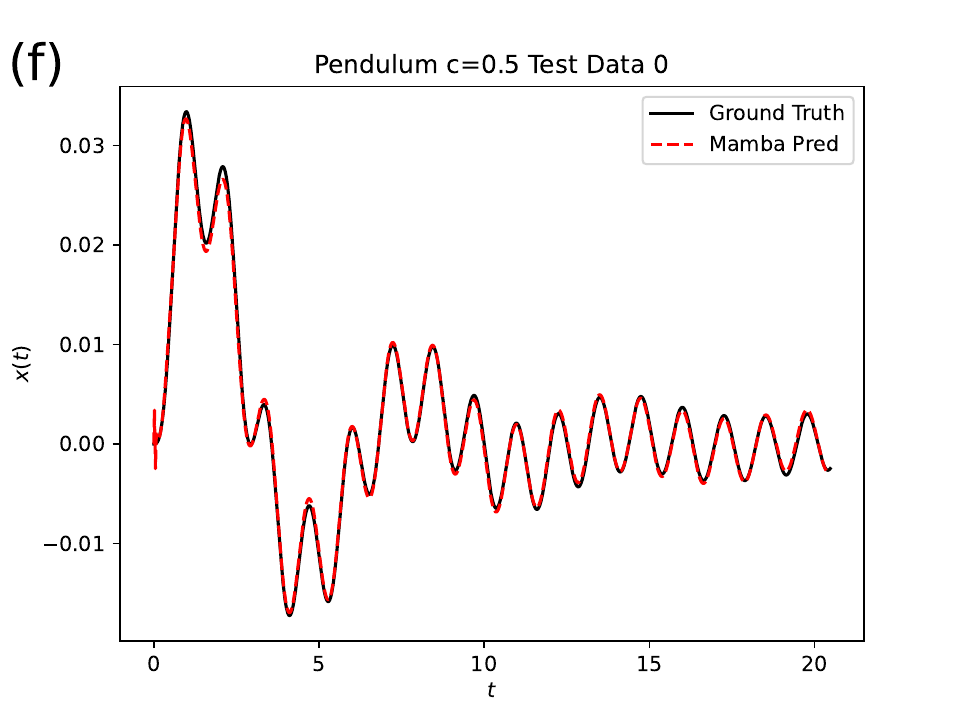}
\caption{Visualization of Mamba's prediction on one test data point in the six test cases following LNO \cite{cao2023lno} corresponding to Section \ref{sec:lnoode} in this paper. 
The quantitative results are presented in Table \ref{tab:Lorenz}.
Mamba's relative $L_2$ error with respect to time is plotted in Figure \ref{fig:lno_rel_error_t}.
Subfigure (a): Lorenz system with $\rho = 5$.
Subfigure (b): Lorenz system with $\rho = 10$.
Subfigure (c): Duffing oscillator with $c = 0$.
Subfigure (d): Duffing oscillator with $c = 0.5$.
Subfigure (e): Pendulum with $c = 0$.
Subfigure (f): Pendulum with $c = 0.5$.}
\label{fig:lno_prediction_visualization}
\end{figure}

\begin{figure}[htbp]
\centering
\includegraphics[width=0.32\linewidth]{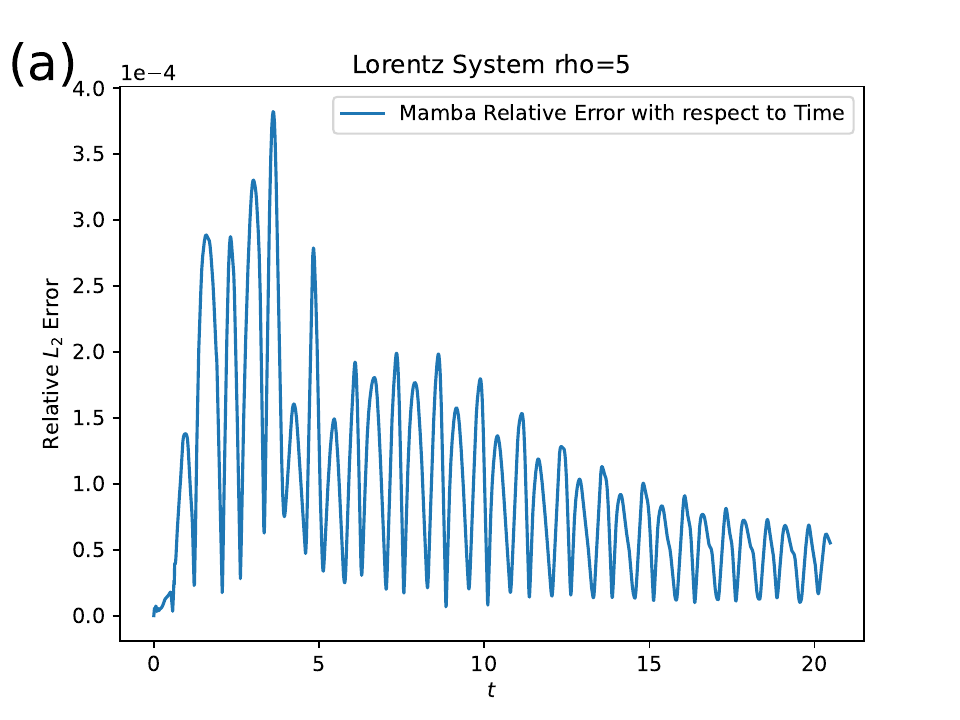}
\includegraphics[width=0.32\linewidth]{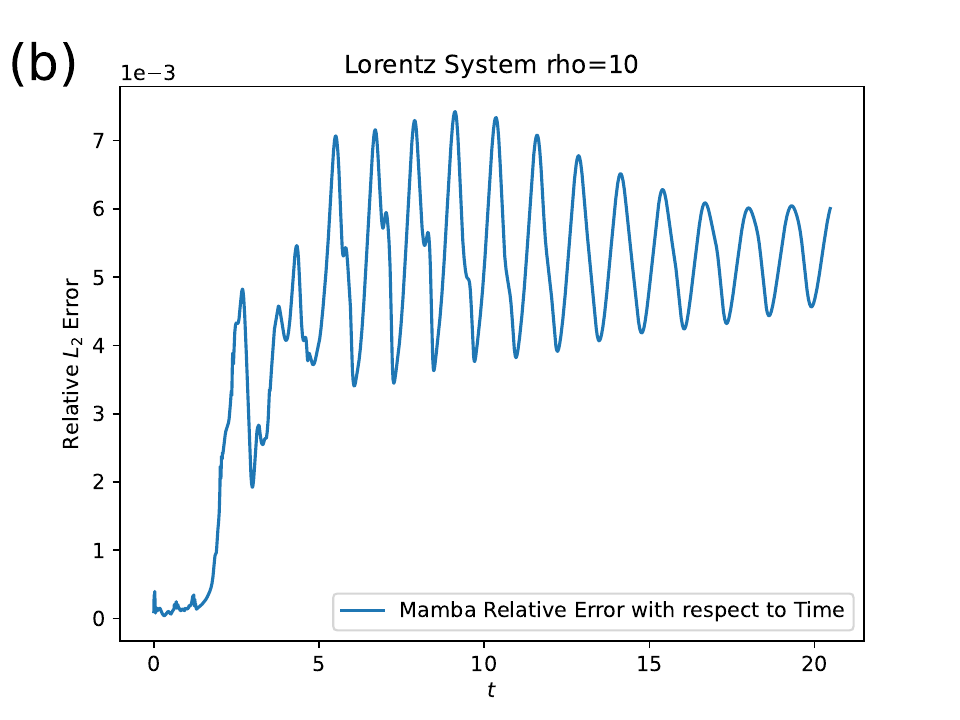}
\includegraphics[width=0.32\linewidth]{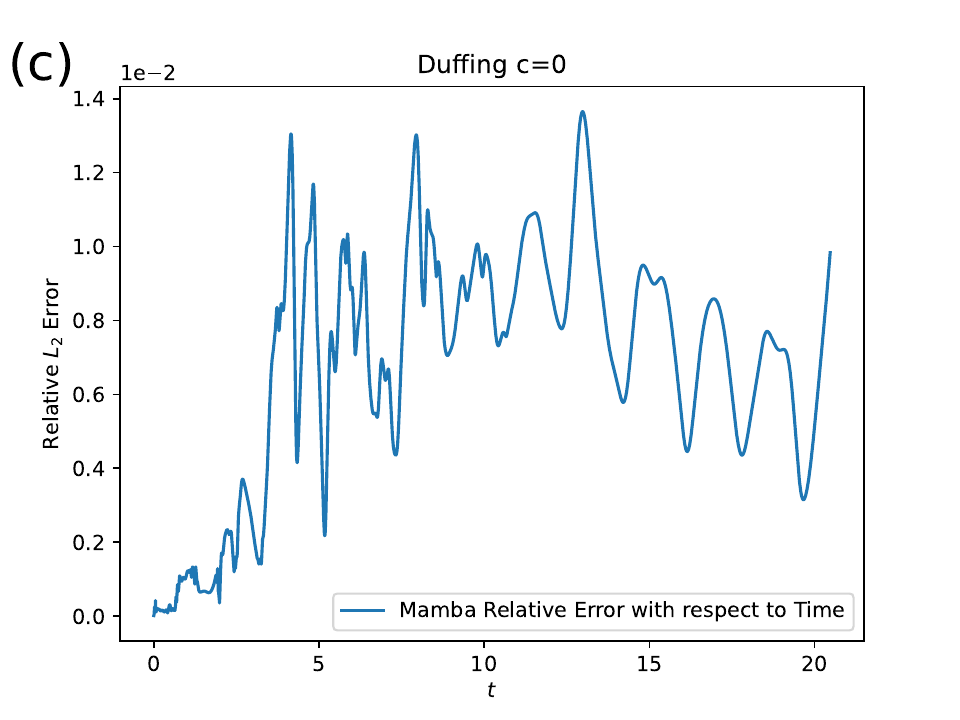}
\includegraphics[width=0.32\linewidth]{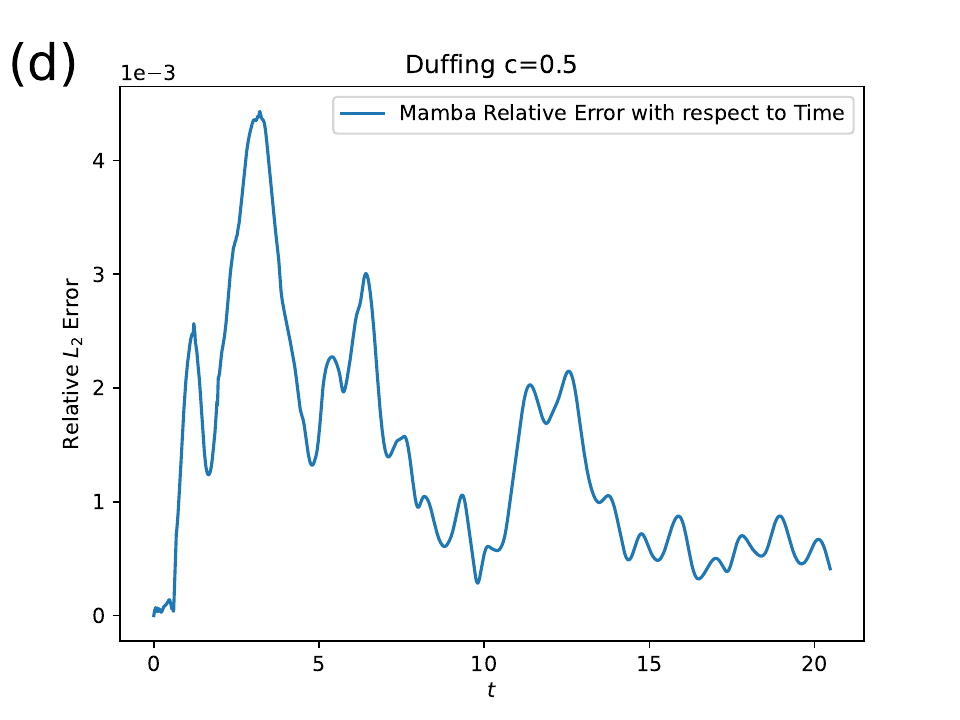}
\includegraphics[width=0.32\linewidth]{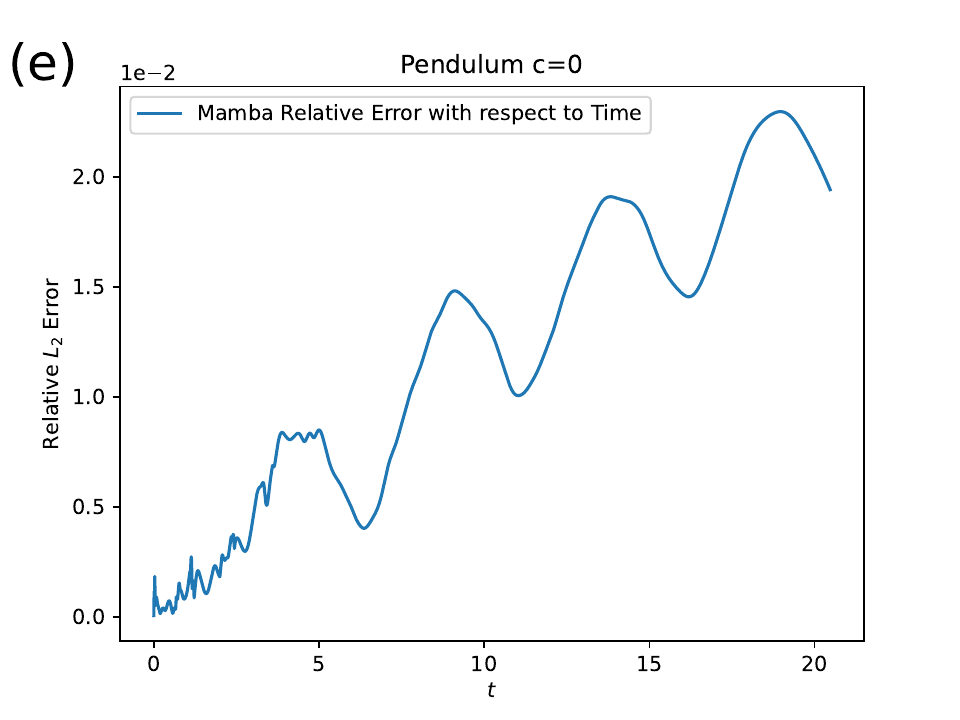}
\includegraphics[width=0.32\linewidth]{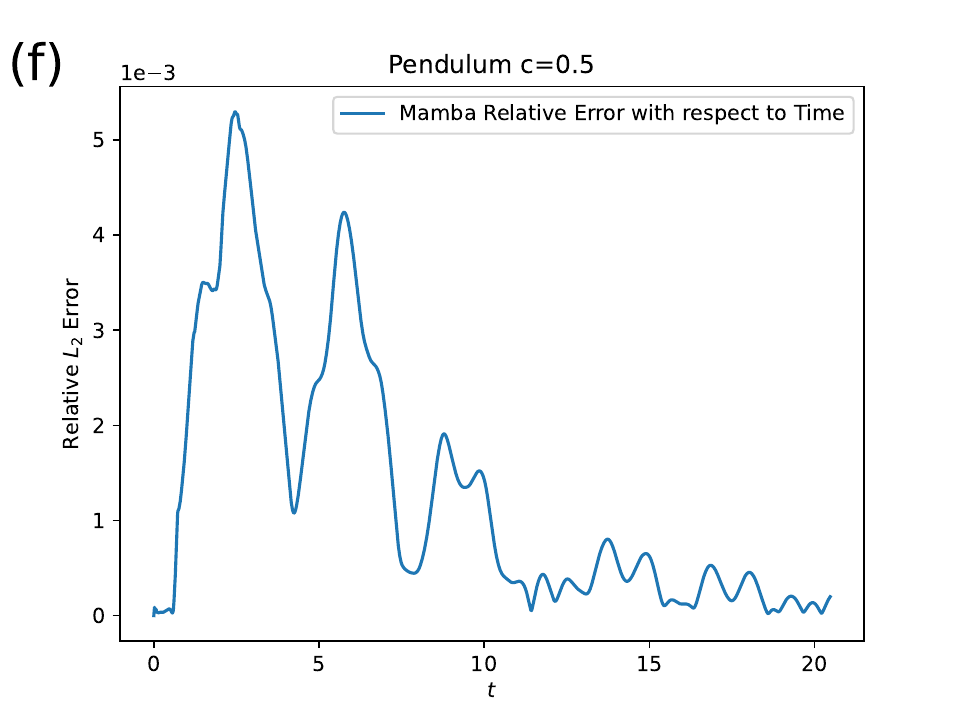}
\caption{Visualization of Mamba's relative $L_2$ prediction error concerning time in the six LNO test cases \cite{cao2023lno} corresponding to Section \ref{sec:lnoode} in this paper.
The quantitative results are presented in Table \ref{tab:Lorenz}.
Some of Mamba's representative predictions are visualized in Figure \ref{fig:lno_prediction_visualization}.
Subfigure (a): Lorenz system with $\rho = 5$.
Subfigure (b): Lorenz system with $\rho = 10$.
Subfigure (c): Duffing oscillator with $c = 0$.
Subfigure (d): Duffing oscillator with $c = 0.5$.
Subfigure (e): Pendulum with $c = 0$.
Subfigure (f): Pendulum with $c = 0.5$.}
\label{fig:lno_rel_error_t}
\end{figure}

We present the results in Table \ref{tab:Lorenz}.
We visualize Mamba's prediction on test data number 0 in Figure \ref{fig:lno_prediction_visualization}.
The main observations are summarized as follows.
Among all the six test cases, Mamba ranks in the top three in five, demonstrating superior stability across various settings.
Specifically, these six cases can be divided into two distinct groups: (1) Lorenz-5, Duffing-.5, and Pend-.5, where all models can achieve relatively low error; (2) Lorenz-10, Duffing-0, and Pend-0, where LNO performs better in test cases without the damping factor or cases with chaotic solutions exhibiting high-frequency solutions.
In the first group, although all models are capable of achieving relatively low errors, Mamba performs better.
In the second group, LNO performs the best since it can learn the transient responses in undamped or chaotic test cases with exponential convergence. Nevertheless, Mamba still performs in the top three in two of the three difficult cases. Other models mainly diverge and get a 100\% relative error, which demonstrates Mamba's ability to capture the transient response in some cases.
Though LNO is good at capturing transient responses and outperforms Mamba in test cases from the second group, it is not necessarily good in normal test cases. Other baselines, such as RNNs, DeepONet, FNO, and Transformers, are less stable than Mamba across various settings. 

To better understand Mamba's prediction error more deeply, we visualize Mamba's relative $L_2$ prediction error concerning time in the six LNO test cases in Figure \ref{fig:lno_rel_error_t}. Here, the relative $L_2$ error is computed over the entire dataset.
For the first group of cases, namely the Lorenz-5, Duffing-.5, and Pend-.5, where all models can achieve relatively low error, Mamba's relative $L_2$ error decreases with time. On the other hand, for the second group, Lorenz-10, Duffing-0, and Pend-0, Mamba's error grows with time.
This demonstrates that the ODE parameters affect Mamba's error distribution. When a damping effect is present, the system's motion gradually loses energy and attenuates, leading to a decrease in Mamba's error over time. Conversely, in the absence of a damping effect or in cases with chaotic solutions, the system's motion oscillates more violently, potentially resulting in more significant error for Mamba over extended long-time periods. At the initial point $t=0$, since we are given the initial condition, Mamba's initial error is usually small.

Overall, this computational experiment further demonstrates Mamba's outstanding stability and excellent OOD generalization ability compared with other models on a variety of ODE systems.

\subsection{Long-Time Integration}\label{sec:long_time_integration_lno_pend}
In this subsection, we will test Mamba for scalability, efficiency, stability, and generality in long-term integration. We solve the pendulum dynamical system over different terminal times and sequence lengths in Section \ref{sec:lnoode} following the work on LNO \cite{cao2023lno}. 
To be self-contained, we restate the problem definition of the driven gravity pendulum:
\begin{align}
\Ddot{x} + c\Dot{x} + \frac{g}{l} \sin(x) = f (t), \quad x(0) = 0, \quad \Dot{x}(0) = 0.
\end{align}
We aim to learn the operator $f(t) \mapsto x(t)$. We choose the damping $c \in \{0.1, 0.3, 0.5\}$ and $g / l = 1$. We also wish to test different models' stability under various damping factors $c$.
We manipulate the input and output sequence lengths in 2048, 4096, 8192, 16384, and 32768, corresponding to terminal times in 20.48, 40.96, 81,92, 163.84, and 327.68.
The original setting in Cao et al. \cite{cao2023lno} only considers terminal time up to 20.48, while we consider a much longer time.
We test the out-of-distribution (OOD) generalization problem following Cao et al. \cite{cao2023lno}, i.e., the training and testing functions/data will follow different distributions. 
Specifically, we sample 200 training samples from $f_{train}(t) =
A\sin(5t)$, where the $A \in \{0.05, 0.1, 0.15, \cdots, 9.95, 10\}$. We sample 50 validation samples and 130 testing samples from $f_{test}(t) = A\exp(-0.001t)\sin(5t)$, where $A \in \{0.14, 0.19, \cdots, 9.04, 9.09\}$. The temporal discretization step size is 0.01.
The implementation details, such as model structure, training procedure, and test metrics, are the same as those in Section \ref{sec:lnoode} due to the similar data generation setup.

Table \ref{tab:mamba_long_time_integration_scalability} presents different models' scalability and speed, while Table \ref{tab:mamba_long_time_integration_performance} presents the performance comparison results.
Regarding scalability presented in Table \ref{tab:mamba_long_time_integration_scalability}, we observe a linear growth in running time and memory cost for RNNs, linear Transformers (including Oformer-G, Oformer-F, and GNOT), and Mamba. 
On the other hand, Transformer and Oformer-V, with a conventional quadratic self-attention scheme, exhibit fast growth in memory and running time and cannot scale up to sequences longer than 8192, i.e., their memory costs exceed the limit of the used NVIDIA A100 GPU with 80GB memory. 
This demonstrates Mamba's linear complexity while keeping the model expressive and parallelizable. 
Furthermore, the neural operators' training time and memory cost are more stable with input sequence lengths since they disregard temporal information. However, this also leads to their inferior performance, as explained below.
Regarding performance comparisons presented in Table \ref{tab:mamba_long_time_integration_performance}, we report N.A. for models that require running longer than 10 hours (600 minutes) since it exceeds a reasonable training time or models going out-of-memory. 
Mamba performs best in most cases, especially when dealing with long-term problems. Mamba is only outperformed by other models in the pendulum problem with damping $c=0.1$ and sequence lengths 2048 and 4096, while Mamba is the best in all other settings.
Even in test cases where it is outperformed by other models, Mamba's performance is still close to the best one.
Mamba's errors grow more mildly than RNNs with the sequence length/terminal time, possibly because Mamba is better at capturing long-term dependencies than RNNs, which are known to forget.
The errors of Linear Transformers grow with sequence length, reflecting a tradeoff in Transformers' computational complexity and model expressiveness. More specifically, the  Transformer's costly self-attention scheme is motivated by capturing long sequences. Despite their efficiency, linear Transformers approximate the original quadratic self-attention trade-off computational complexity with the capability to capture long-term dependencies, leading to their fast error growth concerning sequence length. On the other hand, Mamba combines the advantages of both being efficient with linear cost and being expressive enough to capture long-time dependencies.
Regarding neural operators, we also observed a fast growth in their errors with the sequence length because they are general neural operators failing to capture the temporal information intrinsic in dynamical systems.
For instance, DeepONet's model parameters grow linearly with sequence length since it directly inputs the more extended input sequence.
FNO and LNO transform the input function using trainable kernel integration operators on a bounded domain. With a longer time, FNO and LNO compress the original temporal domain $[0, T]$ to a bounded domain, say $[0, 1]$. This domain transformation will cause the input function to contain more high-frequency signals, e.g., the long-frequency function $\sin (x)$ can be transformed to a high-frequency one $\sin (T x)$ when $T$ is large. Consequently, this makes the training and generalization more challenging for FNO and LNO, leading to their inferior performances under long-time integration.

\begin{table}[htbp]
\centering
	\begin{tabular}{|cc|c|c|c|c|c|}
\hline
\multicolumn{2}{|c|}{Sequence Length} & 2048 & 4096 & 8192 & 16384 & 32768 \\ \hline
\multicolumn{2}{|c|}{Terminal Time $T$} & 20.48 & 40.96 & 81.92 & 163.84 & 327.69 \\ \hline\hline
\multicolumn{1}{|c|}{\multirow{3}{*}{LSTM}} & Num of Params & 33537 & 33537 & 33537 & 33537 & 33537 \\ \cline{2-7} 
\multicolumn{1}{|c|}{} & Training Time & 17min & 30min & 57min & 108min & 217min \\ \cline{2-7} 
\multicolumn{1}{|c|}{} & Memory & 2073MiB & 2461MiB & 3231MiB & 4767MiB & 7839MiB \\ \hline\hline
\multicolumn{1}{|c|}{\multirow{3}{*}{GRU}} & Num of Params & 25217 & 25217 & 25217 & 25217 & 25217 \\ \cline{2-7} 
\multicolumn{1}{|c|}{} & Training Time & 16min & 29min & 55min & 110min & 220min \\ \cline{2-7} 
\multicolumn{1}{|c|}{} & Memory & 2017MiB & 2349MiB & 3007MiB & 4319MiB & 6943MiB \\ \hline\hline
\multicolumn{1}{|c|}{\multirow{3}{*}{DeepONet}} & Num of Params & 103176 & 201480 & 398088 & 791304 & 1577736 \\ \cline{2-7} 
\multicolumn{1}{|c|}{} & Training Time & 5min & 5min & 5min & 5min & 5min \\ \cline{2-7} 
\multicolumn{1}{|c|}{} & Memory & 1513MiB & 1527MiB & 1545MiB & 1637MiB & 1753MiB \\ \hline\hline
\multicolumn{1}{|c|}{\multirow{3}{*}{FNO}} & Num of Params & 27873 & 27873 & 27873 & 27873 & 27873 \\ \cline{2-7} 
\multicolumn{1}{|c|}{} & Training Time & 8min & 8min & 9min & 13min & 24min \\ \cline{2-7} 
\multicolumn{1}{|c|}{} & Memory & 1801MiB & 1883MiB & 1987MiB & 2211MiB & 2789MiB \\ \hline\hline
\multicolumn{1}{|c|}{\multirow{3}{*}{LNO}} & Num of Params & 1309 & 1309 & 1309 & 1309 & 1309 \\ \cline{2-7} 
\multicolumn{1}{|c|}{} & Training Time & 2min & 2min & 3min & 5min & 10min \\ \cline{2-7} 
\multicolumn{1}{|c|}{} & Memory & 1797MiB & 2145MiB & 2485MiB & 3255MiB & 4783MiB \\ \hline\hline
\multicolumn{1}{|c|}{\multirow{3}{*}{Transformer}} & Num of Params & 29041 & 29041 & 29041 & 29041 & 29041 \\ \cline{2-7} 
\multicolumn{1}{|c|}{} & Training Time & 233min & 893min & N.A. & N.A. & N.A. \\ \cline{2-7} 
\multicolumn{1}{|c|}{} & Memory & 12769MiB & 46765MiB & $>$80GB & $>$80GB & $>$80GB \\ \hline\hline
\multicolumn{1}{|c|}{\multirow{3}{*}{Oformer-V}} & Num of Params & 33505 & 33505 & 33505 & 33505 & 33505 \\ \cline{2-7} 
\multicolumn{1}{|c|}{} & Training Time & 162min & 606min & N.A. & N.A. & N.A. \\ \cline{2-7} 
\multicolumn{1}{|c|}{} & Memory & 8701MiB & 30315MiB & $>$80GB & $>$80GB & $>$80GB \\ \hline\hline
\multicolumn{1}{|c|}{\multirow{3}{*}{Oformer-G}} & Num of Params & 33361 & 33361 & 33361 & 33361 & 33361 \\ \cline{2-7} 
\multicolumn{1}{|c|}{} & Training Time & 37min & 73min & 144min & 287min & 571min \\ \cline{2-7} 
\multicolumn{1}{|c|}{} & Memory & 1777MiB & 2133MiB & 2879MiB & 4293MiB & 7167MiB \\ \hline\hline
\multicolumn{1}{|c|}{\multirow{3}{*}{Oformer-F}} & Num of Params & 33361 & 33361 & 33361 & 33361 & 33361 \\ \cline{2-7} 
\multicolumn{1}{|c|}{} & Training Time & 37min & 73min & 144min & 287min & 571min \\ \cline{2-7} 
\multicolumn{1}{|c|}{} & Memory & 1777MiB & 2133MiB & 2879MiB & 4293MiB & 7167MiB \\ \hline\hline
\multicolumn{1}{|c|}{\multirow{3}{*}{GNOT}} & Num of Params & 33479 & 33479 & 33479 & 33479 & 33479 \\ \cline{2-7} 
\multicolumn{1}{|c|}{} & Training Time & 78min & 152min & 302min & 605min & 1208 \\ \cline{2-7} 
\multicolumn{1}{|c|}{} & Memory & 1993MiB & 2443MiB & 3459MiB & 5293MiB & 9557MiB \\ \hline\hline
\multicolumn{1}{|c|}{\multirow{3}{*}{Mamba}} & Num of Params & 19233 & 19233 & 19233 & 19233 & 19233 \\ \cline{2-7} 
\multicolumn{1}{|c|}{} & Training Time & 11min & 17min & 32min & 58min & 115min \\ \cline{2-7} 
\multicolumn{1}{|c|}{} & Memory & 1751MiB & 1965MiB & 2489MiB & 3465MiB & 5187MiB \\ \hline
\end{tabular}
\caption{Long-time integration on the pendulum problem in Cao et al. \cite{cao2023lno} corresponding to Section \ref{sec:long_time_integration_lno_pend} in this paper. We show all models' scalability, such as the running time and memory cost in problems with different lengths.
The performance comparison is presented in Table \ref{tab:mamba_long_time_integration_performance}.
In addition, some of Mamba's representative predictions are visualized in Figure \ref{fig:lno_long_time_pend}.
Relative $L_2$ error with respect to time is plotted in Figure \ref{fig:lno_long_time_pend2}.
}
\label{tab:mamba_long_time_integration_scalability}
\end{table}

\begin{table}[]
\footnotesize
\centering
\begin{tabular}{|cccccc|}
\hline
\multicolumn{6}{|c|}{Pendulum with damping $c = 0.1$ taken from Cao et al. \cite{cao2023lno}} \\ \hline
\multicolumn{1}{|c|}{Seq. Length} & \multicolumn{1}{c|}{2048} & \multicolumn{1}{c|}{4096} & \multicolumn{1}{c|}{8192} & \multicolumn{1}{c|}{16384} & 32768 \\ \hline
\multicolumn{1}{|c|}{Time $T$} & \multicolumn{1}{c|}{20.48} & \multicolumn{1}{c|}{40.96} & \multicolumn{1}{c|}{81.92} & \multicolumn{1}{c|}{163.84} & 327.68 \\ \hline
\multicolumn{1}{|c|}{GRU} & \multicolumn{1}{c|}{2.479E-2$\pm$2.561E-3} & \multicolumn{1}{c|}{\textbf{2.640E-2$\pm$3.008E-3}} & \multicolumn{1}{c|}{1.199E-1$\pm$1.854E-2} & \multicolumn{1}{c|}{1.410E-1$\pm$2.691E-2} & 1.654E-1$\pm$3.178E-2 \\ \hline
\multicolumn{1}{|c|}{LSTM} & \multicolumn{1}{c|}{3.273E-2$\pm$4.558E-3} & \multicolumn{1}{c|}{3.919E-2$\pm$5.719E-3} & \multicolumn{1}{c|}{1.049E-1$\pm$1.089E-2} & \multicolumn{1}{c|}{\textcolor{red}{9.521E-2$\pm$1.227E-2}} & \textcolor{red}{9.319E-2$\pm$1.325E-2} \\ \hline
\multicolumn{1}{|c|}{DeepONet} & \multicolumn{1}{c|}{7.124E-1$\pm$6.178E-2} & \multicolumn{1}{c|}{7.527E-1$\pm$9.381E-2} & \multicolumn{1}{c|}{7.837E-1$\pm$1.399E-1} & \multicolumn{1}{c|}{8.308E-1$\pm$1.403E-1} & 8.623E-1$\pm$1.875E-1 \\ \hline
\multicolumn{1}{|c|}{FNO} & \multicolumn{1}{c|}{2.022E-2$\pm$8.840E-4} & \multicolumn{1}{c|}{3.518E-2$\pm$8.963E-4} & \multicolumn{1}{c|}{\textcolor{red}{4.273E-2$\pm$1.465E-3}} & \multicolumn{1}{c|}{1.299E-1$\pm$1.855E-3} & 3.483E-1$\pm$3.590E-3 \\ \hline
\multicolumn{1}{|c|}{LNO} & \multicolumn{1}{c|}{1.051E-1$\pm$3.547E-2} & \multicolumn{1}{c|}{1.435E-1$\pm$4.097E-2} & \multicolumn{1}{c|}{8.787E-1$\pm$8.266E-2} & \multicolumn{1}{c|}{9.722E-1$\pm$7.454E-2} & 9.827E-1$\pm$1.267E-1 \\ \hline
\multicolumn{1}{|c|}{Transformer} & \multicolumn{1}{c|}{4.488E-2$\pm$3.284E-3} & \multicolumn{1}{c|}{N.A.} & \multicolumn{1}{c|}{N.A.} & \multicolumn{1}{c|}{N.A.} & N.A. \\ \hline
\multicolumn{1}{|c|}{Oformer-V} & \multicolumn{1}{c|}{\textbf{1.740E-2$\pm$9.853E-4}} & \multicolumn{1}{c|}{N.A.} & \multicolumn{1}{c|}{N.A.} & \multicolumn{1}{c|}{N.A.} & N.A. \\ \hline
\multicolumn{1}{|c|}{Oformer-G} & \multicolumn{1}{c|}{2.590E-2$\pm$1.720E-3} & \multicolumn{1}{c|}{6.169E-1$\pm$5.541E-2} & \multicolumn{1}{c|}{7.211E-1$\pm$5.965E-2} & \multicolumn{1}{c|}{Diverge} & Diverge \\ \hline
\multicolumn{1}{|c|}{Oformer-F} & \multicolumn{1}{c|}{2.368E-2$\pm$1.998E-3} & \multicolumn{1}{c|}{5.817E-1$\pm$6.874E-2} & \multicolumn{1}{c|}{6.980E-1$\pm$9.506E-2} & \multicolumn{1}{c|}{Diverge} & Diverge \\ \hline
\multicolumn{1}{|c|}{GNOT} & \multicolumn{1}{c|}{2.870E-2$\pm$5.266E-3} & \multicolumn{1}{c|}{4.378E-2$\pm$8.089E-3} & \multicolumn{1}{c|}{7.693E-2$\pm$1.286E-2} & \multicolumn{1}{c|}{N.A.} & N.A. \\ \hline
\multicolumn{1}{|c|}{Mamba} & \multicolumn{1}{c|}{\textcolor{red}{1.785E-2$\pm$6.106E-3}} & \multicolumn{1}{c|}{\textcolor{red}{3.181E-2$\pm$8.263E-3}} & \multicolumn{1}{c|}{\textbf{3.803E-2$\pm$8.615E-3}} & \multicolumn{1}{c|}{\textbf{3.415E-2$\pm$7.128E-3}} & \textbf{3.461E-2$\pm$6.988E-3} \\ \hline
%\end{tabular}
%\begin{tabular}{|cccccc|}
\hline
\multicolumn{6}{|c|}{Pendulum with damping $c = 0.3$ taken from Cao et al. \cite{cao2023lno}} \\ \hline
\multicolumn{1}{|c|}{Seq. Length} & \multicolumn{1}{c|}{2048} & \multicolumn{1}{c|}{4096} & \multicolumn{1}{c|}{8192} & \multicolumn{1}{c|}{16384} & 32768 \\ \hline
\multicolumn{1}{|c|}{Time $T$} & \multicolumn{1}{c|}{20.48} & \multicolumn{1}{c|}{40.96} & \multicolumn{1}{c|}{81.92} & \multicolumn{1}{c|}{163.84} & 327.68 \\ \hline
\multicolumn{1}{|c|}{GRU} & \multicolumn{1}{c|}{\textcolor{red}{5.249E-3$\pm$9.880E-4}} & \multicolumn{1}{c|}{7.939E-3$\pm$1.562E-3} & \multicolumn{1}{c|}{1.009E-2$\pm$1.778E-3} & \multicolumn{1}{c|}{1.496E-2$\pm$3.741E-3} & \textcolor{red}{1.309E-2$\pm$3.268E-3} \\ \hline
\multicolumn{1}{|c|}{LSTM} & \multicolumn{1}{c|}{7.350E-3$\pm$2.064E-3} & \multicolumn{1}{c|}{\textcolor{red}{7.854E-3$\pm$2.775E-3}} & \multicolumn{1}{c|}{\textcolor{red}{7.168E-3$\pm$1.957E-3}} & \multicolumn{1}{c|}{\textcolor{red}{6.867E-3$\pm$1.276E-3}} & 5.743E-2$\pm$8.745E-3 \\ \hline
\multicolumn{1}{|c|}{DeepONet} & \multicolumn{1}{c|}{5.994E-1$\pm$7.845E-2} & \multicolumn{1}{c|}{6.785E-1$\pm$8.233E-2} & \multicolumn{1}{c|}{7.781E-1$\pm$8.547E-2} & \multicolumn{1}{c|}{8.416E-1$\pm$1.067E-1} & 8.954E-1$\pm$1.854E-1 \\ \hline
\multicolumn{1}{|c|}{FNO} & \multicolumn{1}{c|}{7.346E-3$\pm$8.772E-4} & \multicolumn{1}{c|}{1.001E-2$\pm$6.945E-4} & \multicolumn{1}{c|}{1.237E-2$\pm$1.451E-3} & \multicolumn{1}{c|}{4.259E-2$\pm$3.337E-3} & 6.783E-2$\pm$8.976E-3 \\ \hline
\multicolumn{1}{|c|}{LNO} & \multicolumn{1}{c|}{8.079E-2$\pm$8.815E-3} & \multicolumn{1}{c|}{2.535E-1$\pm$1.745E-2} & \multicolumn{1}{c|}{7.661E-1$\pm$5.676E-2} & \multicolumn{1}{c|}{9.687E-1$\pm$9.763E-2} & 9.904E-1$\pm$1.050E-1 \\ \hline
\multicolumn{1}{|c|}{Transformer} & \multicolumn{1}{c|}{1.773E-2$\pm$2.945E-3} & \multicolumn{1}{c|}{N.A.} & \multicolumn{1}{c|}{N.A.} & \multicolumn{1}{c|}{N.A.} & N.A. \\ \hline
\multicolumn{1}{|c|}{Oformer-V} & \multicolumn{1}{c|}{9.266E-3$\pm$1.089E-3} & \multicolumn{1}{c|}{N.A.} & \multicolumn{1}{c|}{N.A.} & \multicolumn{1}{c|}{N.A.} & N.A. \\ \hline
\multicolumn{1}{|c|}{Oformer-G} & \multicolumn{1}{c|}{1.527E-2$\pm$2.062E-3} & \multicolumn{1}{c|}{7.951E-2$\pm$4.712E-3} & \multicolumn{1}{c|}{1.438E-1$\pm$1.046E-2} & \multicolumn{1}{c|}{Diverge} & Diverge \\ \hline
\multicolumn{1}{|c|}{Oformer-F} & \multicolumn{1}{c|}{1.326E-2$\pm$2.556E-3} & \multicolumn{1}{c|}{7.819E-2$\pm$3.561E-3} & \multicolumn{1}{c|}{1.195E-1$\pm$7.892E-3} & \multicolumn{1}{c|}{Diverge} & Diverge \\ \hline
\multicolumn{1}{|c|}{GNOT} & \multicolumn{1}{c|}{1.066E-2$\pm$2.385E-3} & \multicolumn{1}{c|}{1.995E-2$\pm$3.686E-3} & \multicolumn{1}{c|}{3.889E-2$\pm$5.371E-3} & \multicolumn{1}{c|}{N.A.} & N.A. \\ \hline
\multicolumn{1}{|c|}{Mamba} & \multicolumn{1}{c|}{\textbf{4.725E-3$\pm$1.288E-3}} & \multicolumn{1}{c|}{\textbf{4.465E-3$\pm$1.293E-3}} & \multicolumn{1}{c|}{\textbf{5.327E-3$\pm$1.760E-3}} & \multicolumn{1}{c|}{\textbf{4.543E-3$\pm$1.885E-3}} & \textbf{4.992E-3$\pm$1.925E-3} \\ \hline
%\end{tabular}
%\begin{tabular}{|cccccc|}
\hline
\multicolumn{6}{|c|}{Pendulum with damping $c = 0.5$ taken from Cao et al. \cite{cao2023lno}} \\ \hline
\multicolumn{1}{|c|}{Seq. Length} & \multicolumn{1}{c|}{2048} & \multicolumn{1}{c|}{4096} & \multicolumn{1}{c|}{8192} & \multicolumn{1}{c|}{16384} & 32768 \\ \hline
\multicolumn{1}{|c|}{Time $T$} & \multicolumn{1}{c|}{20.48} & \multicolumn{1}{c|}{40.96} & \multicolumn{1}{c|}{81.92} & \multicolumn{1}{c|}{163.84} & 327.68 \\ \hline
\multicolumn{1}{|c|}{GRU} & \multicolumn{1}{c|}{3.325E-3$\pm$5.966E-4} & \multicolumn{1}{c|}{4.434E-3$\pm$4.850E-4} & \multicolumn{1}{c|}{4.053E-3$\pm$4.988E-4} & \multicolumn{1}{c|}{\textcolor{red}{6.893E-3$\pm$5.022E-4}} & 3.371E-2$\pm$4.741E-3 \\ \hline
\multicolumn{1}{|c|}{LSTM} & \multicolumn{1}{c|}{\textcolor{red}{3.222E-3$\pm$6.288E-4}} & \multicolumn{1}{c|}{\textcolor{red}{3.399E-3$\pm$6.796E-4}} & \multicolumn{1}{c|}{\textcolor{red}{3.203E-3$\pm$4.475E-4}} & \multicolumn{1}{c|}{7.658E-3$\pm$6.296E-4} & \textcolor{red}{4.803E-3$\pm$2.953E-4} \\ \hline
\multicolumn{1}{|c|}{DeepONet} & \multicolumn{1}{c|}{6.375E-1$\pm$6.974E-2} & \multicolumn{1}{c|}{7.199E-1$\pm$8.205E-2} & \multicolumn{1}{c|}{8.296E-1$\pm$7.236E-2} & \multicolumn{1}{c|}{8.859E-1$\pm$7.784E-2} & 9.286E-1$\pm$8.006E-2 \\ \hline
\multicolumn{1}{|c|}{FNO} & \multicolumn{1}{c|}{4.633E-3$\pm$9.870E-4} & \multicolumn{1}{c|}{8.257E-3$\pm$1.455E-3} & \multicolumn{1}{c|}{1.264E-2$\pm$3.809E-3} & \multicolumn{1}{c|}{8.723E-2$\pm$1.066E-2} & 1.476E-1$\pm$3.964E-2 \\ \hline
\multicolumn{1}{|c|}{LNO} & \multicolumn{1}{c|}{1.423E-1$\pm$3.871E-2} & \multicolumn{1}{c|}{6.970E-2$\pm$1.055E-2} & \multicolumn{1}{c|}{5.413E-1$\pm$8.674E-2} & \multicolumn{1}{c|}{9.605E-1$\pm$1.048E-1} & 9.915E-1$\pm$1.474E-1 \\ \hline
\multicolumn{1}{|c|}{Transformer} & \multicolumn{1}{c|}{1.278E-2$\pm$6.871E-4} & \multicolumn{1}{c|}{N.A.} & \multicolumn{1}{c|}{N.A.} & \multicolumn{1}{c|}{N.A.} & N.A. \\ \hline
\multicolumn{1}{|c|}{Oformer-V} & \multicolumn{1}{c|}{8.938E-3$\pm$1.065E-3} & \multicolumn{1}{c|}{N.A.} & \multicolumn{1}{c|}{N.A.} & \multicolumn{1}{c|}{N.A.} & N.A. \\ \hline
\multicolumn{1}{|c|}{Oformer-G} & \multicolumn{1}{c|}{1.210E-2$\pm$4.508E-4} & \multicolumn{1}{c|}{5.695E-2$\pm$8.773E-4} & \multicolumn{1}{c|}{1.468E-1$\pm$1.986E-2} & \multicolumn{1}{c|}{Diverge} & Diverge \\ \hline
\multicolumn{1}{|c|}{Oformer-F} & \multicolumn{1}{c|}{1.086E-2$\pm$5.267E-4} & \multicolumn{1}{c|}{5.301E-2$\pm$9.880E-3} & \multicolumn{1}{c|}{1.584E-1$\pm$2.352E-2} & \multicolumn{1}{c|}{Diverge} & Diverge \\ \hline
\multicolumn{1}{|c|}{GNOT} & \multicolumn{1}{c|}{9.174E-3$\pm$2.054E-3} & \multicolumn{1}{c|}{1.483E-2$\pm$3.833E-3} & \multicolumn{1}{c|}{2.197E-2$\pm$3.568E-4} & \multicolumn{1}{c|}{N.A.} & N.A. \\ \hline
\multicolumn{1}{|c|}{Mamba} & \multicolumn{1}{c|}{\textbf{2.131E-3$\pm$7.819E-4}} & \multicolumn{1}{c|}{\textbf{2.490E-3$\pm$8.020E-4}} & \multicolumn{1}{c|}{\textbf{2.533E-3$\pm$8.114E-4}} & \multicolumn{1}{c|}{\textbf{3.862E-3$\pm$1.185E-3}} & \textbf{2.030E-3$\pm$5.640E-4} \\ \hline
\end{tabular}
\caption{Results for long-time integration on the pendulum problem in Cao et al. \cite{cao2023lno} corresponding to Section \ref{sec:long_time_integration_lno_pend} in this paper. 
\textbf{The best model is bold} and \textcolor{red}{the second best is red}.
Different models' scalability and training speeds are presented in Table \ref{tab:mamba_long_time_integration_scalability}.
We visualize Mamba's prediction in Figure \ref{fig:lno_long_time_pend}.
Relative $L_2$ error with respect to time is plotted in Figure \ref{fig:lno_long_time_pend2}.
}
\label{tab:mamba_long_time_integration_performance}
\end{table}

\begin{figure}[htbp]
\centering
\includegraphics[width=0.32\linewidth]{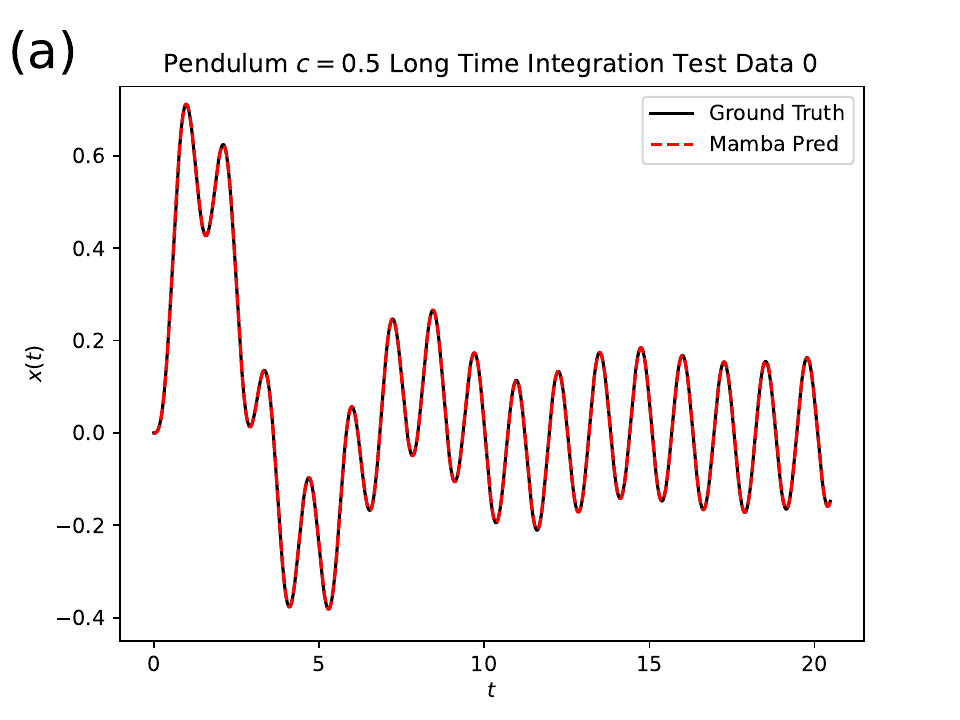}
\includegraphics[width=0.32\linewidth]{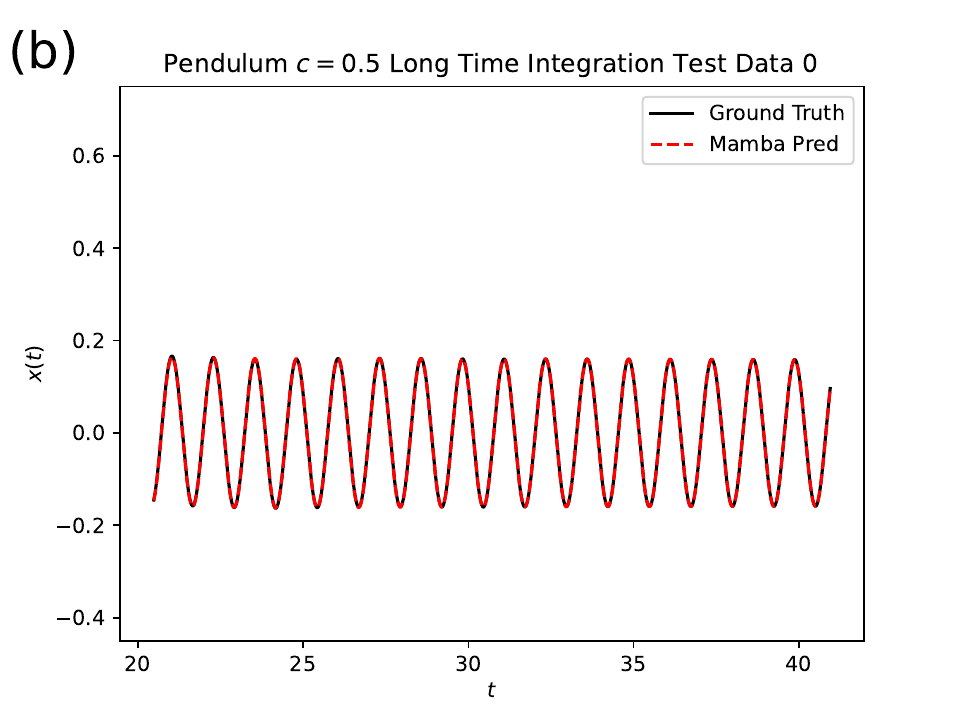}
\includegraphics[width=0.32\linewidth]{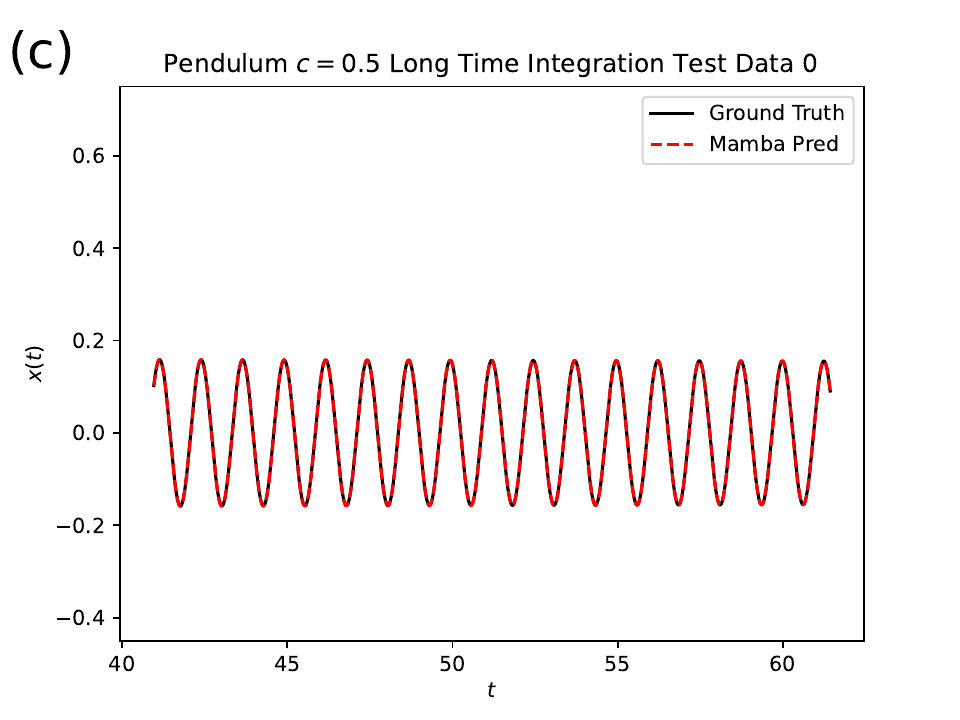}
\includegraphics[width=0.32\linewidth]{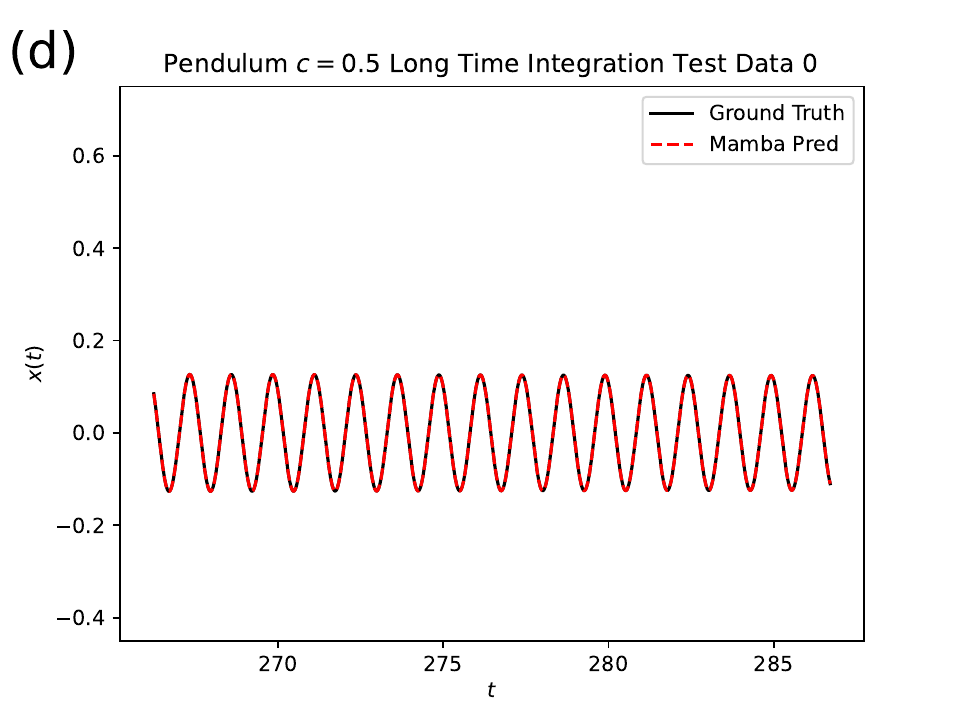}
\includegraphics[width=0.32\linewidth]{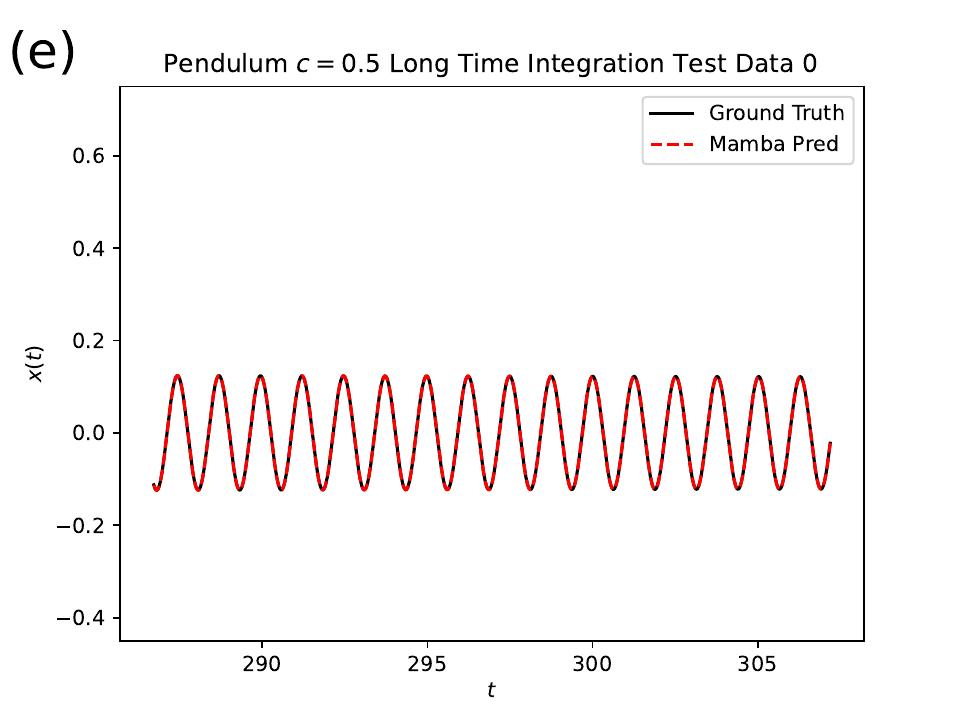}
\includegraphics[width=0.32\linewidth]{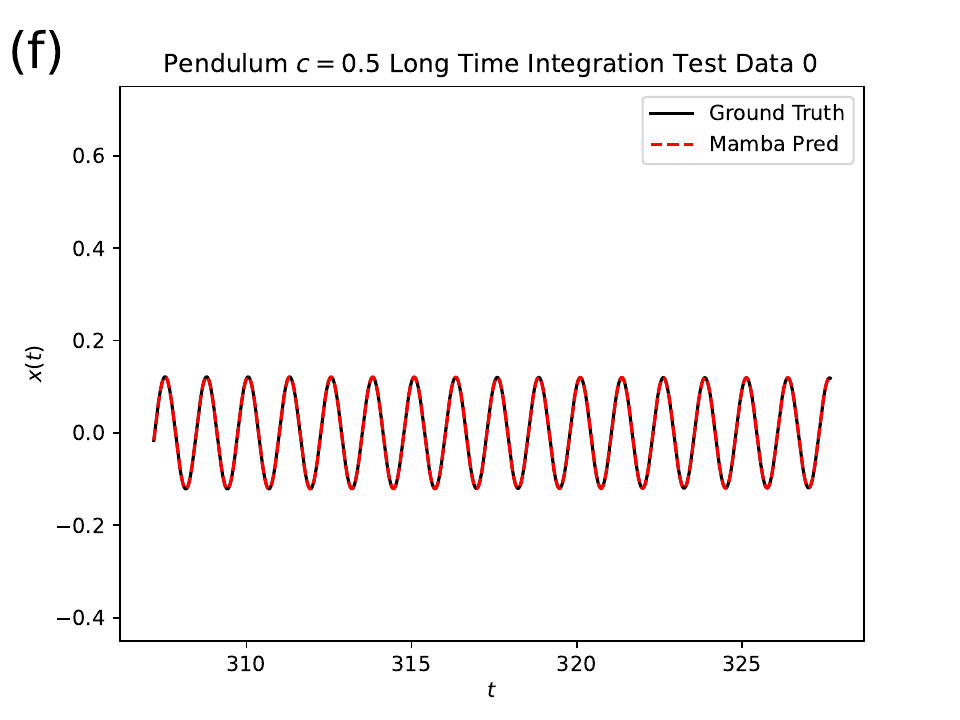}
\caption{Visualization of Mamba's prediction on the pendulum problem in Cao et al. \cite{cao2023lno} for long-time integration corresponding to Section \ref{sec:long_time_integration_lno_pend} in this paper. 
(a) $T \in [0, 20.48]$. 
(b) $T \in [20.48, 40.96]$.
(c) $T \in [40.96, 61.44]$.
(d) $T \in [266.24, 286.72]$.
(e) $T \in [286.72, 307.2]$.
(f) $T \in [307.2, 327.68]$.
The relative $L_2$ error with respect to time is plotted in Figure \ref{fig:lno_long_time_pend2}.
The performance and scalability comparisons between different models are presented in Tables \ref{tab:mamba_long_time_integration_scalability} and \ref{tab:mamba_long_time_integration_performance}.
}
\label{fig:lno_long_time_pend}
\end{figure}

\begin{figure}[htbp]
\centering
\includegraphics[width=0.32\linewidth]{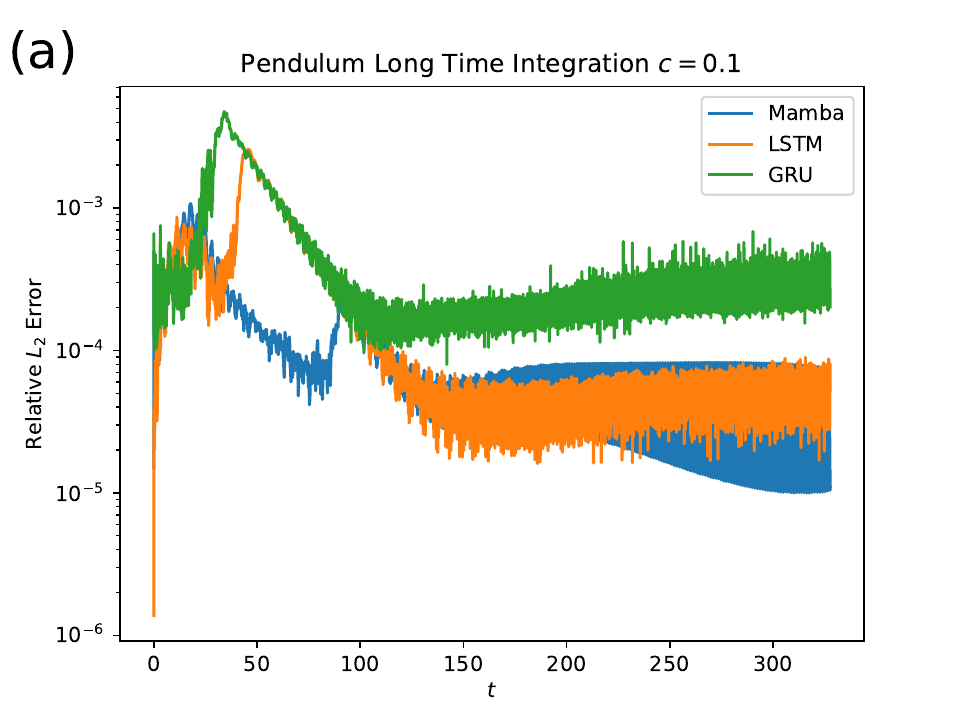}
\includegraphics[width=0.32\linewidth]{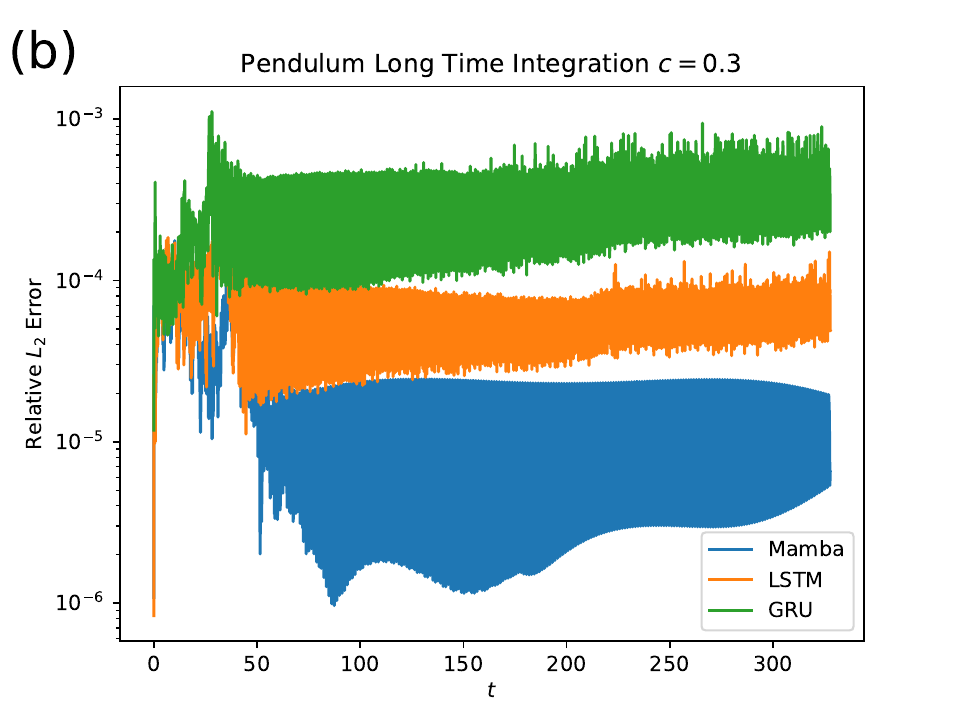}
\includegraphics[width=0.32\linewidth]{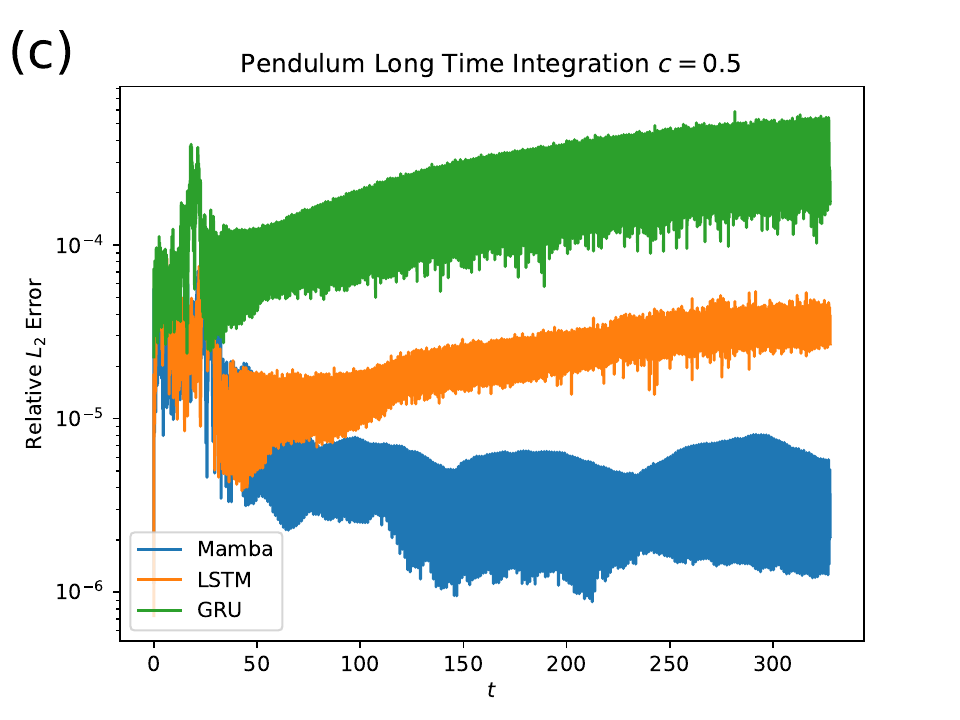}
\caption{Visualization of Mamba’s relative $L_2$ prediction error concerning time in the long-time pendulum problem taken from Cao et al. \cite{cao2023lno} corresponding to Section \ref{sec:long_time_integration_lno_pend} of this paper.
(a) $c = 0.1$. (b) $c = 0.3$. (c) $c = 0.5$.
The performance and scalability comparisons between different models are presented in Tables \ref{tab:mamba_long_time_integration_scalability} and \ref{tab:mamba_long_time_integration_performance}.
We visualize Mamba's prediction in Figure \ref{fig:lno_long_time_pend}.
}
\label{fig:lno_long_time_pend2}
\end{figure}

We plot Mamba's prediction versus the ground truth on test data 0 in the Pendulum problem with $c=0.5$ and the most extended sequence, 32768, in Figure \ref{fig:lno_long_time_pend}. The ODE dynamics are complicated at the beginning. However, as time evolves and with the damping effect, the solution gradually converges to a sinusoidal-like function. Since Mamba achieves a relative $L_2$ error of about 1e-3 in this case, the ground truth and its prediction are almost indistinguishable, further validating Mamba's outstanding accuracy.
We plot the relative $L_2$ error with respect to time for Pendulum problems with $c=0.1, 0.3, 0.5$ and the most extended sequence, 32768, for LSTM, GRU, and Mamba, in Figure \ref{fig:lno_long_time_pend2}, which further validates the above observation. Specifically, the errors are larger at the beginning due to the relatively more complex dynamics there, as shown in the figure. At later times, the dynamics gradually converge to a sinusoidal-like function while the errors drop and maintain a steady level.

Overall, the long-time integration test case further validates Mamba's linear complexity, efficiency, and strong expressiveness to capture long-term dependencies.

\subsection{Extrapolation to Long-Time Integration Problems}\label{sec:extrapolation_long_time_1D_DS_deeponet}
We revisit the \textbf{gravity pendulum} with an external force in Section \ref{sec:1D_DS_DeepONet} proposed in DeepONet \cite{lu2019deeponet}:
\begin{align*}
s_1'(t) = s_2(t), \quad s_2'(t) = -\sin\left(s_1(t)\right) + u(t), \quad t \in [0, T],
\end{align*}
with an initial condition $s_1(0) = s_2(0) = 0$. We aim to learn the operator $u(t) \mapsto s_1(t)$. 
For the extrapolation experiment, we train the models on $[0, 1]$ and test them on $[0, 1], [0, 2], [0, 3], [0, 4]$. 
Specifically, we train the models to only learn $u(t) \mapsto s_1(t)$ for $0 \leq t \leq 1$, i.e., the input is the $u(t)$'s value in $[0, 1]$ and the output is $s_1(t)$'s value in $[0, 1]$. But we test them to $u(t) \mapsto s_1(t)$ for $0 \leq t \leq T$, where $T=2,3,4$, i.e., the input is the $u(t)$'s value in $[0, T]$ and the output is $s_1(t)$'s value in $[0, T]$.
Here, testing on $[0, 1]$ is interpolation, while the other mappings are more challenging extrapolations. The goal is to test whether various models capture the ODE operator over a long time rather than overfitting the short interval.

Here are the implementation details. We generate 10K train and 10K test data with a sensor/discretization size of $100 T$ where $T$ is the terminal time, i.e., the sensors are on $\{0.01, 0.02, \cdots, T - 0.01, T\}$ where the input function $u(t)$ and the target function $s(t)/s_1(t)$ are discretized. 
The training loss is MSE, while the test metric is the relative $L_2$ error to ensure comparability between tests on various intervals.
We train all models for 10,001 epochs. 
In each epoch, we traverse the dataset with a minibatch size of 128.
We use the Adam optimizer \cite{kingma2014adam} with a 1e-3 initial learning rate, which decays linearly to zero at the end of training.
{\color{green!50!black}We keep the number of parameters and model structures of different models similar for a fair comparison, following the first experiment (Section \ref{sec:1D_DS_DeepONet}) exactly since the training dataset and the validation setup are the same.}

Regarding tested models, this extrapolation test case requires the model to be able to intake variable-length inputs. The model is trained on shorter inputs within $[0, 1]$ but is tested on longer intervals $[0, T]$ where $T=2,3,4$. Hence, we test GRU, LSTM, Mamba, and various transformers that are capable of predicting variable inputs. We adopt exactly the same model structures as in Section \ref{sec:1D_DS_DeepONet}. Since the training set and procedure are the same as those in Section \ref{sec:1D_DS_DeepONet}, we report the extrapolation results only as other information, including training time, memory cost, and parameter count are precisely the same as the results in Section \ref{sec:1D_DS_DeepONet} due to the same model setting and training data. The only difference is we test extrapolation.

\begin{table}[htbp]
\centering
\begin{tabular}{|c|c|c|c|c|}
\hline
Test Interval & GRU & LSTM & Transformer & Oformer-V \\ \hline
{[}0,1{]} & 8.292E-4$\pm$3.426E-4 & 7.718E-4$\pm$2.517E-4 & 2.207E-2$\pm$2.697E-4 & 4.324E-4$\pm$2.897E-5 \\ \hline
{[}0,2{]} & 9.169E-2$\pm$1.545E-2 & 1.596E-1$\pm$2.144E-2 & 9.190E-1$\pm$1.702E-2 & 8.118E-1$\pm$4.582E-2 \\ \hline
{[}0,3{]} & 2.669E-1$\pm$3.573E-2 & 4.302E-1$\pm$6.090E-2 & 9.625E-1$\pm$9.573E-3 & 9.452E-1$\pm$1.244E-2 \\ \hline
{[}0,4{]} & 4.411E-1$\pm$4.031E-2 & 6.161E-1$\pm$8.320E-2 & 9.747E-1$\pm$3.231E-3 & 9.638E-1$\pm$1.075E-2 \\ \hline\hline
Test Interval & Oformer-G & Oformer-F & GNOT & Mamba \\ \hline
{[}0,1{]} & 4.228E-4$\pm$1.457E-4 & 4.096E-4$\pm$1.479E-4 & 3.700E-4$\pm$8.855E-6 & \textbf{2.175E-4$\pm$1.533E-5} \\ \hline
{[}0,2{]} & 8.145E-1$\pm$5.908E-2 & 8.165E-1$\pm$1.849E-3 & 7.099E-1$\pm$1.626E-2 & \textbf{2.823E-2$\pm$5.125E-3} \\ \hline
{[}0,3{]} & 9.473E-1$\pm$1.444E-2 & 9.460E-1$\pm$4.959E-3 & 9.120E-1$\pm$1.538E-2 & \textbf{1.475E-1$\pm$1.431E-2} \\ \hline
{[}0,4{]} & 9.622E-1$\pm$4.119E-3 & 9.522E-1$\pm$8.819E-3 & 9.515E-1$\pm$1.623E-2 & \textbf{3.451E-1$\pm$1.952E-2} \\ \hline
\end{tabular}
\caption{Results for extrapolation on the long time gravity pendulum problem following DeepONet \cite{lu2019deeponet} corresponding to Section \ref{sec:extrapolation_long_time_1D_DS_deeponet} in this paper. 
We report the different models' relative $L_2$ errors for clear comparisons between tests on various intervals. The best-performing model, Mamba, is shown in bold.
The visualization of Mamba's and GRU's predictions are presented in Figure \ref{fig:1D_DS_pend_extrapolation} (a, b, c). Mamba's and GRU's relative $L_2$ errors with respect to time is in Figure \ref{fig:1D_DS_pend_extrapolation} (d).}
\label{tab:extrapolation}
\end{table}

\begin{figure}[htbp]
\centering
\includegraphics[width=0.24\linewidth]{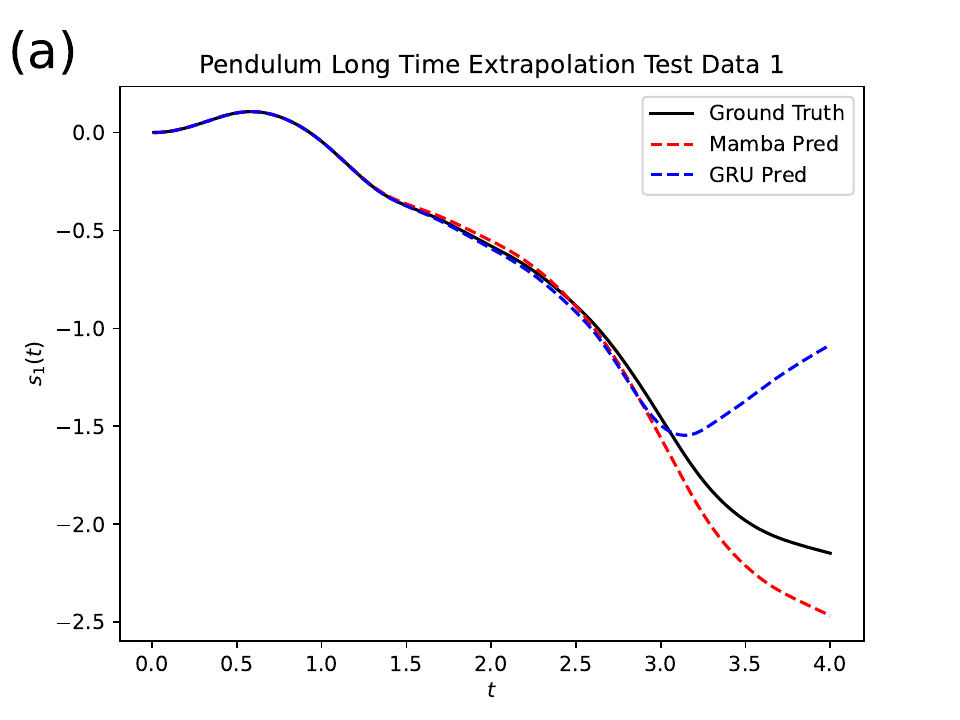}
\includegraphics[width=0.24\linewidth]{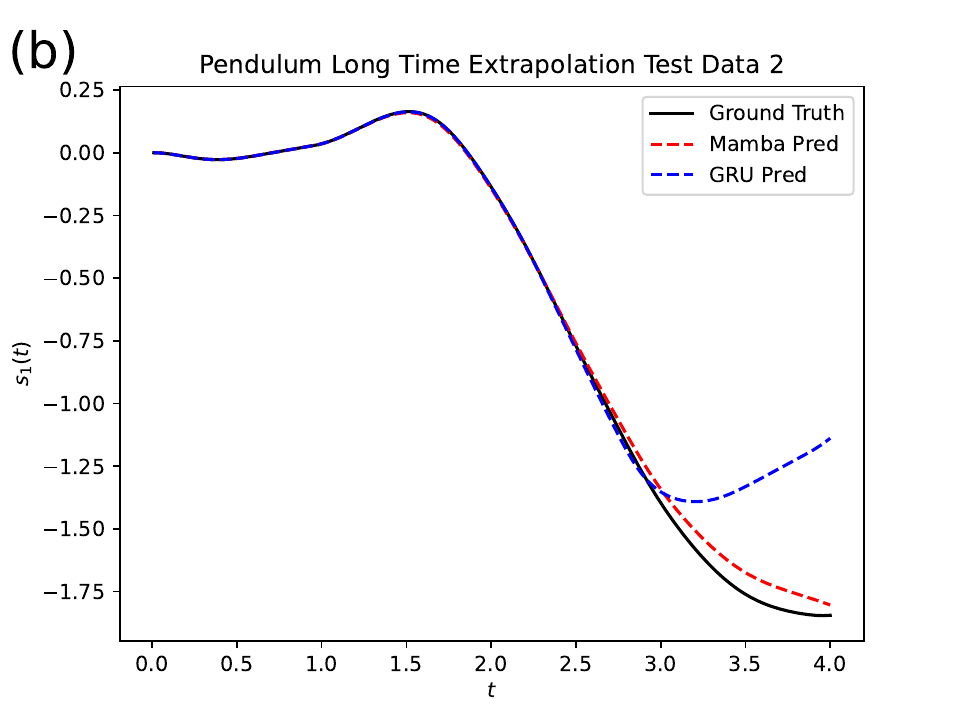}
\includegraphics[width=0.24\linewidth]{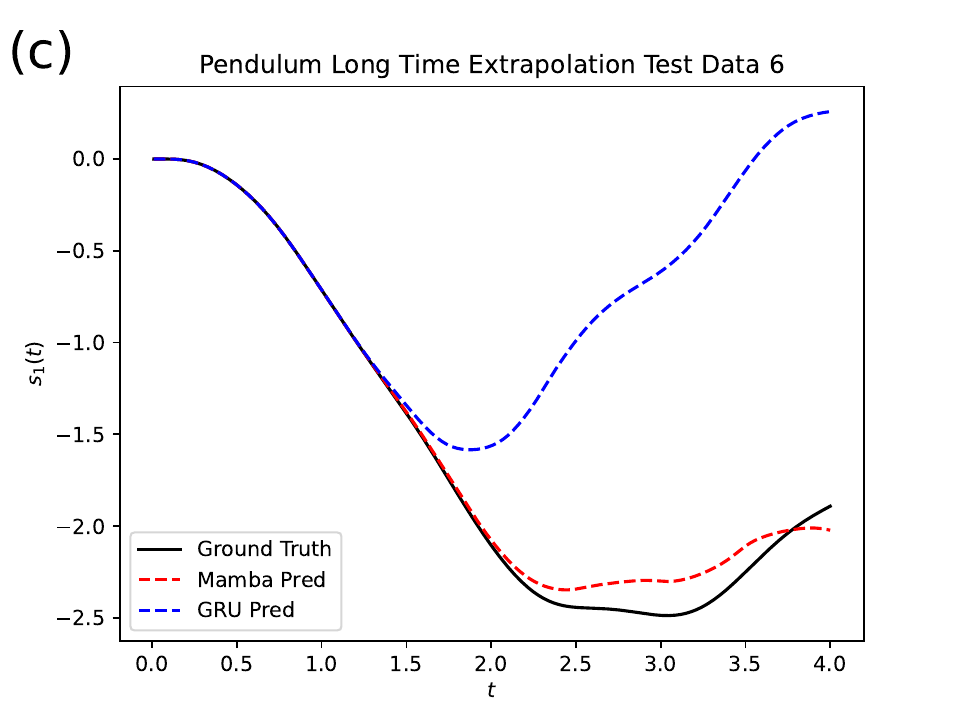}
\includegraphics[width=0.24\linewidth]{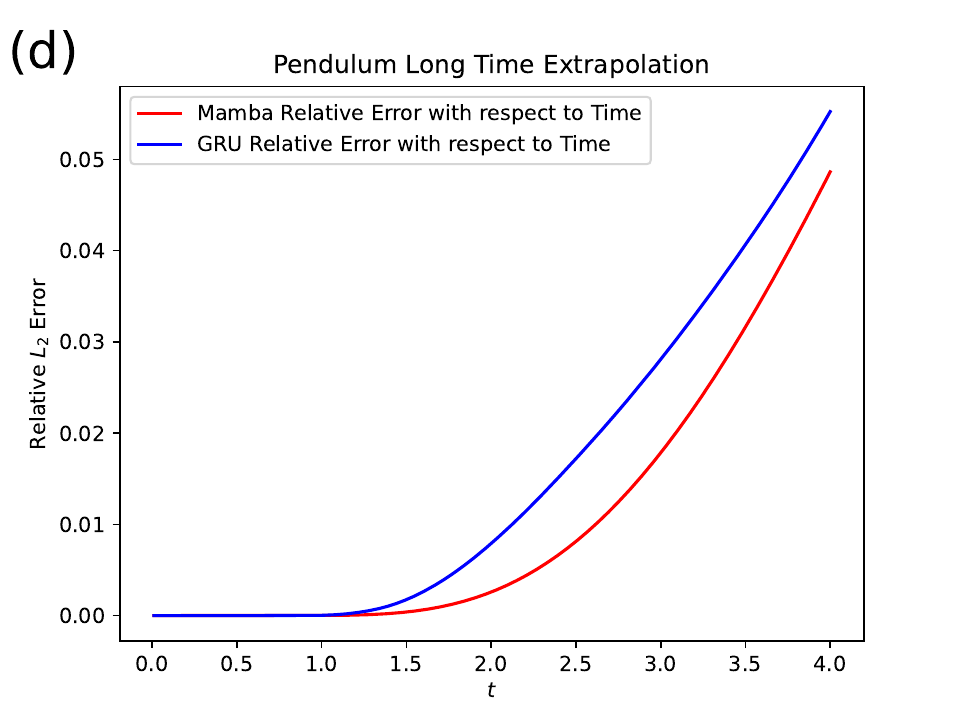}
\caption{Visualization of Mamba's (red) and GRU's (blue) prediction versus the ground truth (black) on the gravity pendulum ODE for extrapolation to long time problems following DeepONet \cite{lu2019deeponet} corresponding to Section \ref{sec:extrapolation_long_time_1D_DS_deeponet} in this paper. (a) Prediction on test data number 1. (b) Prediction on test data number 2. (c) Prediction on test data number 6. (d) Mamba's and GRU's relative $L_2$ errors with respect to time.
The quantitative results are presented in Table \ref{tab:extrapolation}.
}
\label{fig:1D_DS_pend_extrapolation}
\end{figure}

Results are shown in Table \ref{tab:extrapolation}. Mamba outperforms other models in both interpolation and extrapolation tasks, demonstrating its ability to effectively learn from ODE operators and generalize to longer-term problems. 
Regarding models other than Mamba, RNNs are better at extrapolation, while Transformers are not.
The application of extrapolation allows us to train only on short-time data, which is much more cost-effective, and subsequently test on large-scale, long-term problems. Training directly on long-term problems would incur significantly higher costs in comparison.
Since Mamba is the best model and GRU is the second best model, Figure \ref{fig:1D_DS_pend_extrapolation} (a, b, c) visualizes some of Mamba's and GRU's prediction and Figure \ref{fig:1D_DS_pend_extrapolation} (d) shows their relative $L_2$ errors over the entire dataset with respect to time.
Both visualizations demonstrate that Mamba's extrapolation error grows with time since Mamba is only trained on data in $[0,1]$ while predicting on a longer $[0, 4]$.
The visualization results also demonstrate that Mamba outperforms GRU.

\subsection{Chaotic Lorenz System with Various Initial Conditions}\label{sec:Lorenz_init}
So far, we have focused on solving sequence-to-sequence problems in dynamical systems. This is challenging since the model's input sequence is a high-dimensional vector due to the long time plus some additional ODE coefficient/parameter. 
Another crucial setting is to solve ODE and dynamical systems with initial conditions only, i.e., the input is just the initial conditions, and the output is the ODE solution over a long time. 
For the test case, we consider a chaotic Lorenz system with random initial conditions following Li et al. \cite{li2021learning}:
\begin{align}
\Dot{x} &= \sigma (y - x).\\
\Dot{y} &= x(\rho - z) - y.\\
\Dot{z} &= xy - \beta z.
\end{align}
We aim to learn the operator from the random initial condition to the state variable $x(t)$. We opt for
$\sigma = 10$ and $\beta = 8/3$ and $\rho=28$.
More specifically, we generate 10K data with the initial condition drawn from $x(0) \sim \text{Unif}(-10, 10), y(0) \sim \text{Unif}(-20, 30), z(0) \sim \text{Unif}(10, 40)$ and 90\% for training while another 10\% for testing. 
The target is the ODE solution over $[0, T]$ discretized with a step size 1e-3, i.e., we wish to predict the ODE solution values on the $1000 \cdot T$ grid points $\{0.001, 0.002, \cdots, T\}$.
We test for $T=1,2,3$. 
The training loss is MSE, while the test metric is the relative $L_2$ error to ensure comparability between tests on various terminal times.
We train all models for 10,001 epochs. 
In each epoch, we traverse the dataset with a minibatch size of 128.
We use the Adam optimizer \cite{kingma2014adam} with a 1e-3 initial learning rate, which decays linearly to zero at the end of training. We append the temporal grid to the input data tensor
for augmentation.

Regarding tested models, since the problem is a long sequence, we test the efficient LSTM, GRU, and Mamba with two different structures and model sizes.
For LSTM and GRU, we test one-layer models with 32 or 64 hidden units.
For Mamba, we test a one-block model with 16 or 32 model
dimensions and intermediate dimensions.
{\color{green!50!black}The goal is to test models with similar and comparable structures (all of them are one-layer models) but various hidden sizes. LSTM and GRU with 32 (64) hidden units have similar parameter counts with Mamba with 16 (32) hidden units.}

\begin{table}[htbp]
\centering
\begin{tabular}{|cc|c|c|c|}
\hline
\multicolumn{2}{|c|}{} & GRU1 & LSTM1 & Mamba1 \\ \hline
\multicolumn{2}{|c|}{Layer Num} & 1 & 1 & 1 \\ \hline
\multicolumn{2}{|c|}{Hidden Size} & 32 & 32 & 16 \\ \hline
\multicolumn{2}{|c|}{Params} & 6529 & 8641 & 9679 \\ \hline
\multicolumn{1}{|c|}{\multirow{3}{*}{$T=1$}} & Memory & 2397MiB & 2507MiB & 2157MiB \\ \cline{2-5} 
\multicolumn{1}{|c|}{} & Time & 59min & 58min & 95min \\ \cline{2-5} 
\multicolumn{1}{|c|}{} & Rel. $L_2$ Err & 3.909E-2$\pm$9.204E-3 & 3.720E-2$\pm$1.684E-2 & 3.991E-2$\pm$8.022E-3 \\ \hline
\multicolumn{1}{|c|}{\multirow{3}{*}{$T=2$}} & Memory & 3229MiB & 3447MiB & 2919MiB \\ \cline{2-5} 
\multicolumn{1}{|c|}{} & Time & 117min & 94min & 193min \\ \cline{2-5} 
\multicolumn{1}{|c|}{} & Rel. $L_2$ Err & NaN & NaN & 1.338E-1$\pm$5.959E-3 \\ \hline
\multicolumn{1}{|c|}{\multirow{3}{*}{$T=3$}} & Memory & 4063MiB & 4389MiB & 3687MiB \\ \cline{2-5} 
\multicolumn{1}{|c|}{} & Time & 181min & 134min & 296min \\ \cline{2-5} 
\multicolumn{1}{|c|}{} & Rel. $L_2$ Err & NaN & NaN & 2.470E-1$\pm$1.117E-2 \\ \hline\hline
\multicolumn{2}{|c|}{} & GRU2 & LSTM2 & Mamba2 \\ \hline
\multicolumn{2}{|c|}{Layer Num} & 1 & 1 & 1 \\ \hline
\multicolumn{2}{|c|}{Hidden Size} & 64 & 64 & 32 \\ \hline
\multicolumn{2}{|c|}{Params} & 25345 & 33665 & 22593 \\ \hline
\multicolumn{1}{|c|}{\multirow{3}{*}{$T=1$}} & Memory & 3039MiB & 3257MiB & 2265MiB \\ \cline{2-5} 
\multicolumn{1}{|c|}{} &  & 75min & 61min & 116min \\ \cline{2-5} 
\multicolumn{1}{|c|}{} &  & NaN & 6.158E-1$\pm$7.201E-2 & \textbf{3.417E-2$\pm$4.156E-3} \\ \hline
\multicolumn{1}{|c|}{\multirow{3}{*}{$T=2$}} & Memory & 4511MiB & 4949MiB & 3143MiB \\ \cline{2-5} 
\multicolumn{1}{|c|}{} &  & 146min & 114min & 230min \\ \cline{2-5} 
\multicolumn{1}{|c|}{} &  & NaN & NaN & \textbf{1.230E-1$\pm$3.954E-3} \\ \hline
\multicolumn{1}{|c|}{\multirow{3}{*}{$T=3$}} & Memory & 5983MiB & 6637MiB & 4033MiB \\ \cline{2-5} 
\multicolumn{1}{|c|}{} &  & 215min & 169min & 346min \\ \cline{2-5} 
\multicolumn{1}{|c|}{} &  & NaN & NaN & \textbf{2.157E-1$\pm$1.643E-2} \\ \hline
\end{tabular}
\caption{Results for the Lorenz system where we train models to map from the random initial condition to the entire ODE trajectory over a long time introduced from \cite{li2021learning} corresponding to Section \ref{sec:Lorenz_init} in this paper. Some visualization results are  presented in Figure \ref{fig:Lorenz_init_1}.}
\label{tab:Lorenz_init}
\end{table}

\begin{figure}[htbp]
\centering
\includegraphics[width=0.24\linewidth]{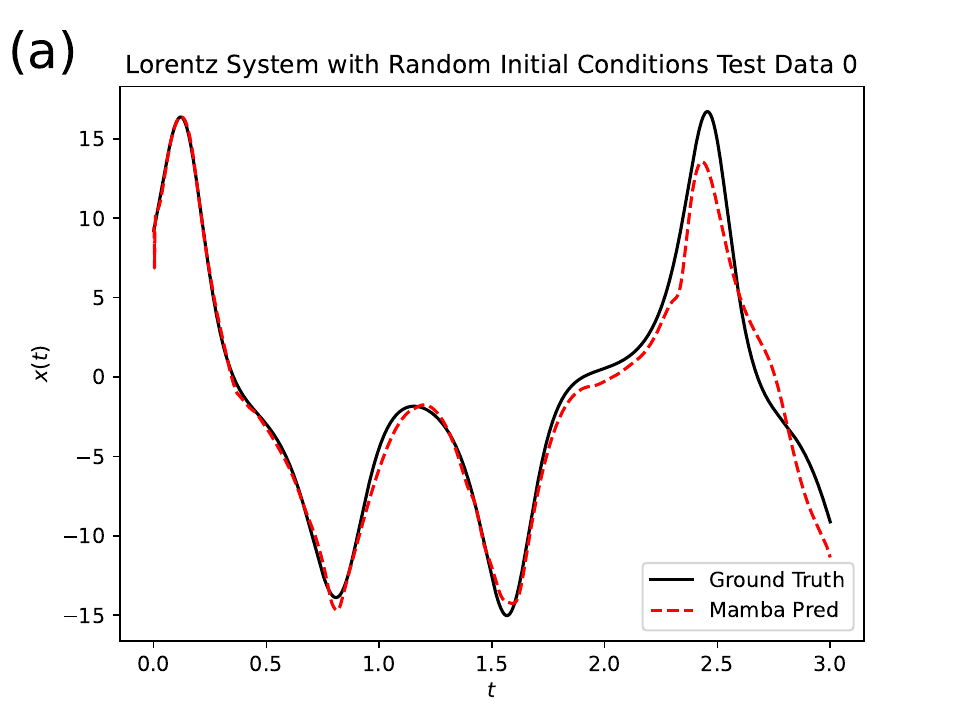}
\includegraphics[width=0.24\linewidth]{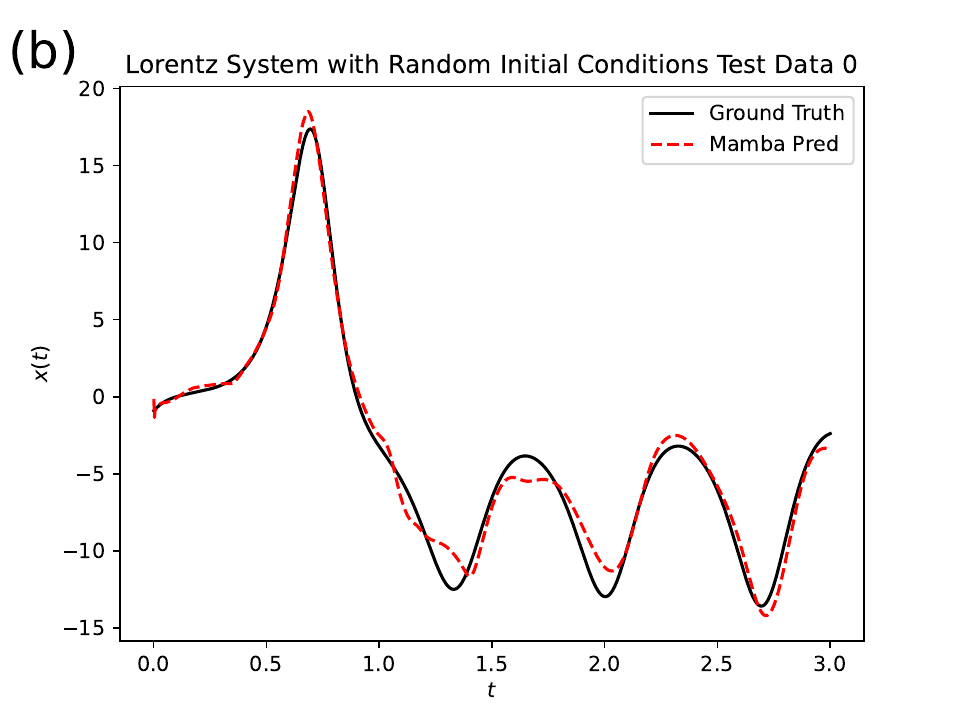}
\includegraphics[width=0.24\linewidth]{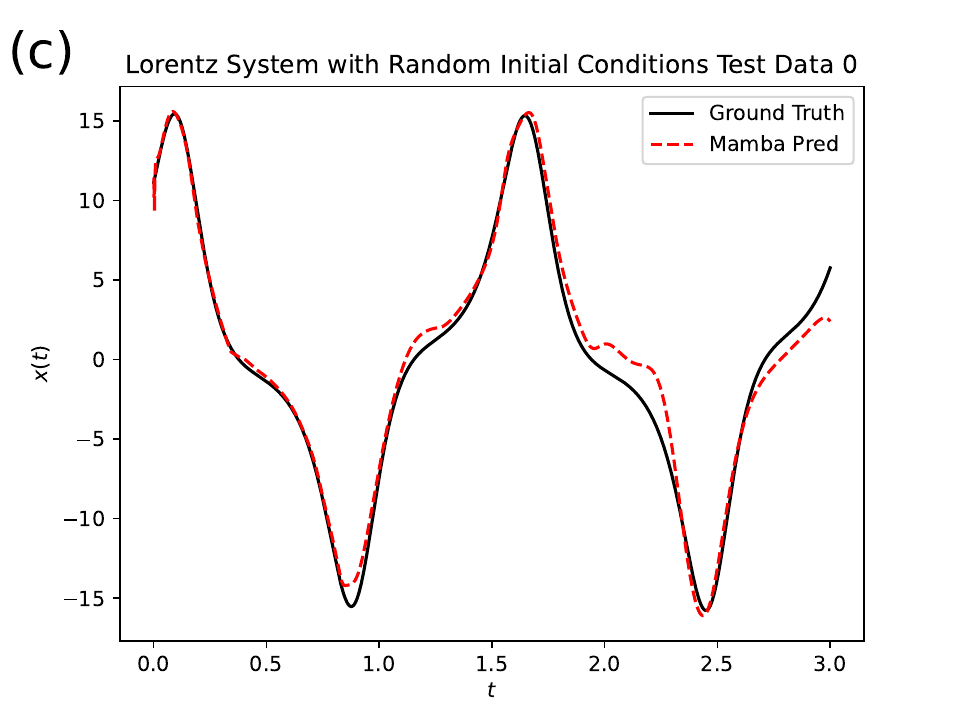}
\includegraphics[width=0.24\linewidth]{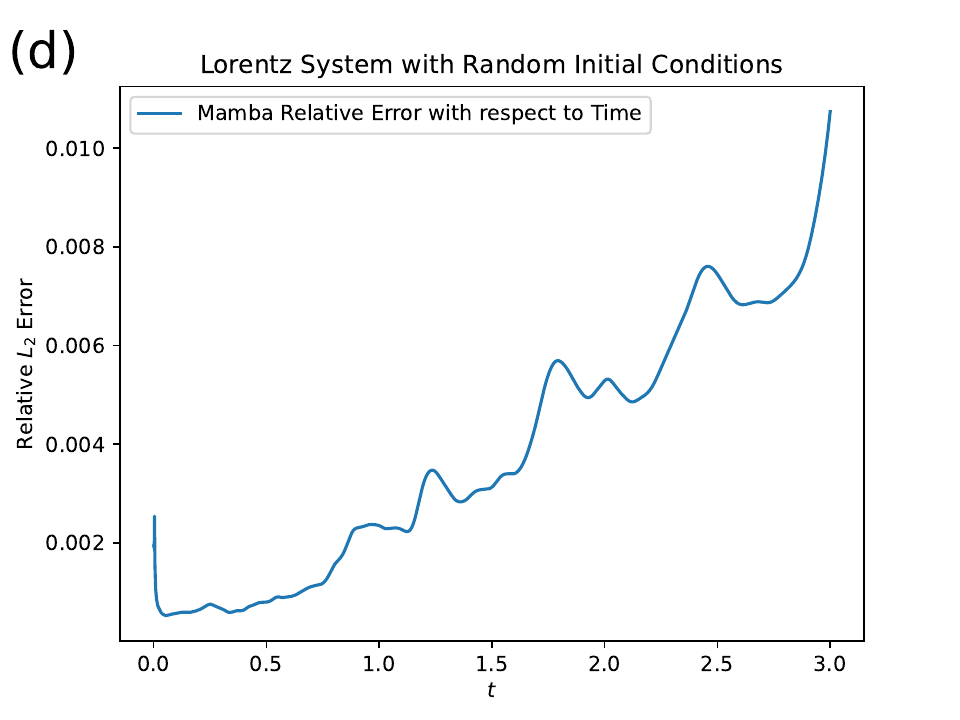}
\caption{Visualization of Mamba's prediction on the Lorenz system with random initial conditions introduced in \cite{li2021learning}, corresponding to Section \ref{sec:Lorenz_init} in this paper. The quantitative results are shown in Table \ref{tab:Lorenz_init}.
We visualize Mamba predictions and ground truth on test data numbers 0, 1, and 2 in subfigures (a), (b), and (c), respectively. 
Subfigure (d) visualizes Mamba's prediction relative $L_2$ error over the entire dataset concerning time.
The error is linear in time.}
\label{fig:Lorenz_init_1}
\end{figure}

Results are shown in Table \ref{tab:Lorenz_init}. 
In the chaotic Lorenz system with a terminal time equal to $T=2$ or $T=3$, we find that LSTM and GRU models of various sizes diverge to not-a-number (NaN) values during training due to their model instability.
Chaotic systems are known for their instability and sensitivity to minor changes in their initial values. A minor difference in initial values will lead to a considerable difference in ODE trajectories, especially during long-time integration. This, in turn, implies that the model has to be able to capture the subtle differences in initial values and react differently, which is a challenging task.
Hence, we observe that LSTM and GRU converge well in the $T=1$ case where the time is short, and the chaotic dynamics have not developed yet. However, as we reach longer times, LSTM and GRU diverge due to the chaotic dynamics, even though we train them using a decaying learning rate for stabilization. 
This reflects RNNs' inherent instability in lengthy sequence modeling (length = 1K, 2K, and 3K here) due to unstable gradient explosion even for its modern variants, namely LSTM and GRU~\cite{bengio1994learning, pascanu2013difficulty}.
In contrast, Mamba always converges in all systems, demonstrating its superior stability and robustness regardless of the long-time integration.

Figure \ref{fig:Lorenz_init_1} visualizes some of Mamba’s predictions and Mamba’s relative $L_2$ error over the entire dataset concerning time. We notice a linear increase in Mamba's error with growing time, demonstrating the challenge due to the chaotic Lorenz system in long-time integration.

More specifically, the rapid growth of error in the Mamba neural network model fitting the Lorenz ODE solution can be explained using the concept of Lyapunov exponents. The Lyapunov exponents measure the rates at which trajectories in the system diverge or converge. For the Lorenz system, the presence of a positive Lyapunov exponent indicates chaos, meaning that small differences in initial conditions can lead to exponentially growing deviations over time.
Wolf et al. \cite{wolf1985determining} provided a method for calculating Lyapunov exponents from time series data, which is $\lambda_1 \approx 0.9$ for the chaotic Lorenz system. 
Intuitively, for a chaotic system with Lyapunov exponent $\lambda_1$, if its initial condition deviates by a small $\delta$, then the ODE solution at time $t$ may deviate up to $\exp(\lambda_1 t)$, which is an exponential increase in error. This further justifies the difficulty of learning the chaotic Lorenz system.

In Figure \ref{fig:Lorenz_init_1} (d), we also observe that the error is slightly larger at the beginning. Initially, the only available information is the initial condition, which is quite sparse, and this makes it challenging for Mamba to effectively utilize this limited information to fit the ODE solution for the subsequent time step. However, as time progresses, the hidden state of Mamba gradually acquires more information, reducing the error and bringing the model's performance back on track. Nevertheless, the initial error remains relatively small compared to the error observed in the larger time.

\subsection{Quantifying the Extrapolation Error}\label{sec:quantify_extrapolation_error_zhu2023reliable}

We have extensively tested Mamba's extraordinary out-of-distribution (OOD) extrapolation capability in (1) OOD operator learning following Cao et al. \cite{cao2023lno} in Section \ref{sec:lnoode}, (2) extrapolation to long-time integration problems while only training on shorter time \ref{sec:extrapolation_long_time_1D_DS_deeponet}, (3) extrapolation to treatment schedule parameters in the PK-PD model in Section \ref{sec:pk_pd_model_extrapolation}. 
In this section, we follow Zhu et al. \cite{zhu2023reliable} to quantify further various models' extrapolation errors under different train and test input function distributions.

Regarding the ODE operators, we test the antiderivative and the gravity pendulum operators in Section \ref{sec:1D_DS_DeepONet}.
Following Zhu et al. \cite{zhu2023reliable}, the input functions $u(t)$ are randomly drawn from a mean zero Gaussian random field (GRF) with a length scale parameter $l$, i.e., the input functions follow the distribution $\mathcal{GP}\left(0, k_l(x, y)\right)$ where $k_l(x, y) = \exp\left(-\frac{(x - y)^2}{2 l^2}\right)$ is the RBF kernel with length scale $l$ and $\mathcal{GP}$ denotes a Gaussian process. A smaller correlation length scale $l$ leads to less smooth functions, while a larger correlation length scale $l$ leads to smoother functions.

We use different length scales for training (denoted $l_{train}$) and testing (denoted $l_{test}$) functions to construct the OOD generalization setting.
We define Ex+ as extrapolation with $l_{train} > l_{test}$, which is an extrapolation from smoother inputs to less smooth functions. 
Ex+ is regarded as more difficult in Zhu et al. \cite{zhu2023reliable} since it requires OOD generalization from ``good" smooth functions during training to ``rougher" functions during testing. 
Similarly, we define Ex- as extrapolation with $l_{train} < l_{test}$.
\begin{itemize}
\item Ex+: We train on $l_{train} = 0.1$ and test on $l_{test} \in \{0.1, 0.2, \cdots, 1.0\}$.
\item Ex-: We train on $l_{train} = 1.0$ and test on $l_{test} \in \{0.1, 0.2, \cdots, 1.0\}$.
\end{itemize}

{\color{green!50!black}We generate 10K training and validation data under $l_{train}$ and 10K testing data under each $l_{test}$. The validation is conducted under an IID set following the same $l_{train}$. In real-world applications, we cannot access the testing distribution in advance and must use the same source for validation.
We keep all settings, e.g., implementation details, training, and validation procedures, the same as those in Section \ref{sec:1D_DS_DeepONet} except the train and test input function distributions.
We finally obtain the best hyperparameters and model structures the same as in Section \ref{sec:1D_DS_DeepONet}.}

\begin{figure}[htbp]
\centering
\includegraphics[width=0.32\linewidth]{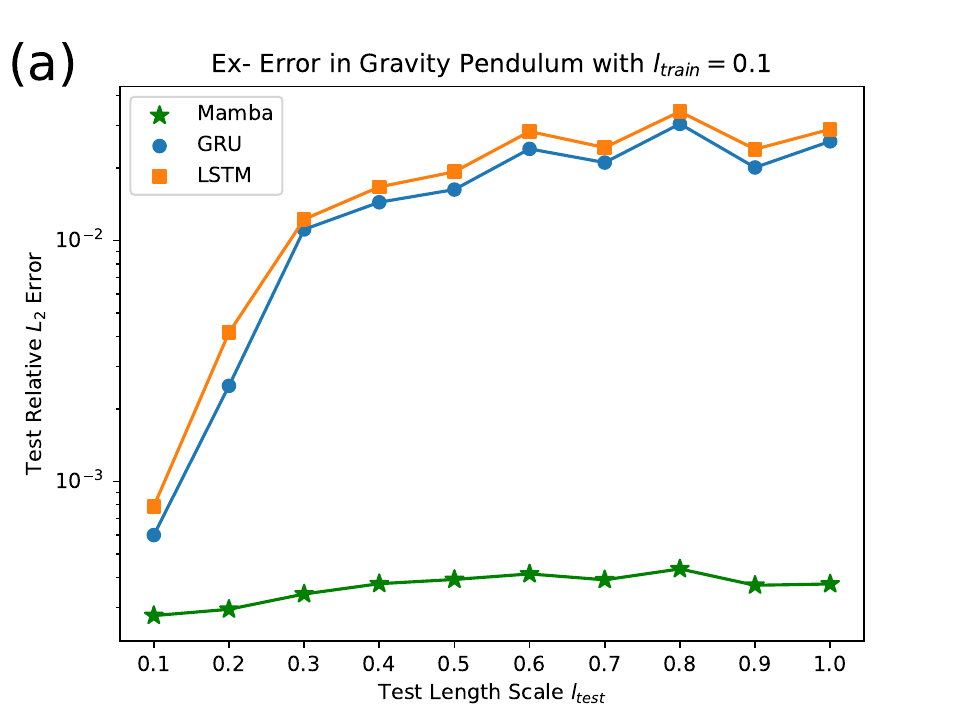}
\includegraphics[width=0.32\linewidth]{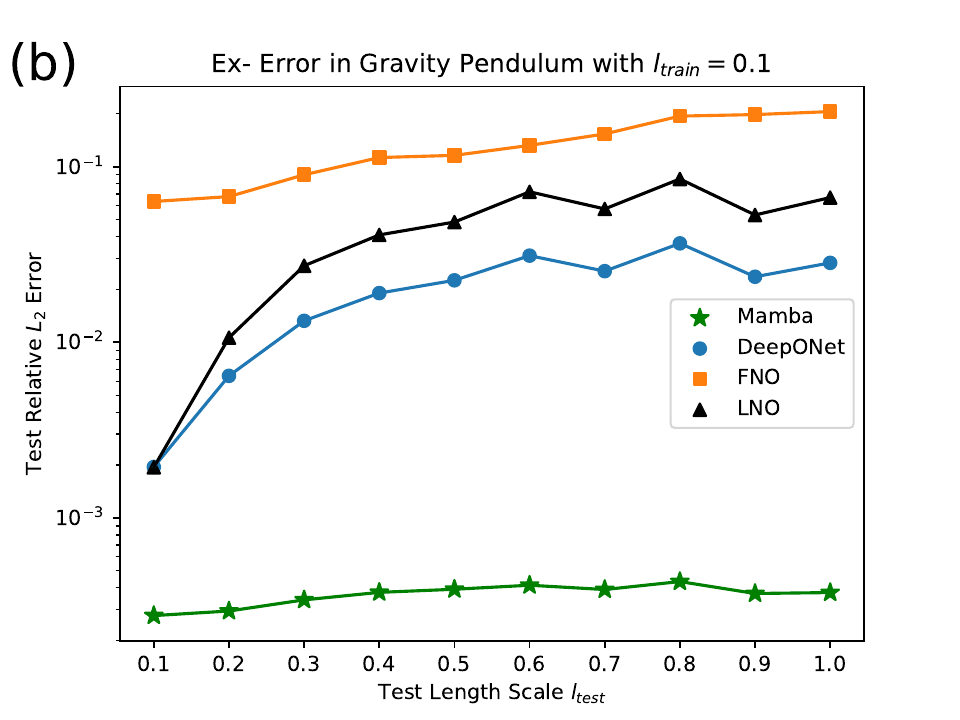}
\includegraphics[width=0.32\linewidth]{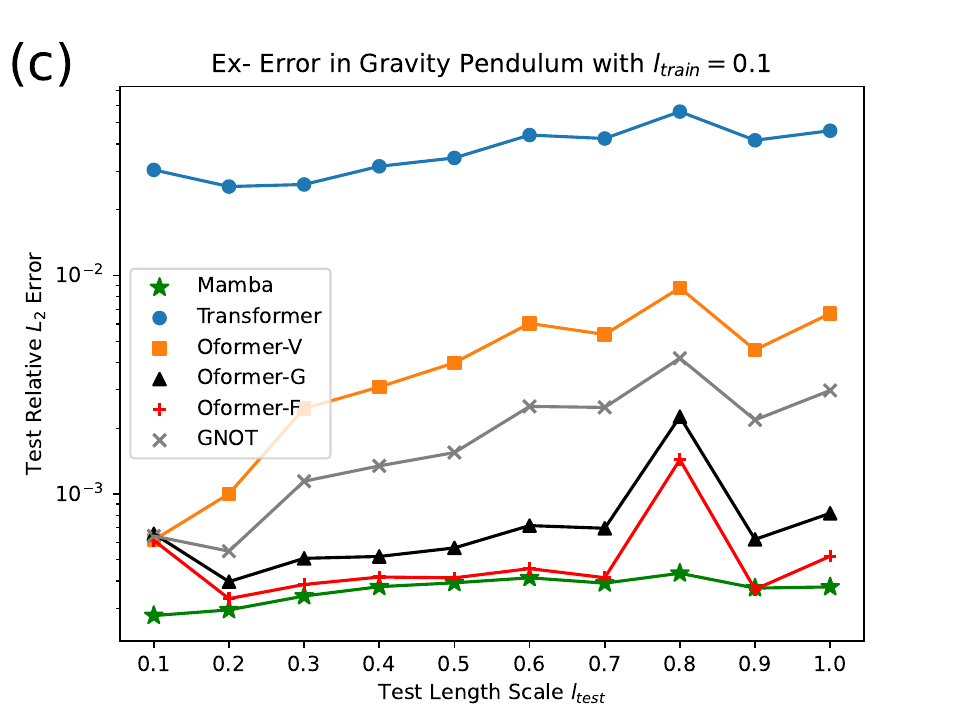}
\caption{Gravity pendulum Ex- setting with $l_{train} = 0.1$ and $l_{test} \in \{0.1,0.2,\cdots,1.0\}$ to quantify the extrapolation error following Zhu et al. \cite{zhu2023reliable}, corresponding to Section \ref{sec:quantify_extrapolation_error_zhu2023reliable}. The raw data is from Table \ref{tab:Ex-_Pend}.
(a) RNN models (GRU and LSTM) versus Mamba.
(b) Neural operators (DeepONet, FNO, and LNO) versus Mamba.
(c) Transformer models (Transformer, Oformers, GNOT) versus Mamba.
}
\label{fig:Ex-_Pend}
\end{figure}

\begin{figure}[htbp]
\centering
\includegraphics[width=0.32\linewidth]{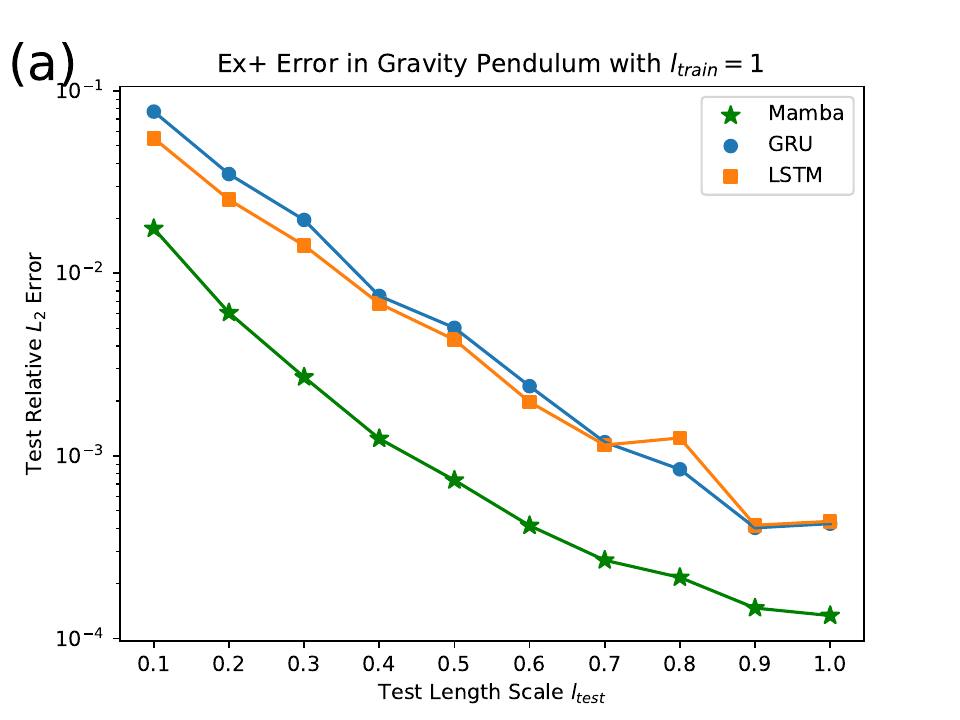}
\includegraphics[width=0.32\linewidth]{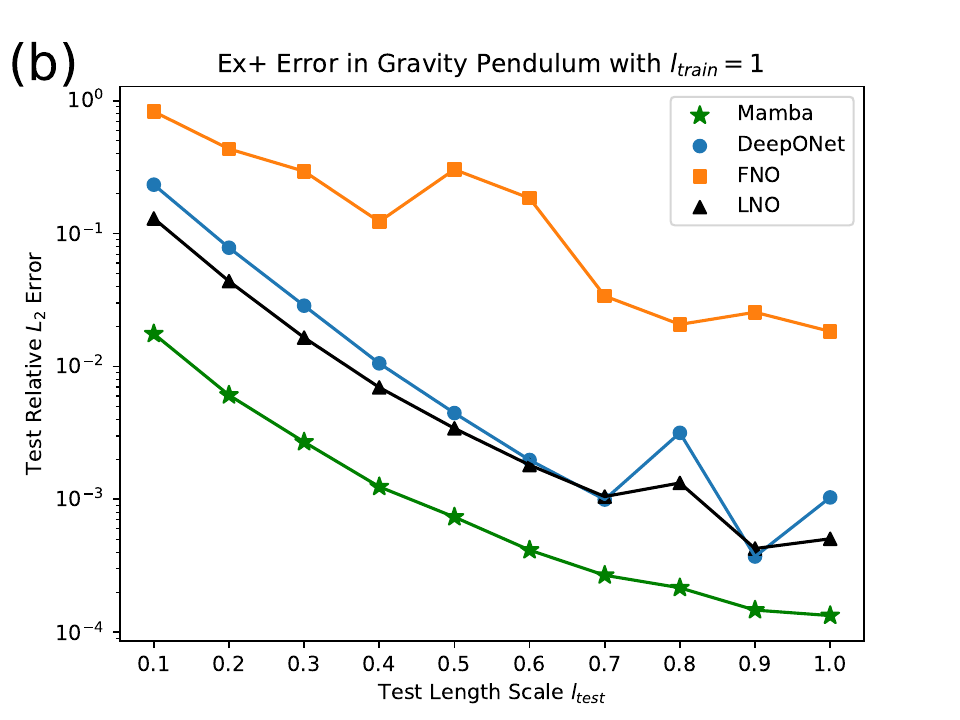}
\includegraphics[width=0.32\linewidth]{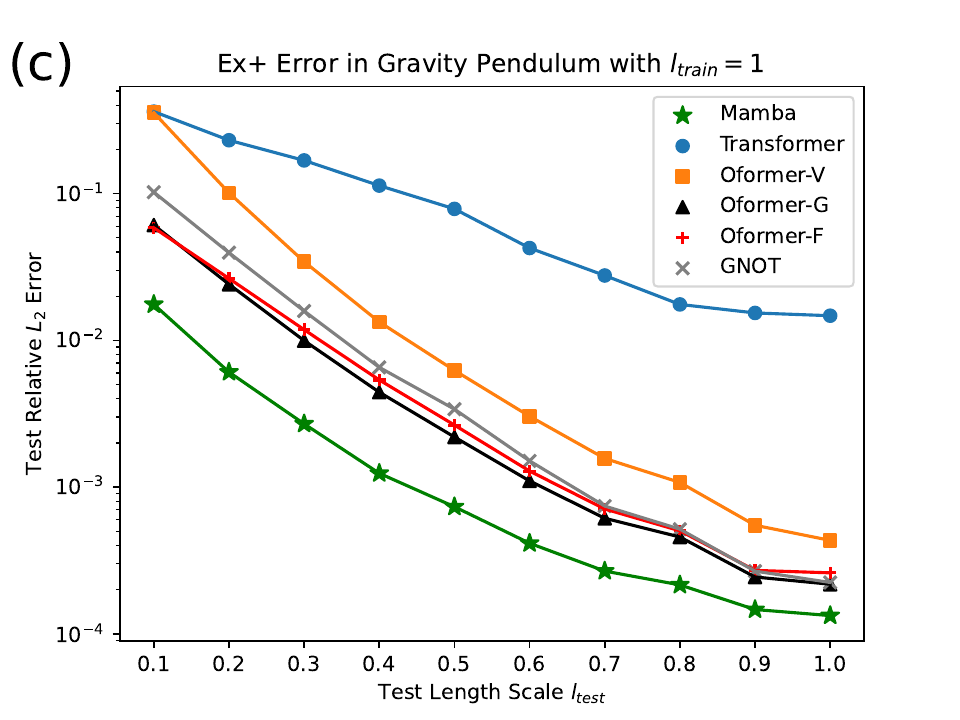}
\caption{Gravity pendulum Ex+ setting with $l_{train} = 1$ and $l_{test} \in \{0.1,0.2,\cdots,1.0\}$ to quantify the extrapolation error following Zhu et al. \cite{zhu2023reliable}, corresponding to Section \ref{sec:quantify_extrapolation_error_zhu2023reliable}. The raw data is from Table \ref{tab:Ex+_Pend}.
(a) RNN models (GRU and LSTM) versus Mamba.
(b) Neural operators (DeepONet, FNO, and LNO) versus Mamba.
(c) Transformer models (Transformer, Oformers, GNOT) versus Mamba.
}
\label{fig:Ex+_Pend}
\end{figure}

\begin{figure}[htbp]
\centering
\includegraphics[width=0.32\linewidth]{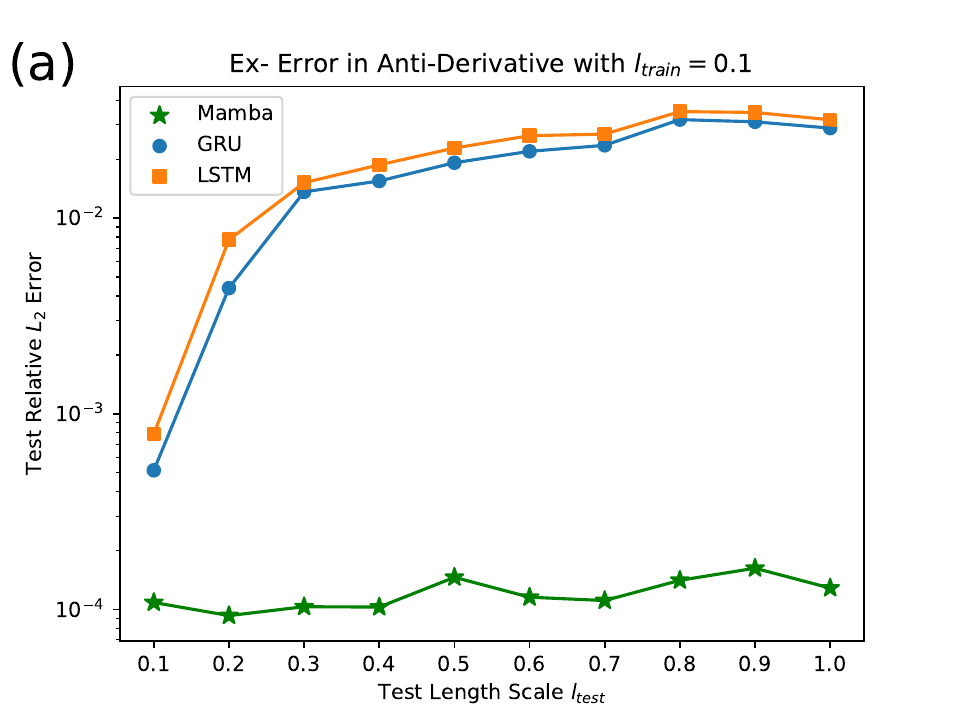}
\includegraphics[width=0.32\linewidth]{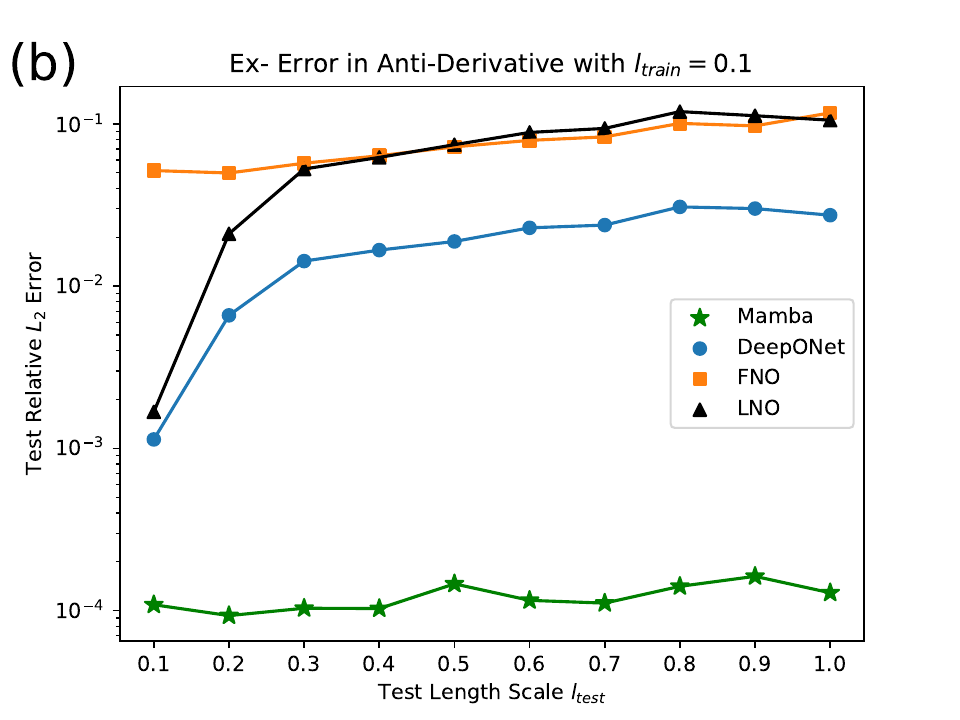}
\includegraphics[width=0.32\linewidth]{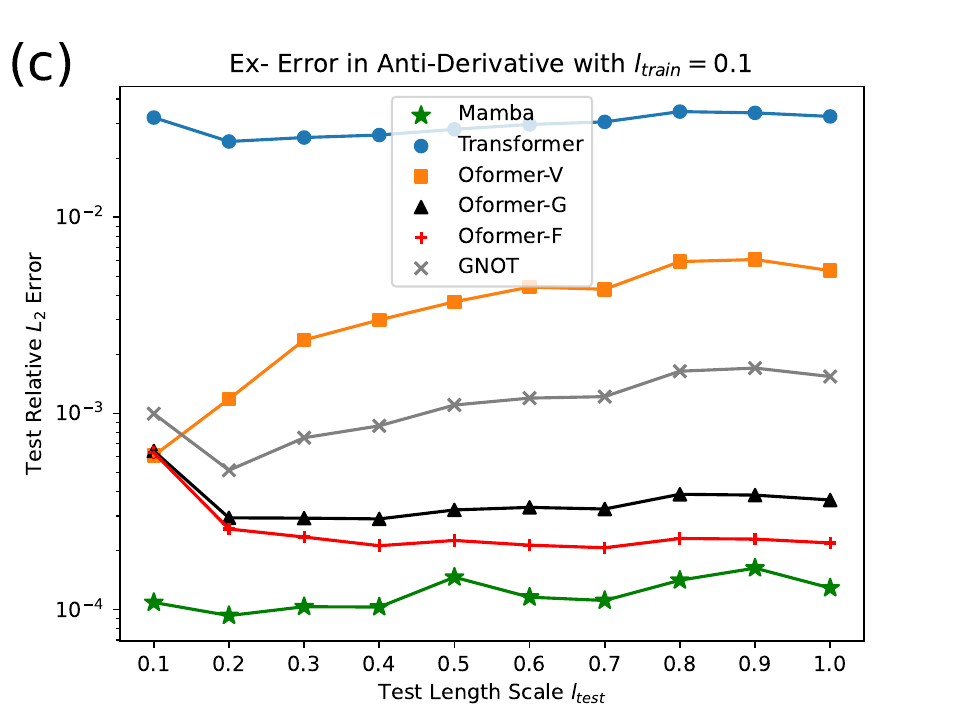}
\caption{Anti-derivative operator Ex- setting with $l_{train} = 0.1$ and $l_{test} \in \{0.1,0.2,\cdots,1.0\}$ to quantify the extrapolation error following Zhu et al. \cite{zhu2023reliable}, corresponding to Section \ref{sec:quantify_extrapolation_error_zhu2023reliable}. The raw data is from Table \ref{tab:Ex-_AD}.
(a) RNN models (GRU and LSTM) versus Mamba.
(b) Neural operators (DeepONet, FNO, and LNO) versus Mamba.
(c) Transformer models (Transformer, Oformers, GNOT) versus Mamba.
}
\label{fig:Ex-_AD}
\end{figure}

\begin{figure}[htbp]
\centering
\includegraphics[width=0.32\linewidth]{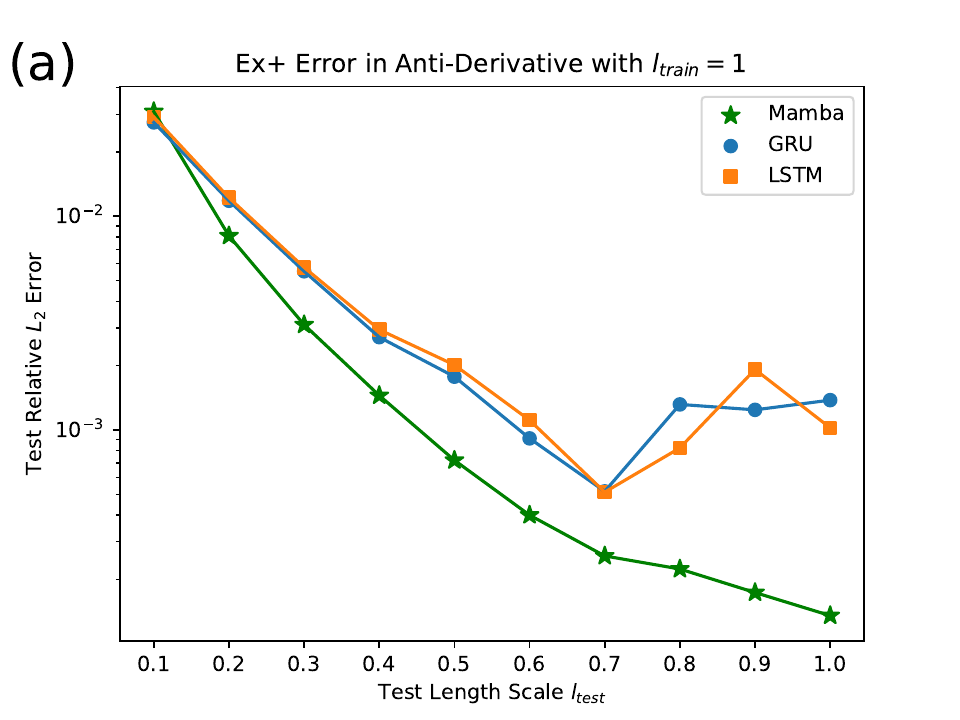}
\includegraphics[width=0.32\linewidth]{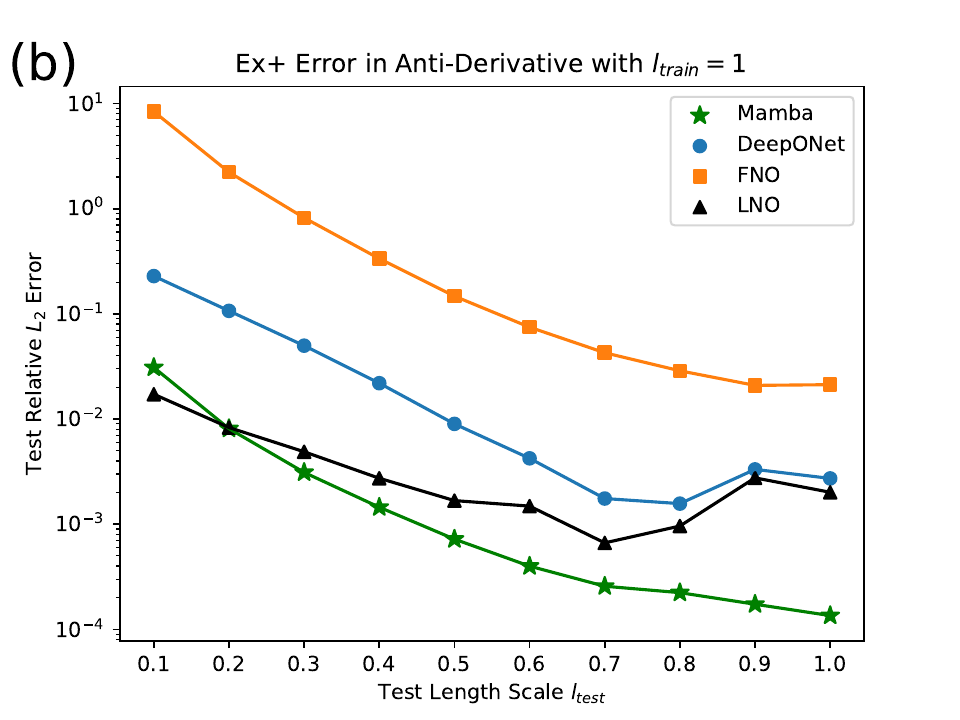}
\includegraphics[width=0.32\linewidth]{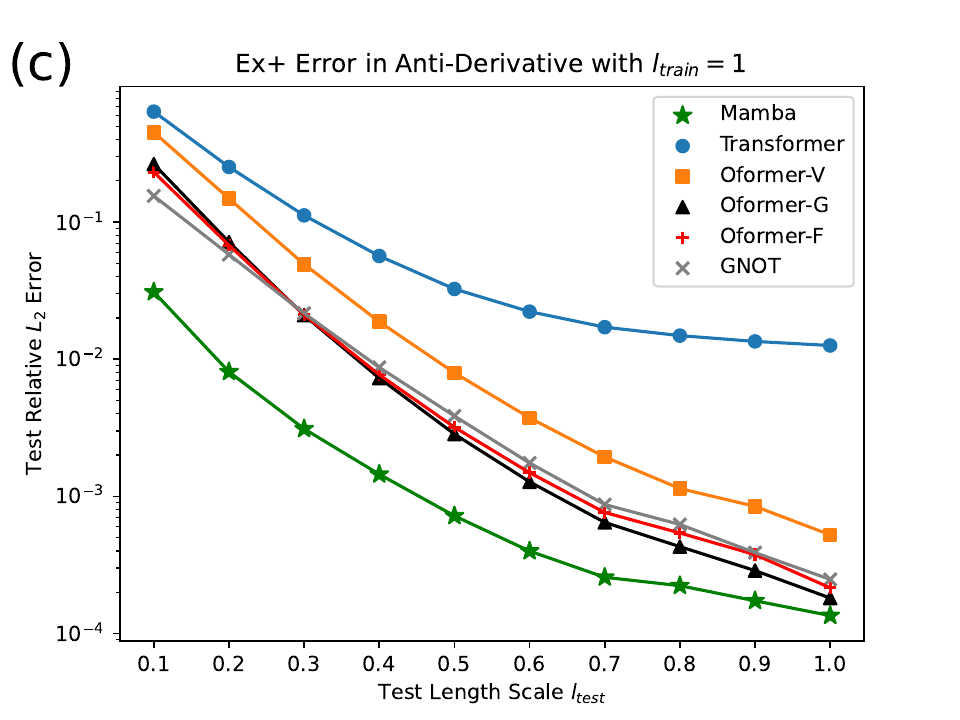}
\caption{Anti-derivative operator Ex+ setting with $l_{train} = 1$ and $l_{test} \in \{0.1,0.2,\cdots,1.0\}$ to quantify the extrapolation error following Zhu et al. \cite{zhu2023reliable}, corresponding to Section \ref{sec:quantify_extrapolation_error_zhu2023reliable}. The raw data is from Table \ref{tab:Ex+_AD}.
(a) RNN models (GRU and LSTM) versus Mamba.
(b) Neural operators (DeepONet, FNO, and LNO) versus Mamba.
(c) Transformer models (Transformer, Oformers, GNOT) versus Mamba.
}
\label{fig:Ex+_AD}
\end{figure}

The detailed quantitative results are shown in Table \ref{tab:Ex-_Pend} (gravity pendulum Ex-), Table \ref{tab:Ex+_Pend} (gravity pendulum Ex+), Table \ref{tab:Ex-_AD} (anti-derivative Ex-), Table \ref{tab:Ex+_AD} (anti-derivative Ex+) in  \ref{appendix:quantify_extrapolation_error_zhu2023reliable}.
Their visualization is shown in Figure \ref{fig:Ex-_Pend} (gravity pendulum Ex-), Figure \ref{fig:Ex+_Pend} (gravity pendulum Ex+), Figure \ref{fig:Ex-_AD} (anti-derivative Ex-), Figure \ref{fig:Ex+_AD} (anti-derivative Ex+).
Each figure shows the visualized results for one of the four settings, gravity pendulum or anti-derivative operator in the Ex- or the Ex+ setting.
Furthermore, in each figure, we show Mamba's comparison with RNNs, including GRU and LSTM, in subfigure (a), with neural operators (DeepONet, FNO, and LNO) in (b), and with transformers (Transformer, Oformers, and GNOT) in (c).

Regarding the pendulum problem, Mamba is the best in both Ex- problems (Figure \ref{fig:Ex-_Pend} and Table \ref{tab:Ex-_Pend}) and Ex+ (Figure \ref{fig:Ex+_Pend} and Table \ref{tab:Ex+_Pend}) in general.
More concretely, in the Ex- setting (Figure \ref{fig:Ex-_Pend} and Table \ref{tab:Ex-_Pend}), Mamba's error remains stable under the test distribution shift. Although certain transformers (Oformer-G and Oformer-F) also exhibit stability against test distribution shift, their error is larger than Mamba. Other models, including RNNs and neural operators, exhibit increases in errors when the test distribution shift is more significant, i.e., their error grows larger as $l_{test} = 0.1$ gradually moves to $l_{test} = 1$.
Meanwhile, regarding the Ex+ setting of gravity pendulum (Figure \ref{fig:Ex+_Pend} and Table \ref{tab:Ex+_Pend}), all models' errors grow with test distribution shift as $l_{test} = 1$ gradually moves to $l_{test} = 0.1$ where the training set length scale is $l_{train} = 1$. However, Mamba exhibits the slowest error growth and consistently performs the best.

Regarding the anti-derivative operator, Mamba is the best in the Ex- setting (Figure \ref{fig:Ex-_AD} and Table \ref{tab:Ex-_AD}), and its Ex- error remains stable. 
We also observe that Oformers \cite{litransformer} can exhibit stability against test distribution shift but perform worse than Mamba in general.
Moreover, other models, including RNNs and neural operators, exhibit increases in errors when the test distribution shift is more significant, i.e., their error grows larger as $l_{test} = 0.1$ gradually moves to $l_{test} = 1$.
This is similar to the Ex- case in the gravity pendulum case.
Regarding the anti-derivative operator's Ex+ setting (Figure \ref{fig:Ex+_AD} and Table \ref{tab:Ex+_AD}), Mamba is good in the majority of test cases. Mamba is only beaten in the case with $l_{train} = 1$ and $l_{test} = 0.1$, although not by much, and it is close to the best model. Quantitatively, Mamba's relative $L_2$ error for the anti-derivative operator's Ex+ setting with $l_{train} = 1$ and $l_{test} = 0.1$ is 3.091E-2, while the best model LNO's is 1.721E-2.

These results further demonstrate Mamba's outstanding OOD generalization capability under quantitative settings following Zhu et al. \cite{zhu2023reliable}.
Mamba performs the best in most cases and is stable across all settings.
We also demonstrate how other various models, including RNNs, neural operators, and transformers, generalize under test distribution shift.

\subsection{Real-World Application: Quantitative Systems Pharmacology}\label{sec:pk_pd_model}

In order to assess the efficacy of predictive models in a real-world setting, we evaluate the antitumor drug effect. A pharmacokinetic-pharmacodynamic (PK-PD) model is employed to establish a link between the administration regimen of a candidate drug and tumor growth dynamics. Precision medicine in oncology, an emerging approach to tumor treatment, is predicated on the ability to predict pharmacodynamic outcomes, which in turn allows for the direct assessment of the therapeutic potential of compounds in cancer therapy. Despite the considerable investment of time and resources in the evaluation of drugs for desirable pharmacokinetic characteristics, these efforts do not always result in optimal clinical outcomes. In the context of virtual compound screening, direct observation of clinical outcomes in humans is not a viable option, introducing an element of uncertainty and extending the timeline for drug development. The incorporation of patient-specific attributes, including sex, age, hematological parameters, and other biomarkers, enhances the accuracy of predictions in cancer therapy. The impact of each parameter can be masked, and if they are deemed to be effective in influencing the outcome, they can be incorporated into the patient-specific PD model parameters. By integrating deep sequence models into PK-PD of tumor growth dynamic models, we developed predictive models for drug efficacy in tumor size, which yield three key benefits: 
1. Early prediction of PD outcomes in drug discovery, streamlining the compound screening process.
2. Surrogate models for uncertainty quantification, allowing rapid model evaluation.
3. Assessment of the impact of drug and treatment schedule changes or combination therapies on tumor size.

\subsubsection{Model and Data Generation}\label{sec:PKPD_data_generation_method}
Simeoni et al. \cite{simeoni2004predictive} introduced a streamlined PK-PD model to simplify the complex dynamics of tumor growth, making it more applicable for the preclinical development of oncology drugs. This model, formulated with a system of ODEs, connects the drug dosing regimen to the tumor growth dynamics observed in animal studies. The model differentiates between untreated and treated animals; in untreated animals, tumor growth follows an initial exponential phase and then a linear phase, whereas, in treated animals, the growth rate decreases in relation to the drug concentration and the number of proliferating tumor cells. The differential equation for the unperturbed growth of the tumor mass \( w(t) \) is given by:
\begin{align}
\frac{dw(t)}{dt} = \frac{\lambda_0 \cdot w(t)}{\left[1 + \left(\frac{\lambda_0}{\lambda_1} \cdot w(t)\right)^{\Psi}\right]^{1/\Psi}}, \quad w(0) = w_0.
\end{align}\\
For sufficiently large values of \( \Psi \), this equation accurately replicates the switching behavior observed in tumor growth dynamics. Specifically, when the tumor mass \( w(t) \) is below a threshold \( w_{\text{th}} \), the ratio \( \frac{\lambda_0}{\lambda_1} \cdot w(t) \) in the denominator is negligible compared to 1. This results in exponential growth characterized by \( \lambda_0 \cdot w(t) \). Conversely, as \( w(t) \) exceeds \( w_{\text{th}} \), the value of 1 in the denominator becomes negligible, transitioning the growth rate to a linear model defined by \( \lambda_1 \). Practically, setting \( \Psi \) to 20 or higher ensures a clear transition from exponential (first-order) to linear (zero-order) growth, closely mirroring the sharp switch observed in the original model with switching systems. A novel feature of the model is the inclusion of a transit compartment system to simulate delayed cell death processes common in many signal transduction pathways. The pharmacodynamic parameters are intricately linked to tumor growth characteristics, anticancer drug efficacy, and tumor cell death kinetics. These parameters facilitate comparisons of compound effectiveness and exploration of variations in tumor cell death mechanisms. The overall model is described by the following system of differential equations:

\begin{align}\label{eq:pkpd}
\frac{dx_1(t)}{dt} &= \frac{\lambda_0 \cdot x_1(t)}{\left[1 + \left(\frac{\lambda_0}{\lambda_1} \cdot w(t)\right)^{\Psi}\right]^{1/\Psi}} - k_2 \cdot c(t) \cdot x_1(t), \\
\frac{dx_2(t)}{dt} &= k_2 \cdot c(t) \cdot x_1(t) - k_1 \cdot x_2(t), \\
\frac{dx_3(t)}{dt} &= k_1 \cdot (x_2(t) - x_3(t)), \\
\frac{dx_4(t)}{dt} &= k_1 \cdot (x_3(t) - x_4(t)), \\
w(t) &= x_1(t) + x_2(t) + x_3(t) + x_4(t),
\end{align}\\
where \( x_1(t) \) represents the population of proliferating cells, and \( x_2(t) \) to \( x_4(t) \) represent cells at various stages of damage leading to death. The total tumor mass \( w(t) \) includes all cellular compartments, and \( c(t) \) is the plasma concentration of the anticancer agent. The drug concentration \( c(t) \) in the model is derived from a two-compartment PK model with known parameters, as detailed in the data presented in \cite{simeoni2004predictive}, which was utilized in this study.

The governing equations of the two-compartment PK model are described by the following system of linear ODEs:
\begin{align}
\frac{dA_1(t)}{dt} &= -k_{12} A_1(t) + k_{21} A_2(t) - k_{10} A_1(t) + I_1(t), \\
\frac{dA_2(t)}{dt} &= k_{12} A_1(t) - k_{21} A_2(t),\\
c(t) &= \frac{A_1(t)}{V_1}
\end{align}
Here, $A_1(t)$ and $A_2(t)$ represent the mass or molar amounts in the respective compartments. The constants $k_{ij}$ govern the mass transfer between the compartments and the elimination from each, where the notation $k_{12}$ denotes the rate of transfer from compartment 1 to compartment 2, and $k_{10}$ denotes elimination from compartment 1, and so forth. The units for all $k_{ij}$ rate constants are $(1/\text{time})$. The input rates $I_1(t)$ may be zero, constant, or time-dependent, indicating the amount or dose of administered drug. Initial conditions for the amounts in each compartment, $A_1(0)$ and $A_2(0)$, must also be specified. For the concentration of the drug in the PD model, we need to divide $A_1(t)$ by $V_1$, which is the volume of distribution in the first compartment.

In this work, we aim to learn the mapping from $I_1(t)$ in the PK model, representing the drug dose and administration schedule in a multi-dose regimen, to the patient-specific function $w(t)$, which reflects tumor weight based on individual patient PD model parameters $k_1$ and $k_2$.  These parameters indicate disparate responses to treatment across different species.

\textbf{Data}. The objective of the proposed compartmental model is to generate data for a defined population of mice, which can be explained within a specified range of PD parameters. Each set of parameters ($k_1$ and $k_2$) represents a distinct individual, indicating that the response to a given treatment may vary based on the presence or absence of known or unknown biomarkers. The PK  parameters obtained following intravenous administration of paclitaxel are as follows: $V_1 = 0.81 \, \text{L/kg}$, $k_{10} = 0.868 \, \text{h}^{-1}$, $k_{12} = 0.0060 \, \text{h}^{-1}$, and $k_{21} = 0.0838 \, \text{h}^{-1}$.
Synthetic data was generated for the purpose of training and testing the model, with the paclitaxel PK parameters and treatment schedule selected from a discrete random uniform distribution. The administration dose is selected from the range of 20 to 45 mg/kg, with increments of 5 mg/kg. The start day of the administration is selected from the range of 1 to 13 days, and the intervals for a 3-dose schedule are chosen from the range of 2 to 6 days (For example, 30 mg/kg is administered every 4 days for 3 doses starting from day 10). With regards to the PD parameters, $k_1$ has been selected from a random normal distribution with $k_1 = 1 \pm 0.5 \, \text{day}^{-1}$, while $k_2$ has been chosen from  $6 \times 10^{-4} \pm 2 \times 10^{-4} \, \text{ng}^{-1} \cdot \text{ml} \cdot \text{day}^{-1}$ distribution. The parameters related to unperturbed tumor growth have been maintained at constant values, with $w_0 = 0.05 \, \text{g}$, $\Psi = 20$, $\lambda_1 = 0.273 \, \text{day}^{-1}$, and $\lambda_2 = 0.814 \, \text{g} \cdot \text{day}^{-1}$.

We generated synthetic data by sampling the solutions of the system of ODEs, obtained by solving the forward problem using numerical methods. Each species' data was generated using its unique PK and PD models. For the PK model, which involves a multi-dose administration regimen, we employed the explicit Euler forward method. The PD model was solved using the \texttt{odeint} function from the \texttt{scipy.integrate} library. The \texttt{odeint} function utilizes the LSODA algorithm, which automatically switches between non-stiff and stiff ODE solvers. For non-stiff problems, LSODA applies an Adams-Bashforth-Moulton method (a multi-step method). For stiff problems, it resorts to a Backward Differentiation Formula (BDF) method, an implicit method. The PD system appears to be mildly stiff, making LSODA's adaptive approach particularly suitable for this problem.

\subsubsection{Learning the PK-PD Model's Operator}\label{sec:pk_pd_model_1}
We generated a dataset containing 5000 (5K) species for both the PK and PD models based on the procedure in Section \ref{sec:PKPD_data_generation_method}. The dataset comprises the dose amount and dose schedule as inputs (represented as $I_1(t)$ in the PK model), the species-specific PD model parameters $k_1$ and $k_2$, as well as $w(t)$, which is the model output. Upon feeding the model with the values of $k_1$, $k_2$, and $I_1(t)$, the resulting output is the tumor size in grams. It is observed that different dynamics and tumor sizes are exhibited by different species or treatment schedules.
{\color{green!50!black} We divide the dataset into 4K training and 1K testing. During validation and model hyperparameter selection, we leave 20\% of the training data as validation and train on the rest 80\%. After the model hyperparameter selection, we returned the validation data to training and optimized the model with a total of 4K training data. We append the temporal grid to the input data tensor for augmentation.}

% The dataset includes the dose amount and dose schedule as inputs (represented as $I_1(t)$ in the PK model), the species-specific PD model parameters $k_1$ and $k_2$, as well as the PD outputs.

Here are the implementation details. 
The training loss is MSE, and the test metric is the relative $L_2$ error.
We train all models for 10,000 epochs. 
In each epoch, we traverse the dataset with a minibatch size of 128.
We use the Adam optimizer \cite{kingma2014adam} with a 1e-3 initial learning rate, which decays linearly to zero at the end of training.
We keep the number of parameters of different models similar for a fair comparison. 
We run the code for five independent random seeds and report the average.

The model structure hyperparameter settings are summarized below.
{\color{green!50!black} Similar to Section \ref{sec:1D_DS_DeepONet}, we set a parameter count budget of approximately 10,000.
For all time series models, we search the model depth in $1\sim3$ and choose the corresponding hidden sizes to satisfy the budget. We found that shallow one-layer and wide models are better than deeper ones in speed and accuracy. 
LSTM and GRU are one-layer models with 32 hidden dimensions.
Regarding Mamba, we opt for a block/layer with skip connection and layer normalization, with 16 model dimensions and 16 intermediate dimensions.
We opt for a one-layer Transformer with 40 hidden and feedforward dimensions, and four heads.
For all three Oformers with vanilla, Galerkin, or Fourier attention, we opt for one-layer Oformers with 24 hidden and feedforward dimensions and four heads.
GNOT has one layer, 16 hidden feedforward dimensions, four heads, and two experts.
We choose the same one-layer comparable structures for all time series models for a fair comparison.
Regarding NOs, we keep DeepONet's trunk and branch nets as MLP with the same structure, whose depth is $2\sim5$. We find the deepest model is the best, i.e., DeepONet's trunk and branch nets are all five-layer networks with 24 hidden dimensions. We keep the same structure as in Section \ref{sec:1D_DS_DeepONet}, but since the input sequence is longer, DeepONet here has a larger parameter count.
We searched for the best FNO structure among $1\sim4$ layers and $8$ modes and obtained the best four-layer model with 16 hidden dimensions and eight modes.
We searched for the best LNO models with $8/16$ hidden dimensions and $4/8$ modes, and finally, we chose a one-layer LNO with eight hidden dimensions and four modes and another one-layer LNO with 16 hidden dimensions and eight modes. We also tried larger LNOs, but this small LNO model performs the best on the validation set. We report both LNO performances.
}

\begin{figure}[htbp]
\centering
\includegraphics[width=0.32\linewidth]{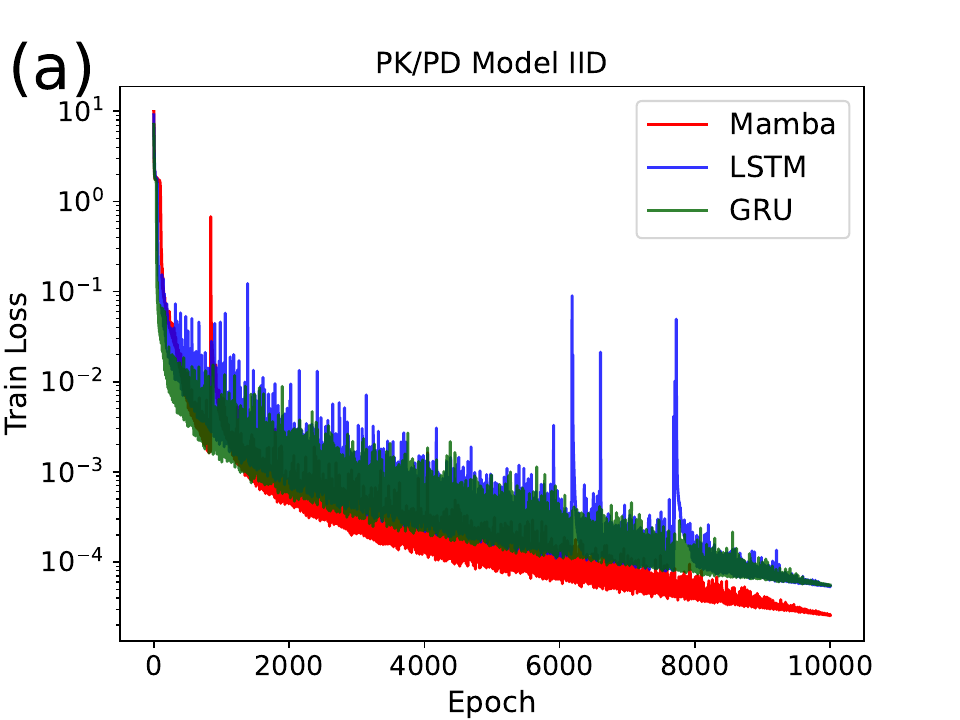}
\includegraphics[width=0.32\linewidth]{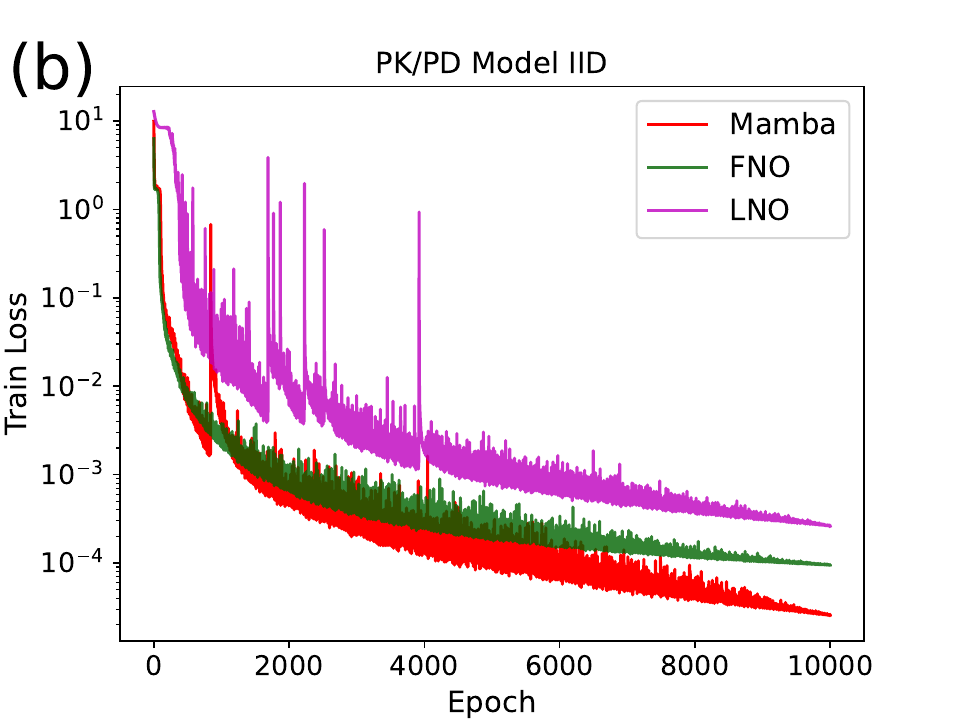}
\includegraphics[width=0.32\linewidth]{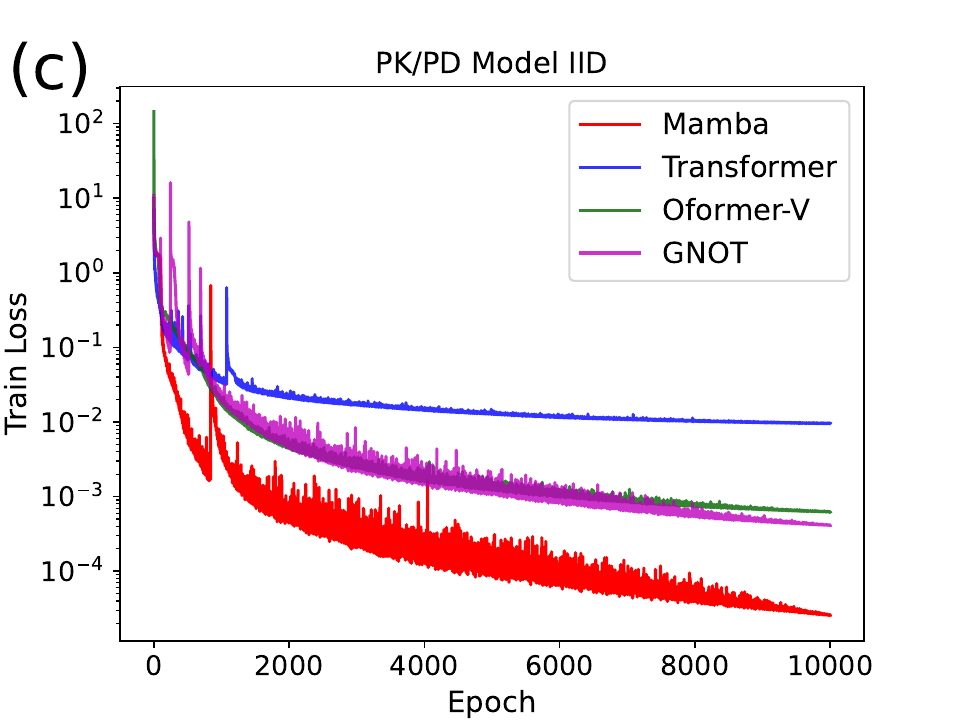}
\caption{{\color{green!50!black}Visualization of various models' loss trajectories for the PK/PD model with IID train-test data corresponding to Section \ref{sec:pk_pd_model_1} in this paper. The detailed quantitative results for this setting are presented in Table \ref{tab:pk_pd}. 
We visualize some inputs and Mamba's predictions in Figure \ref{fig:PKPD_1}.
We visualize Mamba's relative $L_2$ error concerning time in Figure \ref{fig:PKPD_2}.
Subfigure (a) is Mamba versus RNNs; subfigure (b) is Mamba versus NOs; and subfigure (c) is Mamba versus Transformers.}}
\label{fig:pk_pd_loss}
\end{figure}

\begin{table}[htbp]
\centering
\begin{tabular}{|c|c|c|c|c|}
\hline
Model & Params & Memory & Time & Relative $L_2$ Error \\ \hline
LSTM & 8641 & 1985MiB & 16min & \textcolor{red}{2.117E-3$\pm$1.512E-4} \\ \hline
GRU & 6529 & 1925MiB & 16min & \textcolor{blue}{2.125E-3$\pm$1.201E-4} \\ \hline
DeepONet & 38568 & 1425MiB & 15min & 4.732E-1$\pm$1.169E-1 \\ \hline
FNO & 9377 & 2021MiB & 30min & 2.917E-3$\pm$4.158E-5 \\ \hline
LNO (Large) & 6769 & 1991MiB & 24min & 8.269E-1$\pm$5.178E-2 \\ \hline
LNO (Small) & 1913 & 1935MiB & 22min & 5.295E-3$\pm$1.063E-3 \\ \hline
Transformer & 10321 & 2615MiB & 56min & 2.847E-2$\pm$5.340E-3 \\ \hline
Oformer-V & 11281 & 2469MiB & 49min & 7.708E-3$\pm$3.901E-4 \\ \hline
Oformer-G & 11233 & 1591MiB & 46min & 9.101E-3$\pm$7.357E-4 \\ \hline
Oformer-F & 11233 & 1591MiB & 46min & 8.129E-3$\pm$6.255E-4 \\ \hline
GNOT & 11235 & 1649MiB & 91min & 5.437E-3$\pm$9.589E-4 \\ \hline
Mamba & 9697 & 1707MiB & 19min & \textbf{1.685E-3$\pm$2.929E-4} \\ \hline
\end{tabular}
\caption{Results for the PK-PD model corresponding to Section \ref{sec:pk_pd_model_1}. \textbf{The best model is bold}, \textcolor{red}{the second best is red}, and \textcolor{blue}{the third best is blue}.
We visualize the loss trajectory in Figure \ref{fig:pk_pd_loss}.
We visualize some inputs and Mamba's predictions in Figure \ref{fig:PKPD_1}.
We visualize Mamba's relative $L_2$ error concerning time in Figure \ref{fig:PKPD_2}.}
\label{tab:pk_pd}
\end{table}

\begin{figure}[htbp]
\centering
\includegraphics[width=0.24\linewidth]{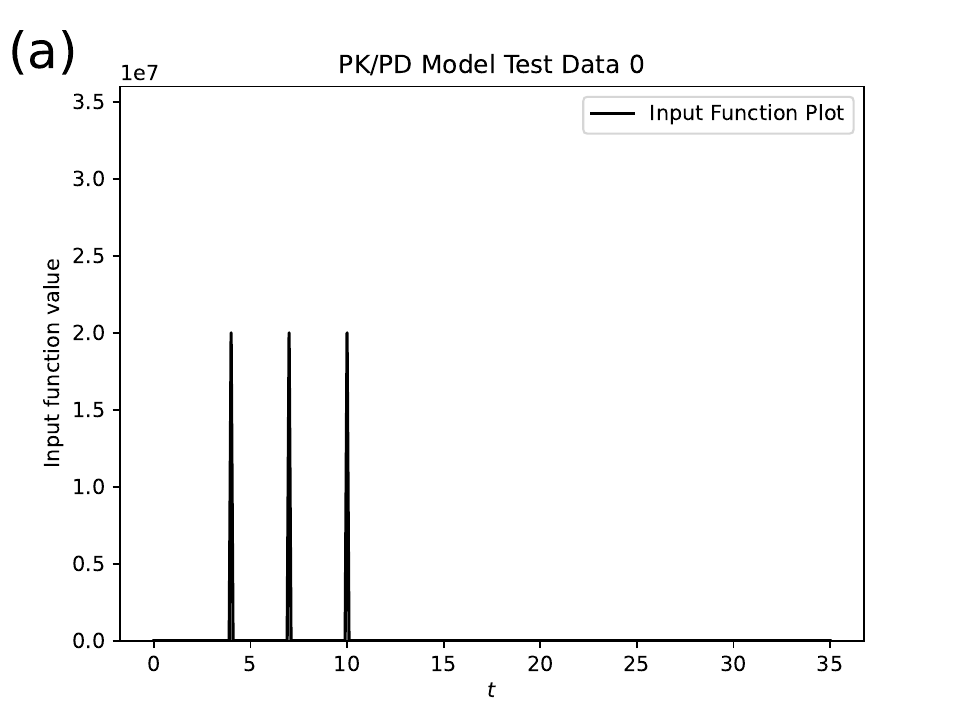}
\includegraphics[width=0.24\linewidth]{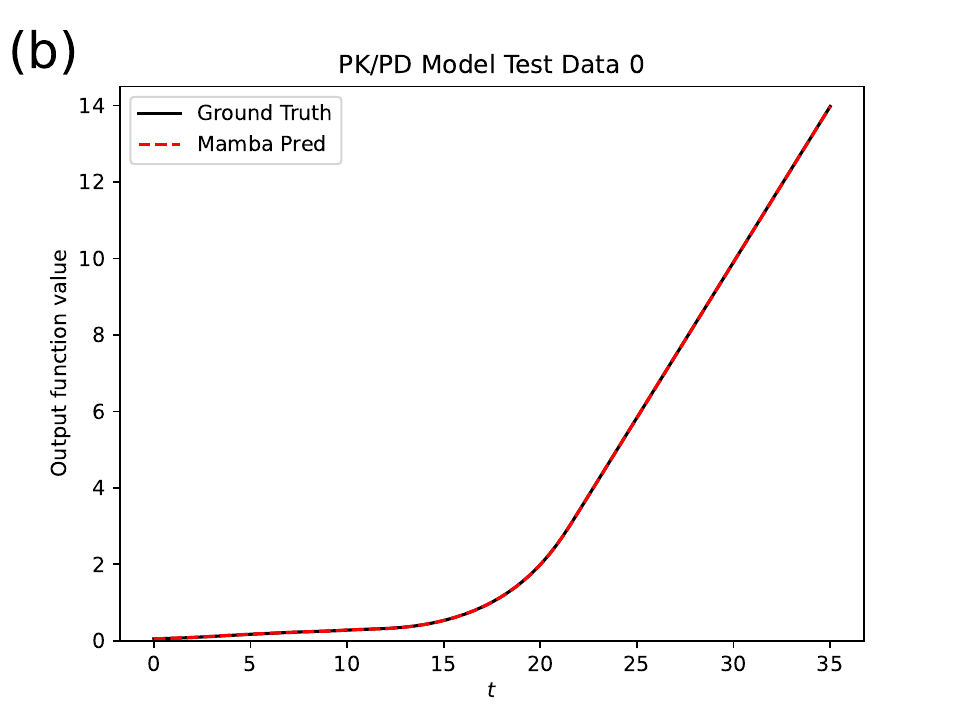}
\includegraphics[width=0.24\linewidth]{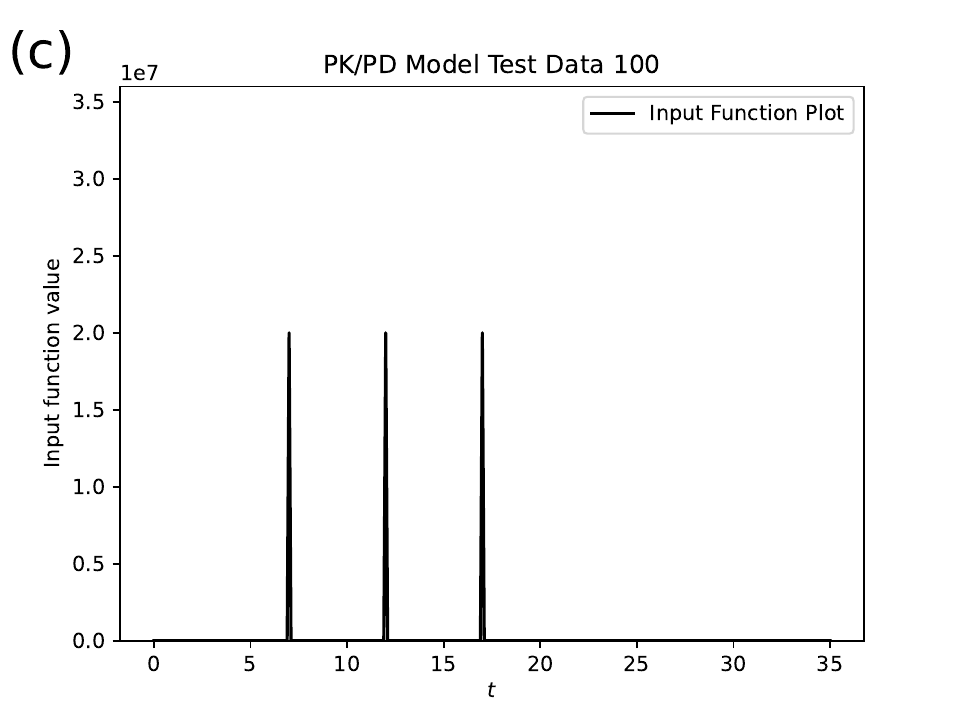}
\includegraphics[width=0.24\linewidth]{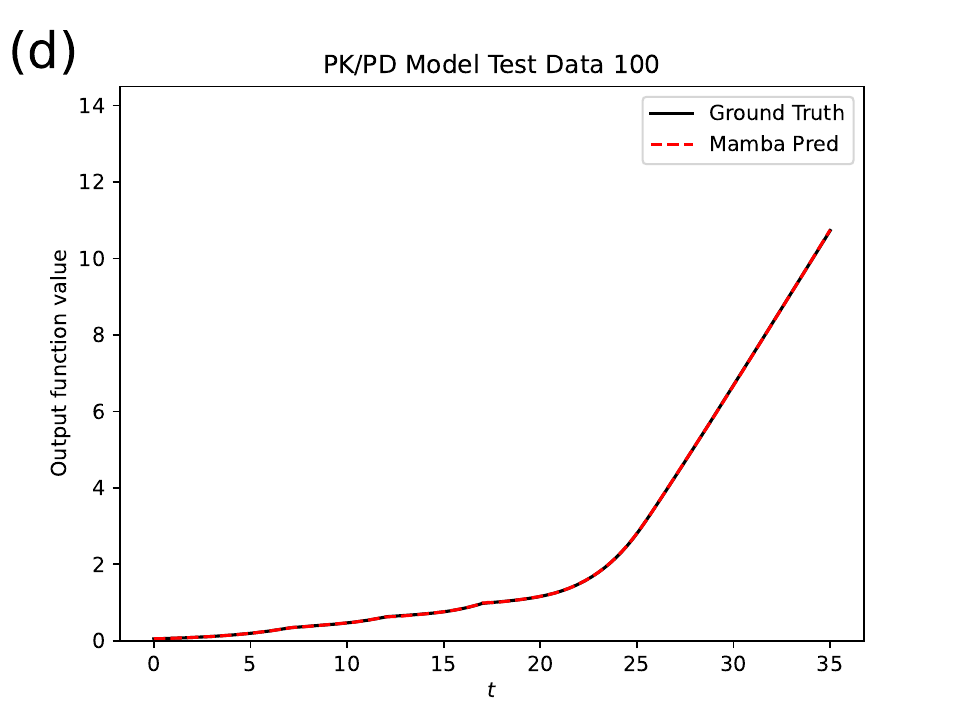}
\includegraphics[width=0.24\linewidth]{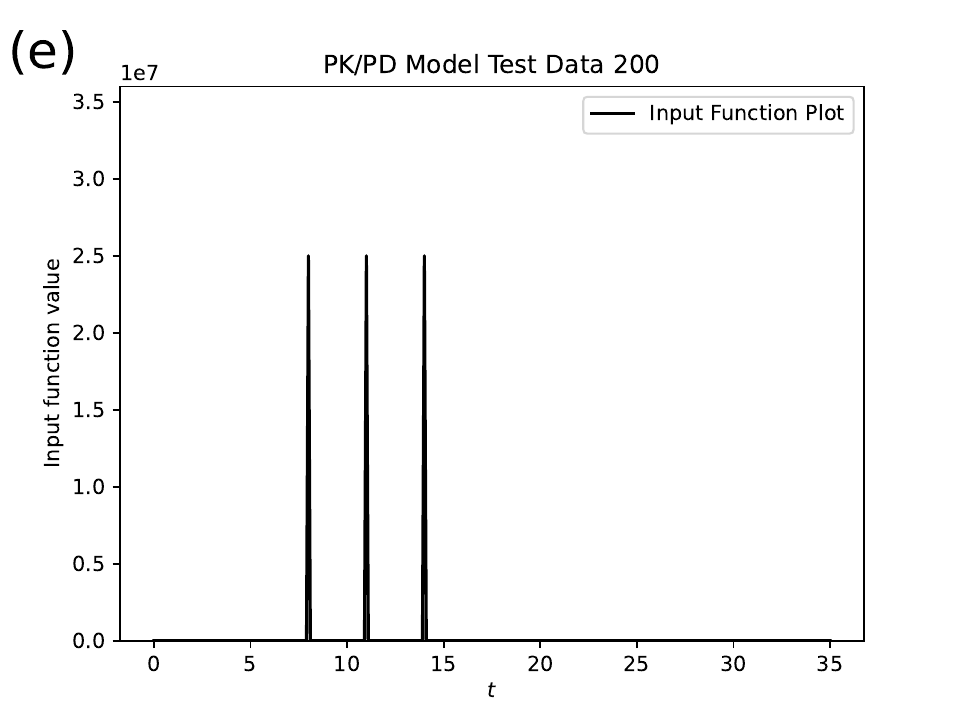}
\includegraphics[width=0.24\linewidth]{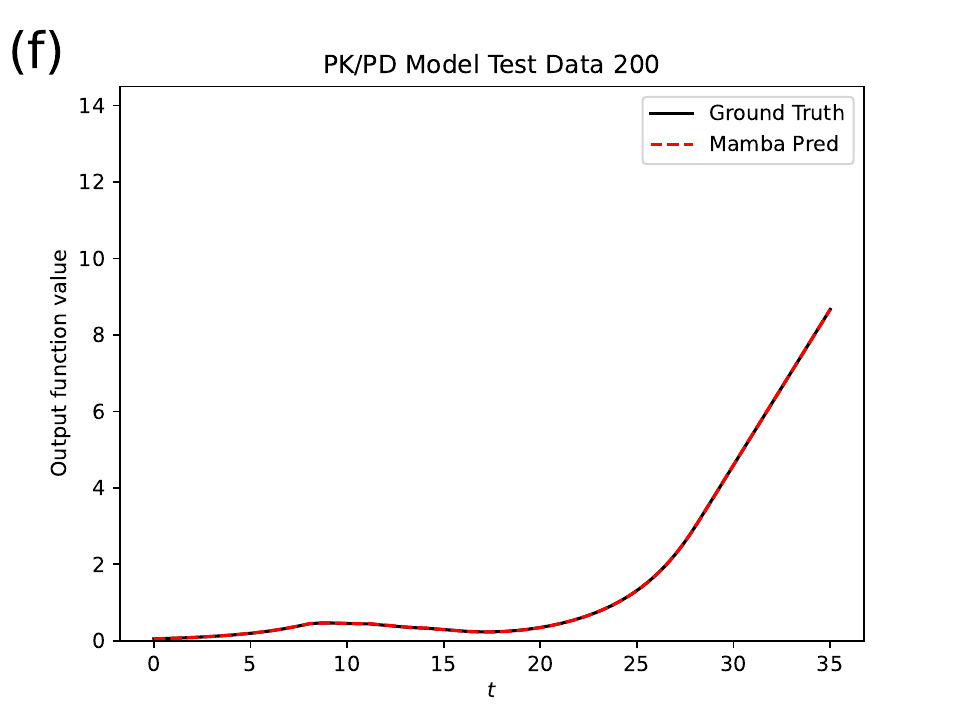}
\includegraphics[width=0.24\linewidth]{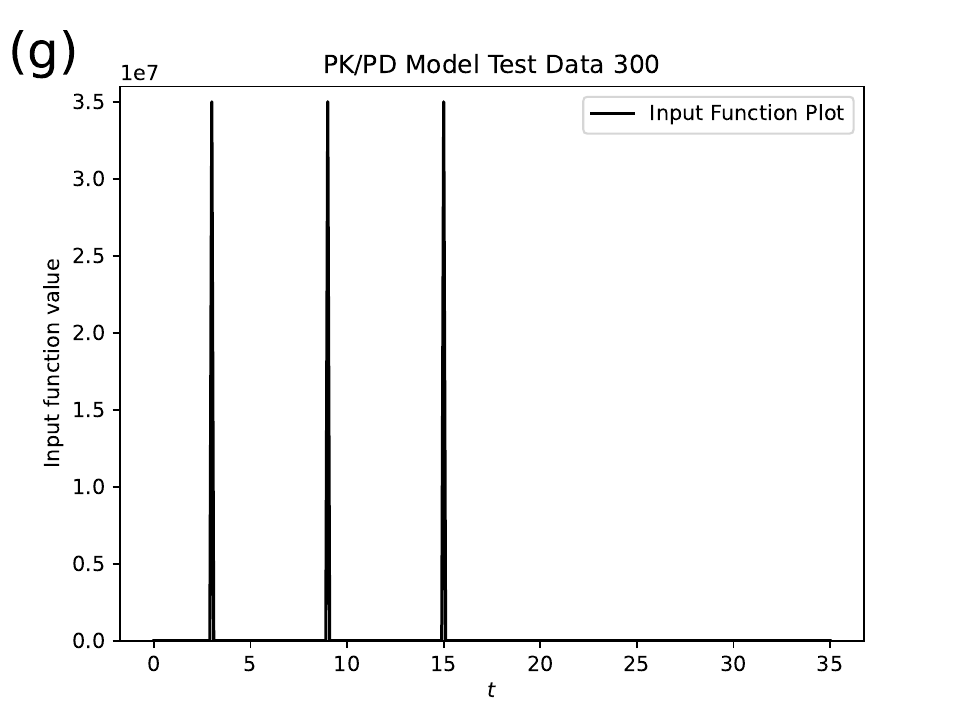}
\includegraphics[width=0.24\linewidth]{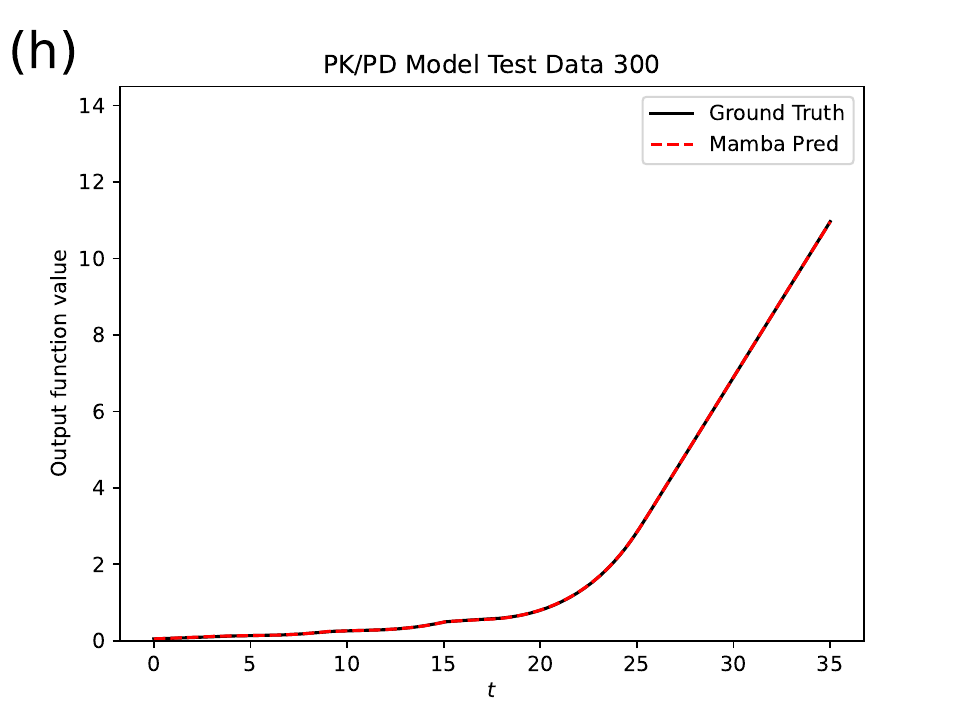}
\caption{Visualization of Mamba's prediction in PK-PD model corresponding to Section \ref{sec:pk_pd_model_1}. We plot the input values and the outputs for test data. 
(a) and (b): input/output (io) for test data No. 0; (c) and (d): io for test data No. 100;
(e) and (f): io for test data No. 200;
(g) and (h): io for test data No. 300. 
Note that the heights and positions of the input spikes vary for different data sets, and the PD parameters differ based on each individual's response to treatment.
Moreover, we visualize loss trajectory in Figure \ref{fig:pk_pd_loss} and Mamba's relative $L_2$ error with respect to time in Figure \ref{fig:PKPD_2}. The full quantitative results are shown in Table \ref{tab:pk_pd}.}
\label{fig:PKPD_1}
\end{figure}

\begin{figure}[htbp]
\centering
\includegraphics[width=0.5\linewidth]{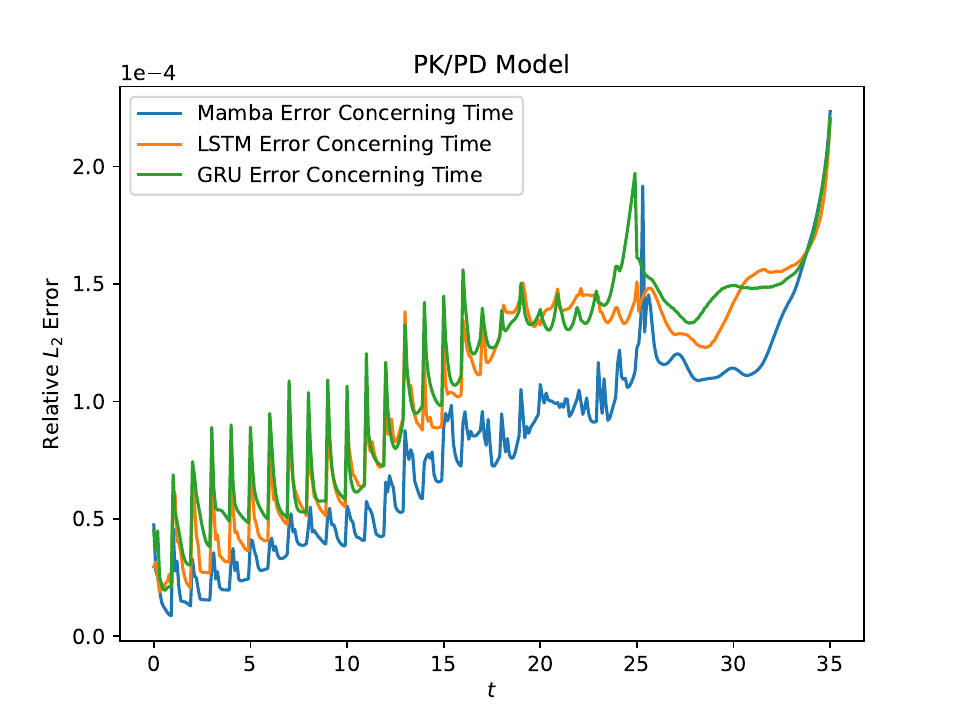}
\caption{Visualization of Mamba’s, GRU's, and LSTM's relative $L_2$ prediction error with respect to time in PK-PD model corresponding to Section \ref{sec:pk_pd_model_1}. The full quantitative results are in Table \ref{tab:pk_pd}. Moreover, we visualize some inputs and Mamba's predictions in Figure \ref{fig:PKPD_1}.}
\label{fig:PKPD_2}
\end{figure}

Results are shown in Table \ref{tab:pk_pd}. Mamba continues to be the best-performing model among multiple strong baselines. We visualize the inputs and outputs of this PK-PD model in Figure \ref{fig:PKPD_1} and plot Mamba's relative $L_2$ error with respect to time in Figure \ref{fig:PKPD_2}. The results show that Mamba can capture the PK-PD model's operator containing discontinuous inputs.
The Mamba model's relative $L_2$ error of 1.685E-3 is equivalent to an error of 0.1685\% of the tumor size. Thus, for a tumor volume of 10 grams, the resulting error is 16.85 milligrams. %\textcolor{blue}{Note that the $L_2$ error should not be interpreted as an error within mg, but rather as a measure of the overall discrepancy between predicted and observed values across the dataset.}
{\color{green!50!black}
Despite a limited 10,000 parameter count budget for Mamba, it can already achieve a satisfactorily low error without the need to further expand the model size for better results.
Lastly, we visualize different models' training loss trajectories in Figure \ref{fig:pk_pd_loss}. Mamba maintains its advantage in faster convergence speed and better optimization results throughout the optimization.}

\subsubsection{Learning the PK-PD Model under Limited Data}\label{sec:pk_pd_model_2}
Due to the constraints of real-world biomedical experiments, the availability of PK-PD model data is often limited. The high cost of conducting experiments, the ethical considerations involved, and the restrictions on data sharing contribute to this scarcity, making it essential to develop models that can efficiently utilize limited data. To this end, we conduct an additional study on Mamba to investigate its generalization with respect to training dataset size under a fixed test set. We generate 5000 data in total as in Section \ref{sec:pk_pd_model_1} and leave 1000 as the fixed test set. We manipulate the train set size within 4000, 2000, 1000, 500, 250, and 125.
Regarding Mamba, we opt for a block with skip connection and layer normalization, with 16 model dimensions and 16 intermediate dimensions, which have 9697 parameters, as in the previous subsection.
We keep all the training settings the same as in the previous subsection except for the number of training epochs. We maintain the product between train set size and the number of training epochs to be 4000 $\times$ 10,000 to maintain the total iterations similar across various settings given less data. For instance, the training epochs number for 2000 data is 20,000.
The results shown in Table \ref{tab:pk_pd_limited_data} demonstrate that Mamba's test error and generalization gap become larger with decreasing training data while the training loss is decreasing. 
As we have less training data, fitting and overfitting the training set becomes easier for the Mamba model, leading to decreasing training loss and increasing the generalization gap. 
However, it can still achieve some good results under limited data settings. 

\begin{table}[htbp]
\centering
\begin{tabular}{|c|c|c|c|}
\hline
Train Data Num & 4000 & 2000 & 1000 \\ \hline
Mamba Params Num & 9697 & 9697 & 9697 \\ \hline
Train Rel. $L_2$ Error & 1.605E-3$\pm$2.514E-4 & 1.512E-3$\pm$1.651E-4 & 1.289E-3$\pm$2.417E-4 \\ \hline
Test Rel. $L_2$ Error & 1.685E-3$\pm$2.929E-4 & 1.940E-3$\pm$2.243E-4 & 2.873E-3$\pm$1.575E-4 \\ \hline\hline
Train Data Num & 500 & 250 & 125 \\ \hline
Mamba Params Num & 9697 & 9697 & 9697 \\ \hline
Train Rel. $L_2$ Error & 1.129E-3$\pm$2.668E-4 & 1.027E-3$\pm$1.479E-4 & 7.137E-4$\pm$1.038E-4 \\ \hline
Test Rel. $L_2$ Error & 6.633E-3$\pm$5.850E-4 & 1.444E-2$\pm$3.052E-3 & 2.009E-2$\pm$1.945E-3 \\ \hline
\end{tabular}
\caption{Results for the PK-PD model under limited data setting corresponding to Section \ref{sec:pk_pd_model_2}.}
\label{tab:pk_pd_limited_data}
\end{table}

\subsubsection{Extrapolation to treatment schedule parameters}\label{sec:pk_pd_model_extrapolation}
In this study, we performed an extrapolation of various treatment schedule parameters to evaluate the efficacy of different dosing regimens. The analysis included three distinct datasets, each containing 200 out-of-distribution test data, and was designed to test the model's robustness and predictive capability under various conditions. The administration dose was maintained within the therapeutic range to avoid the toxic zone, so we did not change this range. Data was generated to ensure that all dose amounts were uniformly distributed across the first and second datasets, differing from the training dataset, where doses were selected from a random uniform function. The input values for the treatment plan were modified to study the efficacy of different schedules in specific populations. Since the population remains the same, the PD parameters were kept within the training range using random uniform values. The first dataset includes treatment schedules with intervals that are out-of-training-range (e.g., every 7 and 8 days) and first administration days that are out-of-training-range (e.g., day 14, 15, or 16). Within the training time range of [0, 35], all patients receive only two doses in a multi-dose administration. In the second dataset, the input values for the first day of administration were set out of range, while the schedule for the second dataset remained the same (e.g., day 2 to day 6). This approach was used to test the impact of starting cancer treatment later than in the training dataset. In the third dataset, the dose is chosen from a random uniform function similar to the training dataset. The same schedule is applied but with a different first day of administration to ensure all patients receive three treatments within the [0, 35] day range.
In addition, the training set is the 5,000 data generated in Section \ref{sec:pk_pd_model_1}, which is a different distribution from all three test sets. We choose the top three models in the IID test in Section \ref{sec:pk_pd_model_1}: GRU, LSTM, and Mamba. We also adopt the same model architecture and optimization procedure as in Section \ref{sec:pk_pd_model_1}.

\begin{table}[htbp]
\centering
\begin{tabular}{|c|c|ccc|}
\hline
 & Train & Test 1 & Test 2 & Test 3 \\ \hline
GRU & 1.818E-3$\pm$2.578E-4 & 7.957E-2$\pm$1.506E-2 & 5.458E-2$\pm$6.455E-3 & 5.790E-2$\pm$1.014E-2 \\ \hline
LSTM & 1.759E-3$\pm$3.761E-4 & 9.427E-2$\pm$4.032E-2 & 7.204E-2$\pm$4.191E-2 & 7.118E-2$\pm$3.782E-2 \\ \hline
Mamba & \textbf{1.321E-3$\pm$2.016E-4} & \textbf{5.554E-2$\pm$2.927E-3} & \textbf{5.414E-2$\pm$2.032E-3} & \textbf{5.382E-2$\pm$2.118E-3} \\ \hline
\end{tabular}
\caption{Performance comparison of different models and different testing datasets for extrapolation corresponding to Section \ref{sec:pk_pd_model_extrapolation}. The best model, Mamba, is bold.}
\label{tab:pkpd_ext}
\end{table}

\textbf{Results}. Results in Table \ref{tab:pkpd_ext}
indicate that the third dataset, which is more akin to the training dataset, has the lowest relative $L_2$ error, while the first dataset has the highest relative $L_2$ error value. In general, we observe the train-test generalization gap due to the out-of-distribution datasets, but we still achieve a relative error of about 5\%, so Mamba performs well. 

\subsubsection{Physics-Informed Mamba Operator Learning}
Herein, we investigate how the incorporation of physics information can improve Mamba's generalization under limited training data.

\textbf{Motivation and Experimental Setting}. As we mentioned, the PK-PD model data require real-world laboratory experiments; we need to learn the PK model under limited data. 
To investigate the efficacy of training the Mamba model with very limited data, specifically with only five labeled training samples and 50 test samples, we explored how incorporating physical information from the ODE might enhance model performance. In addition to the labeled data, we utilized 45 unlabeled data points, which provide input information without corresponding labels. 
All data are generated according to Section \ref{sec:PKPD_data_generation_method}.
Specifically, the loss function we employed consists of a data loss term for the labeled data and a physics loss term for the unlabeled data. 
The time-derivative in the physics-informed loss is computed via the central finite difference scheme.
The data and physics-informed losses are all given unity weights.
Our objective is to examine whether integrating physical information during model training can improve Mamba's generalization capability and assess its practical applicability in real-world scenarios. The Mamba structure and optimization procedure are the same as Section \ref{sec:pk_pd_model_1} except the training epochs number is prolonged to 100,000 due to the smaller dataset.

\textbf{Results}. We obtain a relative $L_2$ error of 8.718E-2$\pm$2.591E-2 with data loss only. If the physics-informed loss on the unlabeled data is employed in addition to the data loss, i.e., the hybrid method with both data loss and physics loss, we improve the error to 1.872E-2$\pm$1.023E-2. However, if we discard the labeled data loss and only train on the physics-informed loss on the unlabeled data, then we obtain an error of 8.791E-1$\pm$1.576E-1. In the physics-informed loss function, the ODE on $A_1(t)$ is stiff and thus rigid to learn due to the discontinuous $\mu$, which may not be an excellent supervision to guide Mamba to the ODE solution. In contrast, the data loss in the hybrid approach pins the model to the correct solution, making it the strongest training methodology. In conclusion, Mamba can be flexibly trained in data-driven, physics-informed, or hybrid scenarios. The hybrid approach is the best, thanks to blending data with physics. This demonstrates Mamba's potential in handling real-world PK-PD models when experimental data is scarce by using physics to improve its data efficiency and generalization.

\section{Summary and Discussion}

In this paper, we addressed the challenges associated with solving dynamical systems using traditional numerical methods and explored the potential of physics-informed machine learning (PIML) as a powerful alternative. While existing PIML approaches, such as recurrent neural networks (RNNs), Transformers, and neural operators, have made significant progress, they still face limitations in handling long-term dependencies, chaotic dynamics, long-time integrations, extrapolation to unseen test functions and longer times, etc.

To overcome these limitations, we introduced the Mamba and state-space models to efficiently capture the complex behavior inherent in dynamical systems. Mamba's architecture dynamically adjusts to capture long-range dependencies and utilizes re-parameterization techniques for enhanced parallelization and computational efficiency.

Through extensive experimentation across diverse dynamical systems, including rough and discontinuous solutions, extrapolation to unseen test functions or long times, long-time integration, chaotic systems, and real-world applications, we demonstrated that Mamba consistently outperforms or matches the state-of-the-art models in interpolation and extrapolation tasks. Mamba also excels in handling long-time integration efficiently.
Specifically, 
Section \ref{sec:1D_DS_DeepONet} investigates three basic systems taken from the first neural operator paper, DeepONet \cite{lu2019deeponet}.
Section \ref{sec:finite_regularity} considers finite regularity and discontinuous solutions following Shih et al. \cite{shih2024transformers}.
Section \ref{sec:lnoode} follows Cao et al. \cite{cao2023lno} and considers out-of-distribution operator learning tasks where the train and test functions follow different distributions. We also investigate the effect of ODE parameters and consider 
chaotic Lorenz systems.
Section \ref{sec:long_time_integration_lno_pend} also follows Cao et al. \cite{cao2023lno} but goes to longer-time integration with a maximal sequence length of 32768.
Section \ref{sec:extrapolation_long_time_1D_DS_deeponet} follows DeepONet's \cite{lu2019deeponet} test cases and considers extrapolation to longer-time problems where the model is trained on a shorter time but is asked to predict on a longer time during test time.
Section \ref{sec:Lorenz_init} considers the chaotic Lorenz system and tests models to directly map from the initial condition to the entire solution trajectory.
Section \ref{sec:quantify_extrapolation_error_zhu2023reliable} quantifies the extrapolation error of various models via gradually moving the test distribution away from the train one.
Section \ref{sec:pk_pd_model} considers a real-world PK-PD model and tests out-of-distribution generalization with test distribution shift, learning under limited data, and physics-informed operator learning. 
 
Our findings underscore the potential of Mamba as a versatile and efficient tool for dynamical system modeling. By bridging the gap between traditional numerical solvers and modern machine learning techniques, Mamba opens new avenues for advancing scientific research and practical applications across multiple domains. Future work will explore further enhancements to the Mamba architecture, expanding its applicability and refining its performance on even more challenging dynamical systems and going beyond ODEs to PDEs. This can be readily accomplished by combining DeepOnet with Mamba or a Unet architecture with Mamba, e.g., following the work in \cite{ovadia2023ditto}, where the spatial representation is based on a Unet while conditioning on time using an extra MLP.

%~\\
%\noindent\textbf{Discussion on the PK-PD model}: 
%
Beyond the superior performance of Mamba on this plurality of benchmarks, we also demonstrated its effectiveness in an important real-world application.
One of the most significant challenges in the field of Quantitative Systems Pharmacology is the prediction of parameters for known compartmental and non-compartmental models. This is particularly the case for parameters of PK and PD of patient-specific models for individuals within a population. The objective of our future work is to integrate the Mamba model as a predictive tool for parameter inference, whether constant or time-varying \cite{cminns}, with the aim of providing tailored results for each individual. Such integration would facilitate the assessment of disparate treatment regimen outcomes for each patient based on a limited set of patient-specific features.
It is important to note that in the present study, a single PK model was used for all patients under the assumption that a specific drug was used. Traditionally, a single PK model is developed for each drug within a given population. However, in reality, the concentration of the same drug in plasma varies among individuals due to differences in ADME (absorption, distribution, metabolism, and excretion) properties of the same drug for each patient. With sufficient data, a more accurate and individualized model could be developed that better reflects real-world applications.

In multi-dose regimens, resistance to chemotherapeutic agents such as paclitaxel is a common challenge. This phenomenon, known as chemotherapy resistance, occurs when tumors that initially respond to treatment begin to grow again. Resistance can result from pharmacodynamic resistance or pharmacokinetic tolerance, particularly in the context of multi-dose regimens. The complexity of modeling these adaptive mechanisms in PK and PD makes it difficult to accurately measure their impact on treatment efficacy. Because they either reduce the effective concentration of the drug or diminish the response of cancer cells to therapy, incorporating additional variables—such as the specific cancer type or other relevant clinical data—can improve the accuracy of patient-specific predictive models.
Therefore, the Mamba models with linear cost, superior accuracy, and robustness could be utilized effectively in such cases.

\newpage

\appendix

\section{More Details on the Baseline Models}\label{appendix:baseline_models}
\subsection{Neural Operators}\label{appendix:baseline_models_NO}
\subsubsection{DeepONet}
A DeepONet \cite{lu2019deeponet} model $\mathcal{O}_\theta$ receives the input function $x$ and a query point $t$, and approximates the output function $y(t) = \left(\mathcal{O}x\right)(t)$ via
\begin{align}
\left(\mathcal{O}_\theta x\right)(t) = B_\theta\left\{\left[x(t_1), x(t_2), \cdots, x\left(t_{N_{\text{grid}}}\right)\right]\right\}^\mathrm{T} T_\theta(t).
\end{align}
The branch net $B_\theta$ encodes the input function's discrete values on the grid $\left[x(t_1), x(t_2), \cdots, x\left(t_{N_{\text{grid}}}\right)\right]$ to a latent vector. Then, the trunk net $T_\theta$ encodes the query point $t$ where the output function will be evaluated to another latent vector. The inner product between these two vectors is the DeepONet output.

\subsubsection{Fourier Neural Operator (FNO)}
FNO \cite{li2020fourier} adopts the trainable kernel integral operator for function transformation on a uniform grid, which can be efficiently computed in the Fourier space. Specifically, an FNO layer maps the input $u(t)$ to the output $v(t)$ as follows:
\begin{align}
v(t) = \left(\kappa_\theta * u\right)(t) = \int \kappa_\theta(t - \tau) u(\tau) d\tau = \mathcal{F}^{-1}\left(\mathcal{F}\left(\kappa_\theta\right)\cdot \mathcal{F}(u)\right)(t),
\end{align}
where $\mathcal{F}$ is the Fourier transform (FT) and $\mathcal{F}^{-1}$ is the inverse FT. Further, $\mathcal{F}(u)$ can be computed via a fast FT algorithm, and FNO parameterized the trainable kernel in the Fourier space:
$
\mathcal{F}(\kappa_\theta) = \sum_{\ell = -\infty}^\infty \alpha_\ell \exp(i\omega_\ell t),
$
where $\omega_\ell = \ell\omega_1$ and $\omega_1$ is the fundamental frequency in rad/s, and $\alpha_\ell$ is the trainable complex Fourier coefficient in FNO.
FNO further stacks multiple of the aforementioned trainable kernel integral operator layers with nonlinear activations to form the final FNO model.
Since FNO works on a uniform grid, the dynamical system is also discretized uniformly in time in our test cases. This enables the direct application of one-dimensional FNO to our computational experiments.

\subsubsection{Laplace Neural Operator (LNO)}
LNO is also built upon the trainable kernel integral operator, but in the Laplace space, mapping the input $u(t)$ to the output $v(t)$:
\begin{align}
v(t) = \left(\kappa_\theta * u\right)(t) = \int \kappa_\theta(t - \tau) u(\tau) d\tau.
\end{align}
Taking the Laplace transform $\mathcal{L}$ yields:
\begin{align}
V(s) = \mathcal{L}\left\{\left(\kappa_\theta * u\right)(t)\right\} = K_\theta(s) U(s),
\end{align}
where $K_\theta(s) = \mathcal{L}\left\{\kappa_\theta(t)\right\}$ and $U(s) = \mathcal{L}\left\{u(t)\right\}$.
Furthermore, $K_\theta(s)$ can be represented in the pole-residue form with poles $\mu_n$ and residues $\beta_n$:
\begin{align}
K_\theta(s) = \sum_{n=1}^N \frac{\beta_n}{s - \mu_n},
\end{align}
where LNO makes the kernel trainable via appointing $\theta = \{\beta_n, \mu_n\}_{n=1}^N$ as model parameters.
To compute the Laplace transform of the input function $u(t)$, we decompose it via Fourier series:
\begin{align}
u(t) = \sum_{\ell = -\infty}^\infty \alpha_\ell \exp(i\omega_\ell t),
\end{align}
where $\omega_\ell = \ell\omega_1$ and $\omega_1$ is the fundamental frequency in rad/s, and $\alpha_\ell$ is the complex Fourier coefficient. Then,
\begin{align}
U(s) = \sum_{\ell = -\infty}^\infty\frac{\alpha_\ell}{s - i\omega_\ell}.
\end{align}
Hence, $V(s) = K_\theta(s) U(s)$ can be derived and further simplified into the pole-residue form:
\begin{align}
V(s) = \left(\sum_{n=1}^N \frac{\beta_n}{s - \mu_n}\right)\left(\sum_{\ell = -\infty}^\infty\frac{\alpha_\ell}{s - i\omega_\ell}\right) = \sum_{n=1}^N \frac{\gamma_n}{s - \mu_n} + \sum_{\ell = -\infty}^\infty\frac{\lambda_\ell}{s - i\omega_\ell},
\end{align}
where 
$
\gamma_n = \beta_n \sum_{\ell = -\infty}^\infty \frac{\alpha_\ell}{\mu_n - i \omega_\ell}
$
and
$
\lambda_\ell = \alpha_\ell \sum_{n=1}^N \frac{\beta_n}{i\omega_\ell - \mu_n}.
$
Consequently, the output function in the original space can be computed via an inverse Laplace transform and be represented by
\begin{align}
v(t) = \sum_{n=1}^N \gamma_n \exp(\mu_n t) + \sum_{\ell = -\infty}^\infty\lambda_\ell \exp(\omega_\ell t).
\end{align}
LNO further stacks multiple aforementioned trainable kernel integral operator layers with nonlinear activations to form the final LNO model.
Since LNO works on a uniform grid, the dynamical system is also discretized uniformly in time in our test cases. This enables the direct application of one-dimensional LNO to our computational experiments.
Overall, LNO parameterized the trainable kernel integral operator in the Laplace space via trainable poles and residue, and it excels at capturing transient responses via the Laplace transform.

\section{Detailed Quantitative Results for Section \ref{sec:quantify_extrapolation_error_zhu2023reliable}}\label{appendix:quantify_extrapolation_error_zhu2023reliable}
\begin{table}[htbp]
\centering\scriptsize
\begin{tabular}{|cc|c|ccccccccc|}
\hline
\multicolumn{2}{|c|}{Pendulum} & IID & \multicolumn{9}{c|}{OOD Extrapolation: Ex-} \\ \hline
\multicolumn{1}{|c|}{Model} & Metric & 0.1 & \multicolumn{1}{c|}{0.2} & \multicolumn{1}{c|}{0.3} & \multicolumn{1}{c|}{0.4} & \multicolumn{1}{c|}{0.5} & \multicolumn{1}{c|}{0.6} & \multicolumn{1}{c|}{0.7} & \multicolumn{1}{c|}{0.8} & \multicolumn{1}{c|}{0.9} & 1 \\ \hline
\multicolumn{1}{|c|}{\multirow{2}{*}{GRU}} & Mean & 5.992E-04 & \multicolumn{1}{c|}{2.494E-03} & \multicolumn{1}{c|}{1.112E-02} & \multicolumn{1}{c|}{1.442E-02} & \multicolumn{1}{c|}{1.628E-02} & \multicolumn{1}{c|}{2.401E-02} & \multicolumn{1}{c|}{2.107E-02} & \multicolumn{1}{c|}{3.049E-02} & \multicolumn{1}{c|}{2.010E-02} & 2.581E-02 \\ \cline{2-12} 
\multicolumn{1}{|c|}{} & Std & 2.301E-05 & \multicolumn{1}{c|}{3.938E-04} & \multicolumn{1}{c|}{1.322E-03} & \multicolumn{1}{c|}{2.017E-03} & \multicolumn{1}{c|}{1.990E-03} & \multicolumn{1}{c|}{2.968E-03} & \multicolumn{1}{c|}{1.595E-03} & \multicolumn{1}{c|}{1.343E-03} & \multicolumn{1}{c|}{2.435E-03} & 2.241E-03 \\ \hline
\multicolumn{1}{|c|}{\multirow{2}{*}{LSTM}} & Mean & 7.893E-04 & \multicolumn{1}{c|}{4.149E-03} & \multicolumn{1}{c|}{1.225E-02} & \multicolumn{1}{c|}{1.669E-02} & \multicolumn{1}{c|}{1.930E-02} & \multicolumn{1}{c|}{2.835E-02} & \multicolumn{1}{c|}{2.432E-02} & \multicolumn{1}{c|}{3.429E-02} & \multicolumn{1}{c|}{2.393E-02} & 2.896E-02 \\ \cline{2-12} 
\multicolumn{1}{|c|}{} & Std & 7.518E-05 & \multicolumn{1}{c|}{3.347E-04} & \multicolumn{1}{c|}{2.760E-04} & \multicolumn{1}{c|}{4.666E-04} & \multicolumn{1}{c|}{3.515E-04} & \multicolumn{1}{c|}{5.668E-04} & \multicolumn{1}{c|}{2.441E-04} & \multicolumn{1}{c|}{2.258E-04} & \multicolumn{1}{c|}{7.072E-04} & 4.910E-04 \\ \hline
\multicolumn{1}{|c|}{\multirow{2}{*}{DeepONet}} & Mean & 1.949E-03 & \multicolumn{1}{c|}{6.438E-03} & \multicolumn{1}{c|}{1.325E-02} & \multicolumn{1}{c|}{1.906E-02} & \multicolumn{1}{c|}{2.253E-02} & \multicolumn{1}{c|}{3.114E-02} & \multicolumn{1}{c|}{2.541E-02} & \multicolumn{1}{c|}{3.658E-02} & \multicolumn{1}{c|}{2.361E-02} & 2.833E-02 \\ \cline{2-12} 
\multicolumn{1}{|c|}{} & Std & 6.547E-04 & \multicolumn{1}{c|}{2.976E-03} & \multicolumn{1}{c|}{5.913E-03} & \multicolumn{1}{c|}{8.861E-03} & \multicolumn{1}{c|}{1.008E-02} & \multicolumn{1}{c|}{1.392E-02} & \multicolumn{1}{c|}{1.148E-02} & \multicolumn{1}{c|}{1.515E-02} & \multicolumn{1}{c|}{1.105E-02} & 1.234E-02 \\ \hline
\multicolumn{1}{|c|}{\multirow{2}{*}{FNO}} & Mean & 6.335E-02 & \multicolumn{1}{c|}{6.755E-02} & \multicolumn{1}{c|}{8.991E-02} & \multicolumn{1}{c|}{1.127E-01} & \multicolumn{1}{c|}{1.160E-01} & \multicolumn{1}{c|}{1.322E-01} & \multicolumn{1}{c|}{1.536E-01} & \multicolumn{1}{c|}{1.941E-01} & \multicolumn{1}{c|}{1.980E-01} & 2.059E-01 \\ \cline{2-12} 
\multicolumn{1}{|c|}{} & Std & 1.015E-02 & \multicolumn{1}{c|}{1.470E-02} & \multicolumn{1}{c|}{2.380E-02} & \multicolumn{1}{c|}{3.336E-02} & \multicolumn{1}{c|}{3.790E-02} & \multicolumn{1}{c|}{3.641E-02} & \multicolumn{1}{c|}{4.457E-02} & \multicolumn{1}{c|}{5.259E-02} & \multicolumn{1}{c|}{5.545E-02} & 5.798E-02 \\ \hline
\multicolumn{1}{|c|}{\multirow{2}{*}{LNO}} & Mean & 1.942E-03 & \multicolumn{1}{c|}{1.062E-02} & \multicolumn{1}{c|}{2.728E-02} & \multicolumn{1}{c|}{4.087E-02} & \multicolumn{1}{c|}{4.838E-02} & \multicolumn{1}{c|}{7.193E-02} & \multicolumn{1}{c|}{5.745E-02} & \multicolumn{1}{c|}{8.510E-02} & \multicolumn{1}{c|}{5.308E-02} & 6.668E-02 \\ \cline{2-12} 
\multicolumn{1}{|c|}{} & Std & 3.373E-04 & \multicolumn{1}{c|}{2.681E-03} & \multicolumn{1}{c|}{7.010E-03} & \multicolumn{1}{c|}{1.093E-02} & \multicolumn{1}{c|}{1.107E-02} & \multicolumn{1}{c|}{1.617E-02} & \multicolumn{1}{c|}{1.313E-02} & \multicolumn{1}{c|}{1.582E-02} & \multicolumn{1}{c|}{1.409E-02} & 1.459E-02 \\ \hline
\multicolumn{1}{|c|}{\multirow{2}{*}{Transformer}} & Mean & 3.043E-02 & \multicolumn{1}{c|}{2.553E-02} & \multicolumn{1}{c|}{2.609E-02} & \multicolumn{1}{c|}{3.160E-02} & \multicolumn{1}{c|}{3.452E-02} & \multicolumn{1}{c|}{4.393E-02} & \multicolumn{1}{c|}{4.235E-02} & \multicolumn{1}{c|}{5.624E-02} & \multicolumn{1}{c|}{4.157E-02} & 4.598E-02 \\ \cline{2-12} 
\multicolumn{1}{|c|}{} & Std & 9.951E-04 & \multicolumn{1}{c|}{3.193E-03} & \multicolumn{1}{c|}{3.195E-03} & \multicolumn{1}{c|}{1.139E-02} & \multicolumn{1}{c|}{1.291E-02} & \multicolumn{1}{c|}{2.369E-02} & \multicolumn{1}{c|}{2.077E-02} & \multicolumn{1}{c|}{3.032E-02} & \multicolumn{1}{c|}{2.073E-02} & 2.361E-02 \\ \hline
\multicolumn{1}{|c|}{\multirow{2}{*}{Oformer-V}} & Mean & 6.149E-04 & \multicolumn{1}{c|}{1.002E-03} & \multicolumn{1}{c|}{2.460E-03} & \multicolumn{1}{c|}{3.088E-03} & \multicolumn{1}{c|}{3.985E-03} & \multicolumn{1}{c|}{6.035E-03} & \multicolumn{1}{c|}{5.370E-03} & \multicolumn{1}{c|}{8.776E-03} & \multicolumn{1}{c|}{4.563E-03} & 6.700E-03 \\ \cline{2-12} 
\multicolumn{1}{|c|}{} & Std & 7.592E-05 & \multicolumn{1}{c|}{3.519E-04} & \multicolumn{1}{c|}{1.225E-03} & \multicolumn{1}{c|}{1.687E-03} & \multicolumn{1}{c|}{2.287E-03} & \multicolumn{1}{c|}{3.649E-03} & \multicolumn{1}{c|}{3.169E-03} & \multicolumn{1}{c|}{5.271E-03} & \multicolumn{1}{c|}{2.822E-03} & 4.020E-03 \\ \hline
\multicolumn{1}{|c|}{\multirow{2}{*}{Oformer-G}} & Mean & 6.577E-04 & \multicolumn{1}{c|}{3.970E-04} & \multicolumn{1}{c|}{5.075E-04} & \multicolumn{1}{c|}{5.177E-04} & \multicolumn{1}{c|}{5.665E-04} & \multicolumn{1}{c|}{7.167E-04} & \multicolumn{1}{c|}{6.968E-04} & \multicolumn{1}{c|}{2.258E-03} & \multicolumn{1}{c|}{6.199E-04} & 8.159E-04 \\ \cline{2-12} 
\multicolumn{1}{|c|}{} & Std & 1.853E-04 & \multicolumn{1}{c|}{1.742E-04} & \multicolumn{1}{c|}{1.748E-04} & \multicolumn{1}{c|}{1.879E-04} & \multicolumn{1}{c|}{1.409E-04} & \multicolumn{1}{c|}{1.995E-04} & \multicolumn{1}{c|}{2.986E-04} & \multicolumn{1}{c|}{1.066E-03} & \multicolumn{1}{c|}{2.723E-04} & 2.525E-04 \\ \hline
\multicolumn{1}{|c|}{\multirow{2}{*}{Oformer-F}} & Mean & 6.134E-04 & \multicolumn{1}{c|}{3.319E-04} & \multicolumn{1}{c|}{3.853E-04} & \multicolumn{1}{c|}{4.168E-04} & \multicolumn{1}{c|}{4.133E-04} & \multicolumn{1}{c|}{4.555E-04} & \multicolumn{1}{c|}{4.138E-04} & \multicolumn{1}{c|}{1.436E-03} & \multicolumn{1}{c|}{\textbf{3.660E-04}} & 5.189E-04 \\ \cline{2-12} 
\multicolumn{1}{|c|}{} & Std & 2.263E-04 & \multicolumn{1}{c|}{1.528E-04} & \multicolumn{1}{c|}{1.996E-04} & \multicolumn{1}{c|}{2.517E-04} & \multicolumn{1}{c|}{2.155E-04} & \multicolumn{1}{c|}{2.440E-04} & \multicolumn{1}{c|}{2.138E-04} & \multicolumn{1}{c|}{1.067E-03} & \multicolumn{1}{c|}{1.778E-04} & 3.079E-04 \\ \hline
\multicolumn{1}{|c|}{\multirow{2}{*}{GNOT}} & Mean & 6.422E-04 & \multicolumn{1}{c|}{5.473E-04} & \multicolumn{1}{c|}{1.144E-03} & \multicolumn{1}{c|}{1.345E-03} & \multicolumn{1}{c|}{1.547E-03} & \multicolumn{1}{c|}{2.516E-03} & \multicolumn{1}{c|}{2.490E-03} & \multicolumn{1}{c|}{4.184E-03} & \multicolumn{1}{c|}{2.182E-03} & 2.978E-03 \\ \cline{2-12} 
\multicolumn{1}{|c|}{} & Std & 1.465E-05 & \multicolumn{1}{c|}{1.373E-04} & \multicolumn{1}{c|}{4.592E-04} & \multicolumn{1}{c|}{6.569E-04} & \multicolumn{1}{c|}{7.168E-04} & \multicolumn{1}{c|}{1.159E-03} & \multicolumn{1}{c|}{1.105E-03} & \multicolumn{1}{c|}{2.033E-03} & \multicolumn{1}{c|}{9.995E-04} & 1.295E-03 \\ \hline
\multicolumn{1}{|c|}{\multirow{2}{*}{Mamba}} & Mean & \textbf{2.776E-04} & \multicolumn{1}{c|}{\textbf{2.948E-04}} & \multicolumn{1}{c|}{\textbf{3.412E-04}} & \multicolumn{1}{c|}{\textbf{3.765E-04}} & \multicolumn{1}{c|}{\textbf{3.921E-04}} & \multicolumn{1}{c|}{\textbf{4.134E-04}} & \multicolumn{1}{c|}{\textbf{3.907E-04}} & \multicolumn{1}{c|}{\textbf{4.337E-04}} & \multicolumn{1}{c|}{3.709E-04} & \textbf{3.755E-04} \\ \cline{2-12} 
\multicolumn{1}{|c|}{} & Std & 5.312E-06 & \multicolumn{1}{c|}{4.985E-06} & \multicolumn{1}{c|}{1.396E-05} & \multicolumn{1}{c|}{2.851E-05} & \multicolumn{1}{c|}{3.648E-05} & \multicolumn{1}{c|}{5.663E-05} & \multicolumn{1}{c|}{5.880E-05} & \multicolumn{1}{c|}{8.012E-05} & \multicolumn{1}{c|}{6.632E-05} & 7.554E-05 \\ \hline
\end{tabular}
\caption{Gravity pendulum Ex- setting with $l_{train} = 0.1$ and $l_{test} \in \{0.1,0.2,\cdots,1.0\}$ to quantify the extrapolation error following Zhu et al. \cite{zhu2023reliable}, corresponding to Section \ref{sec:quantify_extrapolation_error_zhu2023reliable}. The table's content is also visualized in Figure \ref{fig:Ex-_Pend}. Here, we train on $l_{train} = 0.1$, so testing on $l_{test} = 0.1$ is the independent and identically distributed (IID) test setting while $l_{train} \neq 0.1$ is the extrapolation testing. The best model is bold.}
\label{tab:Ex-_Pend}
\end{table}

\begin{table}[htbp]
\centering\scriptsize
\begin{tabular}{|cc|ccccccccc|c|}
\hline
\multicolumn{2}{|c|}{Pendulum} & \multicolumn{9}{c|}{OOD Extrapolation: Ex+} & IID \\ \hline
\multicolumn{1}{|c|}{Model} & Metric & \multicolumn{1}{c|}{0.1} & \multicolumn{1}{c|}{0.2} & \multicolumn{1}{c|}{0.3} & \multicolumn{1}{c|}{0.4} & \multicolumn{1}{c|}{0.5} & \multicolumn{1}{c|}{0.6} & \multicolumn{1}{c|}{0.7} & \multicolumn{1}{c|}{0.8} & 0.9 & 1 \\ \hline
\multicolumn{1}{|c|}{\multirow{2}{*}{GRU}} & Mean & \multicolumn{1}{c|}{7.693E-02} & \multicolumn{1}{c|}{3.499E-02} & \multicolumn{1}{c|}{1.961E-02} & \multicolumn{1}{c|}{7.521E-03} & \multicolumn{1}{c|}{5.027E-03} & \multicolumn{1}{c|}{2.408E-03} & \multicolumn{1}{c|}{1.188E-03} & \multicolumn{1}{c|}{8.439E-04} & 4.031E-04 & 4.249E-04 \\ \cline{2-12} 
\multicolumn{1}{|c|}{} & Std & \multicolumn{1}{c|}{7.460E-03} & \multicolumn{1}{c|}{3.892E-03} & \multicolumn{1}{c|}{3.346E-03} & \multicolumn{1}{c|}{1.011E-03} & \multicolumn{1}{c|}{9.871E-04} & \multicolumn{1}{c|}{5.223E-04} & \multicolumn{1}{c|}{1.868E-04} & \multicolumn{1}{c|}{1.389E-04} & 2.473E-05 & 1.580E-05 \\ \hline
\multicolumn{1}{|c|}{\multirow{2}{*}{LSTM}} & Mean & \multicolumn{1}{c|}{5.494E-02} & \multicolumn{1}{c|}{2.532E-02} & \multicolumn{1}{c|}{1.421E-02} & \multicolumn{1}{c|}{6.832E-03} & \multicolumn{1}{c|}{4.327E-03} & \multicolumn{1}{c|}{1.973E-03} & \multicolumn{1}{c|}{1.146E-03} & \multicolumn{1}{c|}{1.253E-03} & 4.167E-04 & 4.376E-04 \\ \cline{2-12} 
\multicolumn{1}{|c|}{} & Std & \multicolumn{1}{c|}{3.538E-03} & \multicolumn{1}{c|}{3.071E-03} & \multicolumn{1}{c|}{2.270E-03} & \multicolumn{1}{c|}{9.035E-04} & \multicolumn{1}{c|}{7.105E-04} & \multicolumn{1}{c|}{2.483E-04} & \multicolumn{1}{c|}{8.139E-05} & \multicolumn{1}{c|}{2.041E-04} & 3.837E-06 & 5.258E-05 \\ \hline
\multicolumn{1}{|c|}{\multirow{2}{*}{DeepONet}} & Mean & \multicolumn{1}{c|}{2.333E-01} & \multicolumn{1}{c|}{7.821E-02} & \multicolumn{1}{c|}{2.864E-02} & \multicolumn{1}{c|}{1.054E-02} & \multicolumn{1}{c|}{4.455E-03} & \multicolumn{1}{c|}{1.976E-03} & \multicolumn{1}{c|}{9.950E-04} & \multicolumn{1}{c|}{3.164E-03} & 3.709E-04 & 1.033E-03 \\ \cline{2-12} 
\multicolumn{1}{|c|}{} & Std & \multicolumn{1}{c|}{7.506E-02} & \multicolumn{1}{c|}{2.721E-02} & \multicolumn{1}{c|}{9.835E-03} & \multicolumn{1}{c|}{2.740E-03} & \multicolumn{1}{c|}{7.972E-04} & \multicolumn{1}{c|}{2.257E-04} & \multicolumn{1}{c|}{6.357E-05} & \multicolumn{1}{c|}{1.109E-03} & 3.505E-05 & 2.519E-04 \\ \hline
\multicolumn{1}{|c|}{\multirow{2}{*}{FNO}} & Mean & \multicolumn{1}{c|}{8.276E-01} & \multicolumn{1}{c|}{4.336E-01} & \multicolumn{1}{c|}{2.942E-01} & \multicolumn{1}{c|}{1.229E-01} & \multicolumn{1}{c|}{3.033E-01} & \multicolumn{1}{c|}{1.846E-01} & \multicolumn{1}{c|}{3.387E-02} & \multicolumn{1}{c|}{2.071E-02} & 2.557E-02 & 1.835E-02 \\ \cline{2-12} 
\multicolumn{1}{|c|}{} & Std & \multicolumn{1}{c|}{5.038E-02} & \multicolumn{1}{c|}{2.147E-02} & \multicolumn{1}{c|}{6.943E-02} & \multicolumn{1}{c|}{1.413E-02} & \multicolumn{1}{c|}{2.474E-01} & \multicolumn{1}{c|}{1.509E-01} & \multicolumn{1}{c|}{1.509E-03} & \multicolumn{1}{c|}{2.601E-04} & 1.353E-05 & 1.010E-03 \\ \hline
\multicolumn{1}{|c|}{\multirow{2}{*}{LNO}} & Mean & \multicolumn{1}{c|}{1.298E-01} & \multicolumn{1}{c|}{4.380E-02} & \multicolumn{1}{c|}{1.648E-02} & \multicolumn{1}{c|}{6.948E-03} & \multicolumn{1}{c|}{3.417E-03} & \multicolumn{1}{c|}{1.810E-03} & \multicolumn{1}{c|}{1.048E-03} & \multicolumn{1}{c|}{1.327E-03} & 4.259E-04 & 5.044E-04 \\ \cline{2-12} 
\multicolumn{1}{|c|}{} & Std & \multicolumn{1}{c|}{1.567E-03} & \multicolumn{1}{c|}{6.290E-04} & \multicolumn{1}{c|}{4.948E-04} & \multicolumn{1}{c|}{2.661E-04} & \multicolumn{1}{c|}{1.560E-04} & \multicolumn{1}{c|}{7.509E-05} & \multicolumn{1}{c|}{3.838E-05} & \multicolumn{1}{c|}{1.576E-04} & 1.877E-05 & 3.626E-05 \\ \hline
\multicolumn{1}{|c|}{\multirow{2}{*}{Transformer}} & Mean & \multicolumn{1}{c|}{3.623E-01} & \multicolumn{1}{c|}{2.312E-01} & \multicolumn{1}{c|}{1.684E-01} & \multicolumn{1}{c|}{1.134E-01} & \multicolumn{1}{c|}{7.871E-02} & \multicolumn{1}{c|}{4.260E-02} & \multicolumn{1}{c|}{2.768E-02} & \multicolumn{1}{c|}{1.758E-02} & 1.538E-02 & 1.473E-02 \\ \cline{2-12} 
\multicolumn{1}{|c|}{} & Std & \multicolumn{1}{c|}{2.580E-02} & \multicolumn{1}{c|}{2.063E-02} & \multicolumn{1}{c|}{1.765E-02} & \multicolumn{1}{c|}{1.890E-02} & \multicolumn{1}{c|}{9.946E-03} & \multicolumn{1}{c|}{6.458E-03} & \multicolumn{1}{c|}{3.333E-03} & \multicolumn{1}{c|}{7.841E-04} & 4.882E-04 & 6.280E-04 \\ \hline
\multicolumn{1}{|c|}{\multirow{2}{*}{Oformer-V}} & Mean & \multicolumn{1}{c|}{3.562E-01} & \multicolumn{1}{c|}{1.013E-01} & \multicolumn{1}{c|}{3.438E-02} & \multicolumn{1}{c|}{1.335E-02} & \multicolumn{1}{c|}{6.245E-03} & \multicolumn{1}{c|}{3.040E-03} & \multicolumn{1}{c|}{1.569E-03} & \multicolumn{1}{c|}{1.076E-03} & 5.498E-04 & 4.338E-04 \\ \cline{2-12} 
\multicolumn{1}{|c|}{} & Std & \multicolumn{1}{c|}{8.122E-02} & \multicolumn{1}{c|}{1.176E-02} & \multicolumn{1}{c|}{3.195E-03} & \multicolumn{1}{c|}{1.250E-03} & \multicolumn{1}{c|}{6.783E-04} & \multicolumn{1}{c|}{3.392E-04} & \multicolumn{1}{c|}{2.082E-04} & \multicolumn{1}{c|}{9.859E-05} & 6.452E-05 & 5.031E-05 \\ \hline
\multicolumn{1}{|c|}{\multirow{2}{*}{Oformer-G}} & Mean & \multicolumn{1}{c|}{6.109E-02} & \multicolumn{1}{c|}{2.411E-02} & \multicolumn{1}{c|}{9.964E-03} & \multicolumn{1}{c|}{4.436E-03} & \multicolumn{1}{c|}{2.195E-03} & \multicolumn{1}{c|}{1.102E-03} & \multicolumn{1}{c|}{6.137E-04} & \multicolumn{1}{c|}{4.572E-04} & 2.441E-04 & 2.179E-04 \\ \cline{2-12} 
\multicolumn{1}{|c|}{} & Std & \multicolumn{1}{c|}{6.217E-03} & \multicolumn{1}{c|}{7.248E-04} & \multicolumn{1}{c|}{4.695E-04} & \multicolumn{1}{c|}{3.873E-04} & \multicolumn{1}{c|}{2.692E-04} & \multicolumn{1}{c|}{1.396E-04} & \multicolumn{1}{c|}{9.158E-05} & \multicolumn{1}{c|}{8.254E-05} & 4.565E-05 & 6.217E-05 \\ \hline
\multicolumn{1}{|c|}{\multirow{2}{*}{Oformer-F}} & Mean & \multicolumn{1}{c|}{5.853E-02} & \multicolumn{1}{c|}{2.641E-02} & \multicolumn{1}{c|}{1.183E-02} & \multicolumn{1}{c|}{5.376E-03} & \multicolumn{1}{c|}{2.648E-03} & \multicolumn{1}{c|}{1.287E-03} & \multicolumn{1}{c|}{7.096E-04} & \multicolumn{1}{c|}{5.042E-04} & 2.708E-04 & 2.606E-04 \\ \cline{2-12} 
\multicolumn{1}{|c|}{} & Std & \multicolumn{1}{c|}{1.305E-03} & \multicolumn{1}{c|}{7.313E-04} & \multicolumn{1}{c|}{1.203E-03} & \multicolumn{1}{c|}{1.033E-03} & \multicolumn{1}{c|}{5.552E-04} & \multicolumn{1}{c|}{2.691E-04} & \multicolumn{1}{c|}{1.397E-04} & \multicolumn{1}{c|}{1.110E-04} & 5.843E-05 & 6.325E-05 \\ \hline
\multicolumn{1}{|c|}{\multirow{2}{*}{GNOT}} & Mean & \multicolumn{1}{c|}{1.023E-01} & \multicolumn{1}{c|}{3.958E-02} & \multicolumn{1}{c|}{1.585E-02} & \multicolumn{1}{c|}{6.551E-03} & \multicolumn{1}{c|}{3.398E-03} & \multicolumn{1}{c|}{1.509E-03} & \multicolumn{1}{c|}{7.431E-04} & \multicolumn{1}{c|}{5.171E-04} & 2.677E-04 & 2.241E-04 \\ \cline{2-12} 
\multicolumn{1}{|c|}{} & Std & \multicolumn{1}{c|}{1.641E-02} & \multicolumn{1}{c|}{8.068E-03} & \multicolumn{1}{c|}{3.411E-03} & \multicolumn{1}{c|}{1.165E-03} & \multicolumn{1}{c|}{5.416E-04} & \multicolumn{1}{c|}{2.238E-04} & \multicolumn{1}{c|}{7.435E-05} & \multicolumn{1}{c|}{3.796E-05} & 1.373E-05 & 1.238E-05 \\ \hline
\multicolumn{1}{|c|}{\multirow{2}{*}{Mamba}} & Mean & \multicolumn{1}{c|}{\textbf{1.758E-02}} & \multicolumn{1}{c|}{\textbf{6.081E-03}} & \multicolumn{1}{c|}{\textbf{2.699E-03}} & \multicolumn{1}{c|}{\textbf{1.243E-03}} & \multicolumn{1}{c|}{\textbf{7.350E-04}} & \multicolumn{1}{c|}{\textbf{4.146E-04}} & \multicolumn{1}{c|}{\textbf{2.679E-04}} & \multicolumn{1}{c|}{\textbf{2.157E-04}} & \textbf{1.466E-04} & \textbf{1.332E-04} \\ \cline{2-12} 
\multicolumn{1}{|c|}{} & Std & \multicolumn{1}{c|}{6.335E-03} & \multicolumn{1}{c|}{1.933E-03} & \multicolumn{1}{c|}{7.687E-04} & \multicolumn{1}{c|}{3.262E-04} & \multicolumn{1}{c|}{1.925E-04} & \multicolumn{1}{c|}{9.272E-05} & \multicolumn{1}{c|}{5.148E-05} & \multicolumn{1}{c|}{3.555E-05} & 1.939E-05 & 1.467E-05 \\ \hline
\end{tabular}
\caption{Gravity pendulum Ex+ setting with $l_{train} = 1$ and $l_{test} \in \{0.1,0.2,\cdots,1.0\}$ to quantify the extrapolation error following Zhu et al. \cite{zhu2023reliable}, corresponding to Section \ref{sec:quantify_extrapolation_error_zhu2023reliable}. The table's content is also visualized in Figure \ref{fig:Ex+_Pend}. Here, we train on $l_{train} = 1$, so testing on $l_{test} = 1$ is the independent and identically distributed (IID) test setting while $l_{train} \neq 1$ is the extrapolation testing. The best model is bold.}
\label{tab:Ex+_Pend}
\end{table}

\begin{table}[htbp]
\centering\scriptsize
\begin{tabular}{|cc|c|ccccccccc|}
\hline
\multicolumn{2}{|c|}{AD} & IID & \multicolumn{9}{c|}{OOD Extrapolation: Ex-} \\ \hline
\multicolumn{1}{|c|}{Model} & Metric & 0.1 & \multicolumn{1}{c|}{0.2} & \multicolumn{1}{c|}{0.3} & \multicolumn{1}{c|}{0.4} & \multicolumn{1}{c|}{0.5} & \multicolumn{1}{c|}{0.6} & \multicolumn{1}{c|}{0.7} & \multicolumn{1}{c|}{0.8} & \multicolumn{1}{c|}{0.9} & 1 \\ \hline
\multicolumn{1}{|c|}{\multirow{2}{*}{GRU}} & Mean & 5.145E-04 & \multicolumn{1}{c|}{4.388E-03} & \multicolumn{1}{c|}{1.361E-02} & \multicolumn{1}{c|}{1.547E-02} & \multicolumn{1}{c|}{1.918E-02} & \multicolumn{1}{c|}{2.193E-02} & \multicolumn{1}{c|}{2.354E-02} & \multicolumn{1}{c|}{3.185E-02} & \multicolumn{1}{c|}{3.104E-02} & 2.881E-02 \\ \cline{2-12} 
\multicolumn{1}{|c|}{} & Std & 1.675E-04 & \multicolumn{1}{c|}{1.209E-03} & \multicolumn{1}{c|}{4.441E-03} & \multicolumn{1}{c|}{3.018E-03} & \multicolumn{1}{c|}{3.722E-03} & \multicolumn{1}{c|}{3.392E-03} & \multicolumn{1}{c|}{5.762E-03} & \multicolumn{1}{c|}{5.922E-03} & \multicolumn{1}{c|}{6.153E-03} & 6.761E-03 \\ \hline
\multicolumn{1}{|c|}{\multirow{2}{*}{LSTM}} & Mean & 7.893E-04 & \multicolumn{1}{c|}{7.746E-03} & \multicolumn{1}{c|}{1.517E-02} & \multicolumn{1}{c|}{1.870E-02} & \multicolumn{1}{c|}{2.280E-02} & \multicolumn{1}{c|}{2.637E-02} & \multicolumn{1}{c|}{2.686E-02} & \multicolumn{1}{c|}{3.500E-02} & \multicolumn{1}{c|}{3.466E-02} & 3.178E-02 \\ \cline{2-12} 
\multicolumn{1}{|c|}{} & Std & 3.590E-04 & \multicolumn{1}{c|}{5.503E-03} & \multicolumn{1}{c|}{1.076E-02} & \multicolumn{1}{c|}{1.115E-02} & \multicolumn{1}{c|}{1.295E-02} & \multicolumn{1}{c|}{1.461E-02} & \multicolumn{1}{c|}{1.708E-02} & \multicolumn{1}{c|}{1.931E-02} & \multicolumn{1}{c|}{1.795E-02} & 1.814E-02 \\ \hline
\multicolumn{1}{|c|}{\multirow{2}{*}{DeepONet}} & Mean & 1.135E-03 & \multicolumn{1}{c|}{6.607E-03} & \multicolumn{1}{c|}{1.427E-02} & \multicolumn{1}{c|}{1.668E-02} & \multicolumn{1}{c|}{1.885E-02} & \multicolumn{1}{c|}{2.288E-02} & \multicolumn{1}{c|}{2.379E-02} & \multicolumn{1}{c|}{3.080E-02} & \multicolumn{1}{c|}{3.008E-02} & 2.737E-02 \\ \cline{2-12} 
\multicolumn{1}{|c|}{} & Std & 6.133E-05 & \multicolumn{1}{c|}{9.810E-04} & \multicolumn{1}{c|}{2.086E-03} & \multicolumn{1}{c|}{2.273E-03} & \multicolumn{1}{c|}{2.600E-03} & \multicolumn{1}{c|}{3.181E-03} & \multicolumn{1}{c|}{3.392E-03} & \multicolumn{1}{c|}{4.174E-03} & \multicolumn{1}{c|}{4.074E-03} & 3.787E-03 \\ \hline
\multicolumn{1}{|c|}{\multirow{2}{*}{FNO}} & Mean & 5.158E-02 & \multicolumn{1}{c|}{4.985E-02} & \multicolumn{1}{c|}{5.726E-02} & \multicolumn{1}{c|}{6.365E-02} & \multicolumn{1}{c|}{7.224E-02} & \multicolumn{1}{c|}{7.919E-02} & \multicolumn{1}{c|}{8.325E-02} & \multicolumn{1}{c|}{1.008E-01} & \multicolumn{1}{c|}{9.730E-02} & 1.170E-01 \\ \cline{2-12} 
\multicolumn{1}{|c|}{} & Std & 8.791E-04 & \multicolumn{1}{c|}{3.008E-03} & \multicolumn{1}{c|}{6.064E-03} & \multicolumn{1}{c|}{7.578E-03} & \multicolumn{1}{c|}{8.572E-03} & \multicolumn{1}{c|}{8.998E-03} & \multicolumn{1}{c|}{8.328E-03} & \multicolumn{1}{c|}{9.023E-03} & \multicolumn{1}{c|}{1.004E-02} & 1.107E-02 \\ \hline
\multicolumn{1}{|c|}{\multirow{2}{*}{LNO}} & Mean & 1.679E-03 & \multicolumn{1}{c|}{2.103E-02} & \multicolumn{1}{c|}{5.276E-02} & \multicolumn{1}{c|}{6.218E-02} & \multicolumn{1}{c|}{7.448E-02} & \multicolumn{1}{c|}{8.865E-02} & \multicolumn{1}{c|}{9.384E-02} & \multicolumn{1}{c|}{1.191E-01} & \multicolumn{1}{c|}{1.124E-01} & 1.055E-01 \\ \cline{2-12} 
\multicolumn{1}{|c|}{} & Std & 1.609E-04 & \multicolumn{1}{c|}{8.680E-03} & \multicolumn{1}{c|}{1.922E-02} & \multicolumn{1}{c|}{2.044E-02} & \multicolumn{1}{c|}{2.413E-02} & \multicolumn{1}{c|}{2.766E-02} & \multicolumn{1}{c|}{3.131E-02} & \multicolumn{1}{c|}{3.564E-02} & \multicolumn{1}{c|}{3.749E-02} & 3.263E-02 \\ \hline
\multicolumn{1}{|c|}{\multirow{2}{*}{Transformer}} & Mean & 3.220E-02 & \multicolumn{1}{c|}{2.431E-02} & \multicolumn{1}{c|}{2.547E-02} & \multicolumn{1}{c|}{2.623E-02} & \multicolumn{1}{c|}{2.805E-02} & \multicolumn{1}{c|}{2.969E-02} & \multicolumn{1}{c|}{3.061E-02} & \multicolumn{1}{c|}{3.453E-02} & \multicolumn{1}{c|}{3.402E-02} & 3.262E-02 \\ \cline{2-12} 
\multicolumn{1}{|c|}{} & Std & 3.003E-04 & \multicolumn{1}{c|}{1.651E-04} & \multicolumn{1}{c|}{8.204E-04} & \multicolumn{1}{c|}{8.835E-04} & \multicolumn{1}{c|}{1.904E-03} & \multicolumn{1}{c|}{1.275E-03} & \multicolumn{1}{c|}{2.416E-03} & \multicolumn{1}{c|}{3.093E-03} & \multicolumn{1}{c|}{3.925E-03} & 3.595E-03 \\ \hline
\multicolumn{1}{|c|}{\multirow{2}{*}{Oformer-V}} & Mean & 6.087E-04 & \multicolumn{1}{c|}{1.183E-03} & \multicolumn{1}{c|}{2.366E-03} & \multicolumn{1}{c|}{2.995E-03} & \multicolumn{1}{c|}{3.703E-03} & \multicolumn{1}{c|}{4.404E-03} & \multicolumn{1}{c|}{4.291E-03} & \multicolumn{1}{c|}{5.937E-03} & \multicolumn{1}{c|}{6.083E-03} & 5.339E-03 \\ \cline{2-12} 
\multicolumn{1}{|c|}{} & Std & 3.744E-05 & \multicolumn{1}{c|}{5.219E-04} & \multicolumn{1}{c|}{1.361E-03} & \multicolumn{1}{c|}{1.684E-03} & \multicolumn{1}{c|}{2.096E-03} & \multicolumn{1}{c|}{2.574E-03} & \multicolumn{1}{c|}{2.709E-03} & \multicolumn{1}{c|}{3.598E-03} & \multicolumn{1}{c|}{3.677E-03} & 3.385E-03 \\ \hline
\multicolumn{1}{|c|}{\multirow{2}{*}{Oformer-G}} & Mean & 6.431E-04 & \multicolumn{1}{c|}{2.932E-04} & \multicolumn{1}{c|}{2.923E-04} & \multicolumn{1}{c|}{2.897E-04} & \multicolumn{1}{c|}{3.218E-04} & \multicolumn{1}{c|}{3.316E-04} & \multicolumn{1}{c|}{3.254E-04} & \multicolumn{1}{c|}{3.868E-04} & \multicolumn{1}{c|}{3.830E-04} & 3.618E-04 \\ \cline{2-12} 
\multicolumn{1}{|c|}{} & Std & 7.398E-05 & \multicolumn{1}{c|}{2.669E-05} & \multicolumn{1}{c|}{1.716E-05} & \multicolumn{1}{c|}{1.651E-05} & \multicolumn{1}{c|}{1.122E-05} & \multicolumn{1}{c|}{4.508E-05} & \multicolumn{1}{c|}{3.668E-05} & \multicolumn{1}{c|}{6.187E-05} & \multicolumn{1}{c|}{5.957E-05} & 5.335E-05 \\ \hline
\multicolumn{1}{|c|}{\multirow{2}{*}{Oformer-F}} & Mean & 6.310E-04 & \multicolumn{1}{c|}{2.573E-04} & \multicolumn{1}{c|}{2.341E-04} & \multicolumn{1}{c|}{2.115E-04} & \multicolumn{1}{c|}{2.250E-04} & \multicolumn{1}{c|}{2.129E-04} & \multicolumn{1}{c|}{2.065E-04} & \multicolumn{1}{c|}{2.301E-04} & \multicolumn{1}{c|}{2.286E-04} & 2.184E-04 \\ \cline{2-12} 
\multicolumn{1}{|c|}{} & Std & 1.362E-04 & \multicolumn{1}{c|}{3.097E-05} & \multicolumn{1}{c|}{2.657E-05} & \multicolumn{1}{c|}{3.138E-05} & \multicolumn{1}{c|}{2.668E-05} & \multicolumn{1}{c|}{3.435E-05} & \multicolumn{1}{c|}{3.221E-05} & \multicolumn{1}{c|}{3.509E-05} & \multicolumn{1}{c|}{3.426E-05} & 3.533E-05 \\ \hline
\multicolumn{1}{|c|}{\multirow{2}{*}{GNOT}} & Mean & 9.961E-04 & \multicolumn{1}{c|}{5.128E-04} & \multicolumn{1}{c|}{7.509E-04} & \multicolumn{1}{c|}{8.636E-04} & \multicolumn{1}{c|}{1.103E-03} & \multicolumn{1}{c|}{1.196E-03} & \multicolumn{1}{c|}{1.217E-03} & \multicolumn{1}{c|}{1.640E-03} & \multicolumn{1}{c|}{1.702E-03} & 1.541E-03 \\ \cline{2-12} 
\multicolumn{1}{|c|}{} & Std & 9.282E-05 & \multicolumn{1}{c|}{4.917E-05} & \multicolumn{1}{c|}{1.052E-04} & \multicolumn{1}{c|}{2.285E-04} & \multicolumn{1}{c|}{3.337E-04} & \multicolumn{1}{c|}{4.293E-04} & \multicolumn{1}{c|}{3.227E-04} & \multicolumn{1}{c|}{5.873E-04} & \multicolumn{1}{c|}{6.608E-04} & 4.580E-04 \\ \hline
\multicolumn{1}{|c|}{\multirow{2}{*}{Mamba}} & Mean & \textbf{1.086E-04} & \multicolumn{1}{c|}{\textbf{9.311E-05}} & \multicolumn{1}{c|}{\textbf{1.034E-04}} & \multicolumn{1}{c|}{\textbf{1.029E-04}} & \multicolumn{1}{c|}{\textbf{1.461E-04}} & \multicolumn{1}{c|}{\textbf{1.156E-04}} & \multicolumn{1}{c|}{\textbf{1.111E-04}} & \multicolumn{1}{c|}{\textbf{1.410E-04}} & \multicolumn{1}{c|}{\textbf{1.626E-04}} & \textbf{1.288E-04} \\ \cline{2-12} 
\multicolumn{1}{|c|}{} & Std & 1.485E-05 & \multicolumn{1}{c|}{2.302E-05} & \multicolumn{1}{c|}{2.816E-05} & \multicolumn{1}{c|}{2.869E-05} & \multicolumn{1}{c|}{8.059E-05} & \multicolumn{1}{c|}{3.776E-05} & \multicolumn{1}{c|}{5.116E-05} & \multicolumn{1}{c|}{4.718E-05} & \multicolumn{1}{c|}{5.692E-05} & 6.369E-05 \\ \hline
\end{tabular}
\caption{Anti-derivative operator Ex- setting with $l_{train} = 0.1$ and $l_{test} \in \{0.1,0.2,\cdots,1.0\}$ to quantify the extrapolation error following Zhu et al. \cite{zhu2023reliable}, corresponding to Section \ref{sec:quantify_extrapolation_error_zhu2023reliable}. The table's content is also visualized in Figure \ref{fig:Ex-_AD}. Here, we train on $l_{train} = 0.1$, so testing on $l_{test} = 0.1$ is the independent and identically distributed (IID) test setting while $l_{train} \neq 0.1$ is the extrapolation testing. The best model is bold.}
\label{tab:Ex-_AD}
\end{table}

\begin{table}[htbp]
\centering\scriptsize
\begin{tabular}{|cc|ccccccccc|c|}
\hline
\multicolumn{2}{|c|}{AD} & \multicolumn{9}{c|}{OOD Extrapolation: Ex+} & IID \\ \hline
\multicolumn{1}{|c|}{Model} & Metric & \multicolumn{1}{c|}{0.1} & \multicolumn{1}{c|}{0.2} & \multicolumn{1}{c|}{0.3} & \multicolumn{1}{c|}{0.4} & \multicolumn{1}{c|}{0.5} & \multicolumn{1}{c|}{0.6} & \multicolumn{1}{c|}{0.7} & \multicolumn{1}{c|}{0.8} & 0.9 & 1 \\ \hline
\multicolumn{1}{|c|}{\multirow{2}{*}{GRU}} & Mean & \multicolumn{1}{c|}{2.751E-02} & \multicolumn{1}{c|}{1.182E-02} & \multicolumn{1}{c|}{5.528E-03} & \multicolumn{1}{c|}{2.718E-03} & \multicolumn{1}{c|}{1.775E-03} & \multicolumn{1}{c|}{9.128E-04} & \multicolumn{1}{c|}{5.157E-04} & \multicolumn{1}{c|}{1.317E-03} & 1.242E-03 & 1.379E-03 \\ \cline{2-12} 
\multicolumn{1}{|c|}{} & Std & \multicolumn{1}{c|}{1.425E-02} & \multicolumn{1}{c|}{6.274E-03} & \multicolumn{1}{c|}{2.415E-03} & \multicolumn{1}{c|}{1.005E-03} & \multicolumn{1}{c|}{4.525E-04} & \multicolumn{1}{c|}{1.638E-04} & \multicolumn{1}{c|}{9.522E-05} & \multicolumn{1}{c|}{6.882E-04} & 3.052E-04 & 9.488E-04 \\ \hline
\multicolumn{1}{|c|}{\multirow{2}{*}{LSTM}} & Mean & \multicolumn{1}{c|}{2.919E-02} & \multicolumn{1}{c|}{1.221E-02} & \multicolumn{1}{c|}{5.766E-03} & \multicolumn{1}{c|}{2.948E-03} & \multicolumn{1}{c|}{2.014E-03} & \multicolumn{1}{c|}{1.111E-03} & \multicolumn{1}{c|}{5.115E-04} & \multicolumn{1}{c|}{8.236E-04} & 1.919E-03 & 1.023E-03 \\ \cline{2-12} 
\multicolumn{1}{|c|}{} & Std & \multicolumn{1}{c|}{5.071E-03} & \multicolumn{1}{c|}{1.300E-03} & \multicolumn{1}{c|}{3.408E-04} & \multicolumn{1}{c|}{2.638E-04} & \multicolumn{1}{c|}{4.978E-04} & \multicolumn{1}{c|}{1.839E-04} & \multicolumn{1}{c|}{1.073E-05} & \multicolumn{1}{c|}{9.685E-05} & 7.694E-04 & 5.447E-04 \\ \hline
\multicolumn{1}{|c|}{\multirow{2}{*}{DeepONet}} & Mean & \multicolumn{1}{c|}{2.286E-01} & \multicolumn{1}{c|}{1.068E-01} & \multicolumn{1}{c|}{4.983E-02} & \multicolumn{1}{c|}{2.191E-02} & \multicolumn{1}{c|}{8.988E-03} & \multicolumn{1}{c|}{4.229E-03} & \multicolumn{1}{c|}{1.754E-03} & \multicolumn{1}{c|}{1.569E-03} & 3.319E-03 & 2.724E-03 \\ \cline{2-12} 
\multicolumn{1}{|c|}{} & Std & \multicolumn{1}{c|}{1.583E-02} & \multicolumn{1}{c|}{5.847E-03} & \multicolumn{1}{c|}{3.733E-03} & \multicolumn{1}{c|}{1.864E-03} & \multicolumn{1}{c|}{6.847E-04} & \multicolumn{1}{c|}{1.432E-04} & \multicolumn{1}{c|}{1.285E-04} & \multicolumn{1}{c|}{2.862E-04} & 4.322E-04 & 2.817E-04 \\ \hline
\multicolumn{1}{|c|}{\multirow{2}{*}{FNO}} & Mean & \multicolumn{1}{c|}{8.377E+00} & \multicolumn{1}{c|}{2.224E+00} & \multicolumn{1}{c|}{8.190E-01} & \multicolumn{1}{c|}{3.350E-01} & \multicolumn{1}{c|}{1.474E-01} & \multicolumn{1}{c|}{7.468E-02} & \multicolumn{1}{c|}{4.267E-02} & \multicolumn{1}{c|}{2.874E-02} & 2.088E-02 & 2.121E-02 \\ \cline{2-12} 
\multicolumn{1}{|c|}{} & Std & \multicolumn{1}{c|}{1.056E+00} & \multicolumn{1}{c|}{5.543E-01} & \multicolumn{1}{c|}{1.901E-01} & \multicolumn{1}{c|}{5.526E-02} & \multicolumn{1}{c|}{1.419E-02} & \multicolumn{1}{c|}{5.552E-03} & \multicolumn{1}{c|}{5.784E-03} & \multicolumn{1}{c|}{5.938E-03} & 6.394E-03 & 5.104E-03 \\ \hline
\multicolumn{1}{|c|}{\multirow{2}{*}{LNO}} & Mean & \multicolumn{1}{c|}{\textbf{1.721E-02}} & \multicolumn{1}{c|}{8.301E-03} & \multicolumn{1}{c|}{4.883E-03} & \multicolumn{1}{c|}{2.734E-03} & \multicolumn{1}{c|}{1.676E-03} & \multicolumn{1}{c|}{1.485E-03} & \multicolumn{1}{c|}{6.643E-04} & \multicolumn{1}{c|}{9.606E-04} & 2.757E-03 & 2.011E-03 \\ \cline{2-12} 
\multicolumn{1}{|c|}{} & Std & \multicolumn{1}{c|}{4.271E-03} & \multicolumn{1}{c|}{2.477E-03} & \multicolumn{1}{c|}{1.695E-03} & \multicolumn{1}{c|}{9.095E-04} & \multicolumn{1}{c|}{6.982E-04} & \multicolumn{1}{c|}{5.560E-04} & \multicolumn{1}{c|}{2.163E-04} & \multicolumn{1}{c|}{3.904E-04} & 1.260E-03 & 9.660E-04 \\ \hline
\multicolumn{1}{|c|}{\multirow{2}{*}{Transformer}} & Mean & \multicolumn{1}{c|}{6.392E-01} & \multicolumn{1}{c|}{2.523E-01} & \multicolumn{1}{c|}{1.119E-01} & \multicolumn{1}{c|}{5.659E-02} & \multicolumn{1}{c|}{3.253E-02} & \multicolumn{1}{c|}{2.225E-02} & \multicolumn{1}{c|}{1.714E-02} & \multicolumn{1}{c|}{1.487E-02} & 1.349E-02 & 1.260E-02 \\ \cline{2-12} 
\multicolumn{1}{|c|}{} & Std & \multicolumn{1}{c|}{1.094E-02} & \multicolumn{1}{c|}{3.294E-03} & \multicolumn{1}{c|}{1.744E-03} & \multicolumn{1}{c|}{2.322E-03} & \multicolumn{1}{c|}{1.528E-03} & \multicolumn{1}{c|}{1.184E-03} & \multicolumn{1}{c|}{1.184E-03} & \multicolumn{1}{c|}{1.219E-03} & 1.046E-03 & 1.021E-03 \\ \hline
\multicolumn{1}{|c|}{\multirow{2}{*}{Oformer-V}} & Mean & \multicolumn{1}{c|}{4.515E-01} & \multicolumn{1}{c|}{1.485E-01} & \multicolumn{1}{c|}{4.930E-02} & \multicolumn{1}{c|}{1.877E-02} & \multicolumn{1}{c|}{7.960E-03} & \multicolumn{1}{c|}{3.741E-03} & \multicolumn{1}{c|}{1.940E-03} & \multicolumn{1}{c|}{1.141E-03} & 8.460E-04 & 5.243E-04 \\ \cline{2-12} 
\multicolumn{1}{|c|}{} & Std & \multicolumn{1}{c|}{5.774E-02} & \multicolumn{1}{c|}{3.174E-02} & \multicolumn{1}{c|}{1.342E-02} & \multicolumn{1}{c|}{5.487E-03} & \multicolumn{1}{c|}{2.479E-03} & \multicolumn{1}{c|}{1.187E-03} & \multicolumn{1}{c|}{6.441E-04} & \multicolumn{1}{c|}{3.338E-04} & 2.280E-04 & 1.732E-04 \\ \hline
\multicolumn{1}{|c|}{\multirow{2}{*}{Oformer-G}} & Mean & \multicolumn{1}{c|}{2.647E-01} & \multicolumn{1}{c|}{7.146E-02} & \multicolumn{1}{c|}{2.095E-02} & \multicolumn{1}{c|}{7.286E-03} & \multicolumn{1}{c|}{2.848E-03} & \multicolumn{1}{c|}{1.278E-03} & \multicolumn{1}{c|}{6.462E-04} & \multicolumn{1}{c|}{4.302E-04} & 2.877E-04 & 1.816E-04 \\ \cline{2-12} 
\multicolumn{1}{|c|}{} & Std & \multicolumn{1}{c|}{3.578E-02} & \multicolumn{1}{c|}{4.720E-03} & \multicolumn{1}{c|}{6.421E-04} & \multicolumn{1}{c|}{6.479E-04} & \multicolumn{1}{c|}{3.276E-04} & \multicolumn{1}{c|}{1.686E-04} & \multicolumn{1}{c|}{9.763E-05} & \multicolumn{1}{c|}{6.990E-05} & 4.816E-05 & 2.007E-05 \\ \hline
\multicolumn{1}{|c|}{\multirow{2}{*}{Oformer-F}} & Mean & \multicolumn{1}{c|}{2.314E-01} & \multicolumn{1}{c|}{6.783E-02} & \multicolumn{1}{c|}{2.109E-02} & \multicolumn{1}{c|}{7.714E-03} & \multicolumn{1}{c|}{3.195E-03} & \multicolumn{1}{c|}{1.490E-03} & \multicolumn{1}{c|}{7.633E-04} & \multicolumn{1}{c|}{5.409E-04} & 3.762E-04 & 2.156E-04 \\ \cline{2-12} 
\multicolumn{1}{|c|}{} & Std & \multicolumn{1}{c|}{3.078E-02} & \multicolumn{1}{c|}{3.636E-03} & \multicolumn{1}{c|}{1.555E-03} & \multicolumn{1}{c|}{7.555E-04} & \multicolumn{1}{c|}{3.062E-04} & \multicolumn{1}{c|}{1.596E-04} & \multicolumn{1}{c|}{9.401E-05} & \multicolumn{1}{c|}{3.987E-05} & 1.575E-05 & 1.008E-05 \\ \hline
\multicolumn{1}{|c|}{\multirow{2}{*}{GNOT}} & Mean & \multicolumn{1}{c|}{1.552E-01} & \multicolumn{1}{c|}{5.804E-02} & \multicolumn{1}{c|}{2.163E-02} & \multicolumn{1}{c|}{8.720E-03} & \multicolumn{1}{c|}{3.855E-03} & \multicolumn{1}{c|}{1.753E-03} & \multicolumn{1}{c|}{8.726E-04} & \multicolumn{1}{c|}{6.241E-04} & 3.887E-04 & 2.482E-04 \\ \cline{2-12} 
\multicolumn{1}{|c|}{} & Std & \multicolumn{1}{c|}{4.856E-03} & \multicolumn{1}{c|}{3.312E-03} & \multicolumn{1}{c|}{1.625E-03} & \multicolumn{1}{c|}{9.031E-04} & \multicolumn{1}{c|}{6.404E-04} & \multicolumn{1}{c|}{2.752E-04} & \multicolumn{1}{c|}{1.442E-04} & \multicolumn{1}{c|}{4.299E-05} & 5.183E-05 & 2.275E-05 \\ \hline
\multicolumn{1}{|c|}{\multirow{2}{*}{Mamba}} & Mean & \multicolumn{1}{c|}{3.091E-02} & \multicolumn{1}{c|}{\textbf{8.101E-03}} & \multicolumn{1}{c|}{\textbf{3.108E-03}} & \multicolumn{1}{c|}{\textbf{1.450E-03}} & \multicolumn{1}{c|}{\textbf{7.212E-04}} & \multicolumn{1}{c|}{\textbf{3.993E-04}} & \multicolumn{1}{c|}{\textbf{2.571E-04}} & \multicolumn{1}{c|}{\textbf{2.233E-04}} & \textbf{1.733E-04} & \textbf{1.352E-04} \\ \cline{2-12} 
\multicolumn{1}{|c|}{} & Std & \multicolumn{1}{c|}{2.409E-03} & \multicolumn{1}{c|}{2.090E-03} & \multicolumn{1}{c|}{1.030E-03} & \multicolumn{1}{c|}{4.888E-04} & \multicolumn{1}{c|}{1.479E-04} & \multicolumn{1}{c|}{7.192E-05} & \multicolumn{1}{c|}{2.083E-05} & \multicolumn{1}{c|}{1.281E-05} & 1.876E-05 & 9.883E-06 \\ \hline
\end{tabular}
\caption{Anti-derivative operator Ex+ setting with $l_{train} = 1$ and $l_{test} \in \{0.1,0.2,\cdots,1.0\}$ to quantify the extrapolation error following Zhu et al. \cite{zhu2023reliable}, corresponding to Section \ref{sec:quantify_extrapolation_error_zhu2023reliable}. The table's content is also visualized in Figure \ref{fig:Ex+_AD}. Here, we train on $l_{train} = 1$, so testing on $l_{test} = 1$ is the independent and identically distributed (IID) test setting while $l_{train} \neq 1$ is the extrapolation testing. The best model is bold.}
\label{tab:Ex+_AD}
\end{table}

\newpage

\bibliographystyle{plain}
\bibliography{references}

\begin{thebibliography}{10}

\bibitem{ba2016layer}
Jimmy~Lei Ba, Jamie~Ryan Kiros, and Geoffrey~E Hinton.
\newblock Layer normalization.
\newblock {\em arXiv preprint arXiv:1607.06450}, 2016.

\bibitem{bengio1994learning}
Yoshua Bengio, Patrice Simard, and Paolo Frasconi.
\newblock Learning long-term dependencies with gradient descent is difficult.
\newblock {\em IEEE transactions on neural networks}, 5(2):157--166, 1994.

\bibitem{brown2020language}
Tom Brown, Benjamin Mann, Nick Ryder, Melanie Subbiah, Jared~D Kaplan, Prafulla Dhariwal, Arvind Neelakantan, Pranav Shyam, Girish Sastry, Amanda Askell, et~al.
\newblock Language models are few-shot learners.
\newblock {\em Advances in neural information processing systems}, 33:1877--1901, 2020.

\bibitem{cao2023lno}
Qianying Cao, Somdatta Goswami, and George~Em Karniadakis.
\newblock Laplace neural operator for solving differential equations.
\newblock {\em Nature Machine Intelligence}, pages 1--10, 2024.

\bibitem{cao2021choose}
Shuhao Cao.
\newblock Choose a \uppercase{T}ransformer: Fourier or \uppercase{G}alerkin.
\newblock {\em Advances in Neural Information Processing Systems}, 34:24924--24940, 2021.

\bibitem{chen1995universal}
Tianping Chen and Hong Chen.
\newblock Universal approximation to nonlinear operators by neural networks with arbitrary activation functions and its application to dynamical systems.
\newblock {\em IEEE transactions on neural networks}, 6(4):911--917, 1995.

\bibitem{cho2014learning}
Kyunghyun Cho, Bart van Merri{\"e}nboer, {\c{C}}a{\u{g}}lar Gul{\c{c}}ehre, Dzmitry Bahdanau, Fethi Bougares, Holger Schwenk, and Yoshua Bengio.
\newblock Learning phrase representations using \uppercase{RNN} encoder--decoder for statistical machine translation.
\newblock In {\em Proceedings of the 2014 Conference on Empirical Methods in Natural Language Processing (EMNLP)}, pages 1724--1734, 2014.

\bibitem{dao2024mamba2}
Tri Dao and Albert Gu.
\newblock Transformers are \uppercase{SSM}s: Generalized models and efficient algorithms through structured state space duality.
\newblock {\em arXiv preprint arXiv:2405.21060}, 2024.

\bibitem{cminns}
Nazanin~Ahmadi Daryakenari, Shupeng Wang, and George Karniadakis.
\newblock Cminns: Compartment model informed neural networks—unlocking drug dynamics.
\newblock {\em Computers in Biology and Medicine}, 184:109392, 2025.

\bibitem{deng2022approximation}
Beichuan Deng, Yeonjong Shin, Lu~Lu, Zhongqiang Zhang, and George~Em Karniadakis.
\newblock Approximation rates of \uppercase{D}eep\uppercase{ON}ets for learning operators arising from advection--diffusion equations.
\newblock {\em Neural Networks}, 153:411--426, 2022.

\bibitem{devlin2018bert}
Jacob Devlin, Ming-Wei Chang, Kenton Lee, and Kristina Toutanova.
\newblock Bert: Pre-training of deep bidirectional transformers for language understanding.
\newblock {\em arXiv preprint arXiv:1810.04805}, 2018.

\bibitem{franco2023approximation}
Nicola~Rares Franco, Stefania Fresca, Andrea Manzoni, and Paolo Zunino.
\newblock Approximation bounds for convolutional neural networks in operator learning.
\newblock {\em Neural Networks}, 161:129--141, 2023.

\bibitem{fu2022hungry}
Daniel~Y Fu, Tri Dao, Khaled~K Saab, Armin~W Thomas, Atri Rudra, and Christopher R{\'e}.
\newblock Hungry hungry hippos: Towards language modeling with state space models.
\newblock {\em arXiv preprint arXiv:2212.14052}, 2022.

\bibitem{funahashi1993approximation}
Ken-ichi Funahashi and Yuichi Nakamura.
\newblock Approximation of dynamical systems by continuous time recurrent neural networks.
\newblock {\em Neural networks}, 6(6):801--806, 1993.

\bibitem{goswami2023physics}
Somdatta Goswami, Aniruddha Bora, Yue Yu, and George~Em Karniadakis.
\newblock Physics-informed deep neural operator networks.
\newblock In {\em Machine Learning in Modeling and Simulation: Methods and Applications}, pages 219--254. Springer, 2023.

\bibitem{goswami2023learning}
Somdatta Goswami, Ameya~D Jagtap, Hessam Babaee, Bryan~T Susi, and George~Em Karniadakis.
\newblock Learning stiff chemical kinetics using extended deep neural operators.
\newblock {\em Computer Methods in Applied Mechanics and Engineering}, 419:116674, 2024.

\bibitem{goswami2022physics}
Somdatta Goswami, Minglang Yin, Yue Yu, and George~Em Karniadakis.
\newblock A physics-informed variational \uppercase{D}eep\uppercase{ON}et for predicting crack path in quasi-brittle materials.
\newblock {\em Computer Methods in Applied Mechanics and Engineering}, 391:114587, 2022.

\bibitem{gu2023mamba}
Albert Gu and Tri Dao.
\newblock Mamba: Linear-time sequence modeling with selective state spaces.
\newblock {\em arXiv preprint arXiv:2312.00752}, 2023.

\bibitem{gu2021efficiently}
Albert Gu, Karan Goel, and Christopher R{\'e}.
\newblock Efficiently modeling long sequences with structured state spaces.
\newblock {\em arXiv preprint arXiv:2111.00396}, 2021.

\bibitem{hao2023gnot}
Zhongkai Hao, Zhengyi Wang, Hang Su, Chengyang Ying, Yinpeng Dong, Songming Liu, Ze~Cheng, Jian Song, and Jun Zhu.
\newblock \uppercase{GNOT}: A general neural operator transformer for operator learning.
\newblock In {\em International Conference on Machine Learning}, pages 12556--12569. PMLR, 2023.

\bibitem{he2016deep}
Kaiming He, Xiangyu Zhang, Shaoqing Ren, and Jian Sun.
\newblock Deep residual learning for image recognition.
\newblock In {\em Proceedings of the IEEE conference on computer vision and pattern recognition}, pages 770--778, 2016.

\bibitem{hochreiter1997long}
Sepp Hochreiter and J{\"u}rgen Schmidhuber.
\newblock Long short-term memory.
\newblock {\em Neural computation}, 9(8):1735--1780, 1997.

\bibitem{hu2023augmented}
Zheyuan Hu, Ameya~D Jagtap, George~Em Karniadakis, and Kenji Kawaguchi.
\newblock Augmented physics-informed neural networks (\uppercase{APINN}s): A gating network-based soft domain decomposition methodology.
\newblock {\em Engineering Applications of Artificial Intelligence}, 126:107183, 2023.

\bibitem{hu2023tackling}
Zheyuan Hu, Khemraj Shukla, George~Em Karniadakis, and Kenji Kawaguchi.
\newblock Tackling the curse of dimensionality with physics-informed neural networks.
\newblock {\em Neural Networks}, page 106369, 2024.

\bibitem{karniadakis2021physics}
George~Em Karniadakis, Ioannis~G Kevrekidis, Lu~Lu, Paris Perdikaris, Sifan Wang, and Liu Yang.
\newblock Physics-informed machine learning.
\newblock {\em Nature Reviews Physics}, 3(6):422--440, 2021.

\bibitem{katharopoulos2020transformers}
Angelos Katharopoulos, Apoorv Vyas, Nikolaos Pappas, and Fran{\c{c}}ois Fleuret.
\newblock Transformers are \uppercase{RNN}s: Fast autoregressive transformers with linear attention.
\newblock In {\em International conference on machine learning}, pages 5156--5165. PMLR, 2020.

\bibitem{kingma2014adam}
Diederik~P Kingma and Jimmy Ba.
\newblock Adam: A method for stochastic optimization.
\newblock {\em International Conference on Learning Representations}, 2015.

\bibitem{litransformer}
Zijie Li, Kazem Meidani, and Amir~Barati Farimani.
\newblock Transformer for partial differential equations’ operator learning.
\newblock {\em Transactions on Machine Learning Research}, 2023.

\bibitem{li2020fourier}
Zongyi Li, Nikola~Borislavov Kovachki, Kamyar Azizzadenesheli, Burigede liu, Kaushik Bhattacharya, Andrew Stuart, and Anima Anandkumar.
\newblock Fourier neural operator for parametric partial differential equations.
\newblock In {\em International Conference on Learning Representations}, 2021.

\bibitem{li2021learning}
Zongyi Li, Miguel Liu-Schiaffini, Nikola Kovachki, Burigede Liu, Kamyar Azizzadenesheli, Kaushik Bhattacharya, Andrew Stuart, and Anima Anandkumar.
\newblock Learning dissipative dynamics in chaotic systems.
\newblock {\em arXiv preprint arXiv:2106.06898}, 2021.

\bibitem{liu2021swin}
Ze~Liu, Yutong Lin, Yue Cao, Han Hu, Yixuan Wei, Zheng Zhang, Stephen Lin, and Baining Guo.
\newblock Swin transformer: Hierarchical vision transformer using shifted windows.
\newblock In {\em Proceedings of the IEEE/CVF international conference on computer vision}, pages 10012--10022, 2021.

\bibitem{lorenz1963deterministic}
Edward~N Lorenz.
\newblock Deterministic nonperiodic flow.
\newblock {\em Journal of atmospheric sciences}, 20(2):130--141, 1963.

\bibitem{lu2019deeponet}
Lu~Lu, Pengzhan Jin, Guofei Pang, Zhongqiang Zhang, and George~Em Karniadakis.
\newblock Learning nonlinear operators via \uppercase{D}eep\uppercase{ON}et based on the universal approximation theorem of operators.
\newblock {\em Nature Machine Intelligence}, 3(3):218--229, 2021.

\bibitem{lu2021comprehensive}
Lu~Lu, Xuhui Meng, Shengze Cai, Zhiping Mao, Somdatta Goswami, Zhongqiang Zhang, and George~Em Karniadakis.
\newblock A comprehensive and fair comparison of two neural operators (with practical extensions) based on fair data.
\newblock {\em Computer Methods in Applied Mechanics and Engineering}, 393:114778, 2022.

\bibitem{mehta2022long}
Harsh Mehta, Ankit Gupta, Ashok Cutkosky, and Behnam Neyshabur.
\newblock Long range language modeling via gated state spaces.
\newblock {\em arXiv preprint arXiv:2206.13947}, 2022.

\bibitem{michalowska2023neural}
Katarzyna Micha{\l}owska, Somdatta Goswami, George~Em Karniadakis, and Signe Riemer-S{\o}rensen.
\newblock Neural operator learning for long-time integration in dynamical systems with recurrent neural networks.
\newblock {\em arXiv preprint arXiv:2303.02243}, 2023.

\bibitem{nguyen2024hyenadna}
Eric Nguyen, Michael Poli, Marjan Faizi, Armin Thomas, Michael Wornow, Callum Birch-Sykes, Stefano Massaroli, Aman Patel, Clayton Rabideau, Yoshua Bengio, et~al.
\newblock Hyenadna: Long-range genomic sequence modeling at single nucleotide resolution.
\newblock {\em Advances in neural information processing systems}, 36, 2024.

\bibitem{ovadia2023ditto}
Oded Ovadia, Eli Turkel, Adar Kahana, and George~Em Karniadakis.
\newblock Ditto: Diffusion-inspired temporal transformer operator.
\newblock {\em arXiv preprint arXiv:2307.09072}, 2023.

\bibitem{pascanu2013difficulty}
Razvan Pascanu, Tomas Mikolov, and Yoshua Bengio.
\newblock On the difficulty of training recurrent neural networks.
\newblock In {\em International conference on machine learning}, pages 1310--1318. Pmlr, 2013.

\bibitem{patil2023hno}
Saurabh Patil, Zijie Li, and Amir~Barati Farimani.
\newblock \uppercase{HNO}: Hyena neural operator for solving \uppercase{PDE}s.
\newblock {\em arXiv preprint arXiv:2306.16524}, 2023.

\bibitem{peng2023rwkv}
Bo~Peng, Eric Alcaide, Quentin Anthony, Alon Albalak, Samuel Arcadinho, Stella Biderman, Huanqi Cao, Xin Cheng, Michael Chung, Matteo Grella, et~al.
\newblock \uppercase{RWKV}: Reinventing \uppercase{RNN}s for the transformer era.
\newblock {\em arXiv preprint arXiv:2305.13048}, 2023.

\bibitem{pmlr-v202-poli23a}
Michael Poli, Stefano Massaroli, Eric Nguyen, Daniel~Y Fu, Tri Dao, Stephen Baccus, Yoshua Bengio, Stefano Ermon, and Christopher Re.
\newblock Hyena hierarchy: Towards larger convolutional language models.
\newblock In Andreas Krause, Emma Brunskill, Kyunghyun Cho, Barbara Engelhardt, Sivan Sabato, and Jonathan Scarlett, editors, {\em Proceedings of the 40th International Conference on Machine Learning}, volume 202 of {\em Proceedings of Machine Learning Research}, pages 28043--28078. PMLR, 23--29 Jul 2023.

\bibitem{raissi2019physics}
Maziar Raissi, Paris Perdikaris, and George~E Karniadakis.
\newblock Physics-informed neural networks: A deep learning framework for solving forward and inverse problems involving nonlinear partial differential equations.
\newblock {\em Journal of Computational Physics}, 378:686--707, 2019.

\bibitem{robinson2022physics}
Haakon Robinson, Suraj Pawar, Adil Rasheed, and Omer San.
\newblock Physics guided neural networks for modelling of non-linear dynamics.
\newblock {\em Neural Networks}, 154:333--345, 2022.

\bibitem{saha2021physics}
Priyabrata Saha, Saurabh Dash, and Saibal Mukhopadhyay.
\newblock Physics-incorporated convolutional recurrent neural networks for source identification and forecasting of dynamical systems.
\newblock {\em Neural Networks}, 144:359--371, 2021.

\bibitem{shih2024transformers}
Benjamin Shih, Ahmad Peyvan, Zhongqiang Zhang, and George~Em Karniadakis.
\newblock Transformers as neural operators for solutions of differential equations with finite regularity.
\newblock {\em arXiv preprint arXiv:2405.19166}, 2024.

\bibitem{shukla2023deep}
Khemraj Shukla, Vivek Oommen, Ahmad Peyvan, Michael Penwarden, Luis Bravo, Anindya Ghoshal, Robert~M Kirby, and George~Em Karniadakis.
\newblock Deep neural operators can serve as accurate surrogates for shape optimization: a case study for airfoils.
\newblock {\em arXiv preprint arXiv:2302.00807}, 2023.

\bibitem{simeoni2004predictive}
Monica Simeoni, Paolo Magni, Cristiano Cammia, Giuseppe~De Nicolao, Valter Croci, Enrico Pesenti, Massimiliano Germani, Italo Poggesi, and Maurizio Rocchetti.
\newblock Predictive pharmacokinetic-pharmacodynamic modeling of tumor growth kinetics in xenograft models after administration of anticancer agents.
\newblock {\em Cancer Research}, 64(3):1094--1101, 2004.

\bibitem{sun2023retentive}
Yutao Sun, Li~Dong, Shaohan Huang, Shuming Ma, Yuqing Xia, Jilong Xue, Jianyong Wang, and Furu Wei.
\newblock Retentive network: A successor to transformer for large language models.
\newblock {\em arXiv preprint arXiv:2307.08621}, 2023.

\bibitem{sutskever2014sequence}
Ilya Sutskever, Oriol Vinyals, and Quoc~V Le.
\newblock Sequence to sequence learning with neural networks.
\newblock {\em Advances in neural information processing systems}, 27, 2014.

\bibitem{tallec2018can}
Corentin Tallec and Yann Ollivier.
\newblock Can recurrent neural networks warp time?
\newblock {\em arXiv preprint arXiv:1804.11188}, 2018.

\bibitem{vaswani2017attention}
Ashish Vaswani, Noam Shazeer, Niki Parmar, Jakob Uszkoreit, Llion Jones, Aidan~N Gomez, {\L}ukasz Kaiser, and Illia Polosukhin.
\newblock Attention is all you need.
\newblock {\em Advances in neural information processing systems}, 30, 2017.

\bibitem{vlachas2020backpropagation}
Pantelis-Rafail Vlachas, Jaideep Pathak, Brian~R Hunt, Themistoklis~P Sapsis, Michelle Girvan, Edward Ott, and Petros Koumoutsakos.
\newblock Backpropagation algorithms and reservoir computing in recurrent neural networks for the forecasting of complex spatiotemporal dynamics.
\newblock {\em Neural Networks}, 126:191--217, 2020.

\bibitem{wang2021learning}
Sifan Wang, Hanwen Wang, and Paris Perdikaris.
\newblock Learning the solution operator of parametric partial differential equations with physics-informed \uppercase{D}eep\uppercase{ON}ets.
\newblock {\em Science advances}, 7(40):eabi8605, 2021.

\bibitem{wolf1985determining}
Alan Wolf, Jack~B Swift, Harry~L Swinney, and John~A Vastano.
\newblock Determining \uppercase{L}yapunov exponents from a time series.
\newblock {\em Physica D: nonlinear phenomena}, 16(3):285--317, 1985.

\bibitem{zhu2023reliable}
Min Zhu, Handi Zhang, Anran Jiao, George~Em Karniadakis, and Lu~Lu.
\newblock Reliable extrapolation of deep neural operators informed by physics or sparse observations.
\newblock {\em Computer Methods in Applied Mechanics and Engineering}, 412:116064, 2023.

\end{thebibliography}
\end{document}